\documentclass[11pt]{article}

\makeatletter
\@ifundefined{nolinenumbers}{
  \newcommand{\nolinenumbers}{}
}{}
\makeatother

\usepackage[final]{acl} % Changed to final for typical submission, adjust if needed

\usepackage{times}
\usepackage{latexsym}
\usepackage{graphicx}      % <- for \resizebox
\usepackage{tikz}
\usetikzlibrary{arrows.meta,positioning,calc,shapes.geometric}
\usepackage[ruled,vlined,linesnumbered]{algorithm2e}
\usepackage{cuted} % For the 'strip' environment
\usepackage{caption} % For the '\captionof' command

\usepackage[T1]{fontenc}
\usepackage[utf8]{inputenc}

\usepackage{microtype}

\usepackage{inconsolata}

\usepackage{multirow}
\usepackage{array}

\usepackage{hyperref}

\usepackage{amsmath}

\usepackage{booktabs,tabularx,longtable,rotating,pdflscape,xurl}

\usepackage[section]{placeins}
\usepackage{url}     % for breakable \url
\usepackage{stfloats}
\title{ESGenius: Benchmarking LLMs on Environmental, Social, and Governance (ESG) and Sustainability Knowledge}

\author{
  \textbf{Chaoyue He}\textsuperscript{1} \quad
  \textbf{Xin Zhou}\textsuperscript{1} \thanks{Corresponding author}\quad
  \textbf{Yi Wu}\textsuperscript{1} \quad
  \textbf{Xinjia Yu}\textsuperscript{1} \quad
  \textbf{Yan Zhang}\textsuperscript{1} \quad
  \textbf{Lei Zhang}\textsuperscript{1} \quad
  \textbf{Di Wang}\textsuperscript{1} \\
  \textbf{Shengfei Lyu}\textsuperscript{1} \quad
  \textbf{Hong Xu}\textsuperscript{1} \quad
  \textbf{Xiaoqiao Wang}\textsuperscript{2} \quad
  \textbf{Wei Liu}\textsuperscript{2} \quad
  \textbf{Chunyan Miao}\textsuperscript{1} \\
  \textsuperscript{1}Alibaba-NTU Global e-Sustainability CorpLab (ANGEL), Singapore; \qquad
  \textsuperscript{2}Alibaba Group, China \\
  \texttt{\{cyhe, xin.zhou, xinjia.yu, zhang.yan, wangdi, shengfei.lyu, xuhong, ascymiao\}@ntu.edu.sg} \\
  \texttt{\{wuyi0614, leizhanzzl.1103\}@gmail.com} \qquad
  \texttt{\{nerissa.wxq, weiliu.liuwei\}@alibaba-inc.com} \\
  \textbf{GitHub:} \url{https://github.com/ANGEL-NTU/ESGenius} \\
  \textbf{Web Portal:} \url{https://angel-ntu.github.io/ESGenius}
}

\begin{document}
\nolinenumbers
\maketitle % Creates title and author block

\begin{strip}
\centering
    \vspace{-8pt}
    \includegraphics[width=0.96\textwidth, trim=5 2 10 5, clip]{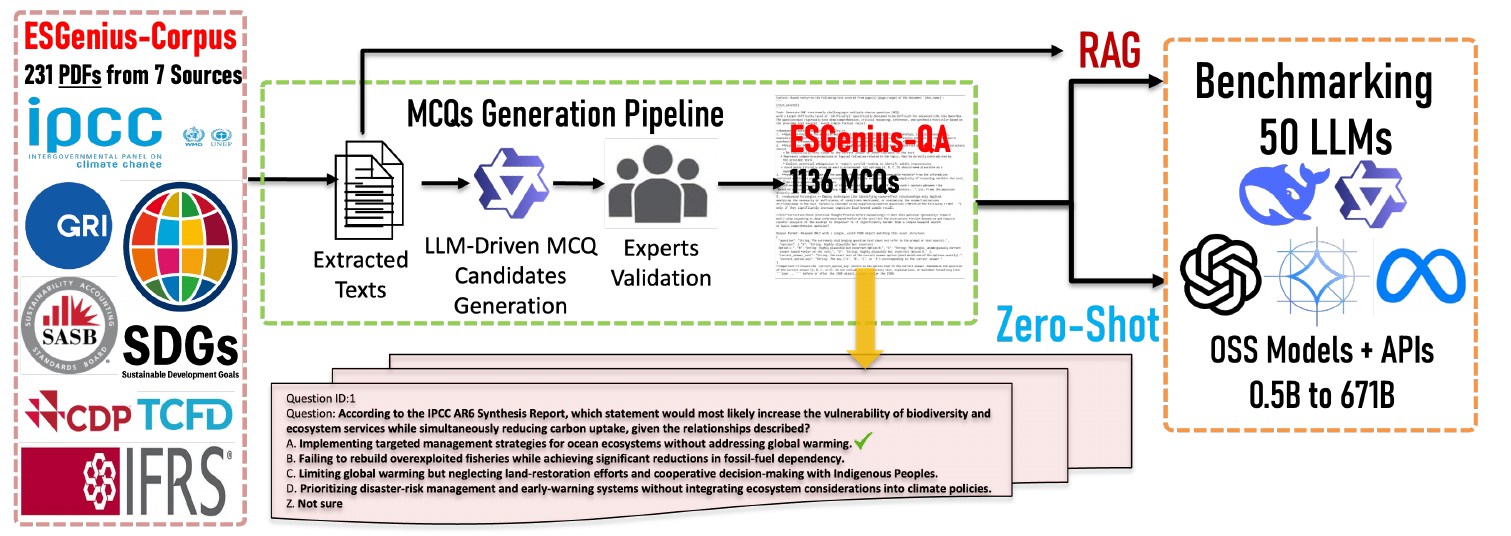}
    \captionof{figure}{The \textbf{ESGenius} pipeline for benchmark creation and model evaluation. The process begins with the \textbf{ESGenius-Corpus}, a curated collection of 231 authoritative PDFs from 7 key ESG sources. Text is extracted from these documents and used in an LLM-driven pipeline to generate candidate MCQs. These questions then undergo rigorous validation by domain experts to produce the final \textbf{ESGenius-QA} dataset, which contains 1136 high-quality MCQs (An example question is shown above). Finally, this dataset is used to benchmark 50 different LLMs from 5 renowned families, with sizes ranging from 0.5B to 671B parameters, via open-source models (OSS) or proprietary APIs. The evaluation is performed in two settings: a \textbf{zero-shot} setup to test the models' inherent ESG knowledge and a \textbf{RAG} setup to assess their ability to synthesize information from the source documents.}
    \vspace{-6pt}
    \label{fig:esgenius_pipeline}
\end{strip}

\begin{abstract}
  We introduce \textbf{ESGenius}, a comprehensive benchmark for evaluating and enhancing the proficiency of Large Language Models (LLMs) in Environmental, Social, and Governance (ESG) and sustainability-focused question answering. \textbf{ESGenius} comprises two key components: (i) \textbf{ESGenius-QA}, a collection of \textbf{1,136} Multiple-Choice Questions (MCQs) generated by LLMs and rigorously validated by domain experts, covering a broad range of ESG pillars and sustainability topics. Each question is systematically linked to its corresponding source text, enabling transparent evaluation and supporting Retrieval-Augmented Generation (RAG) methods; and (ii) \textbf{ESGenius-Corpus}, a meticulously curated repository of \textbf{231} foundational frameworks, standards, reports, and recommendation documents from \textbf{7} authoritative sources. Moreover, to fully assess the capabilities and adaptation potential of LLMs, we implement a rigorous two-stage evaluation protocol---\emph{Zero-Shot} and \emph{RAG}. Extensive experiments across \textbf{50} LLMs (0.5B to 671B) demonstrate that state-of-the-art models achieve only moderate performance in zero-shot settings, with accuracies around 55--70\%, highlighting a significant knowledge gap for LLMs in this specialized, interdisciplinary domain. However, models employing RAG demonstrate significant performance improvements, particularly for smaller models. For example, \texttt{DeepSeek-R1-Distill-Qwen-14B} improves from 63.82\% (zero-shot) to 80.46\% with RAG. These results demonstrate the necessity of grounding responses in authoritative sources for enhanced ESG understanding. To the best of our knowledge, ESGenius is the first comprehensive QA benchmark designed to rigorously evaluate LLMs on ESG and sustainability knowledge, providing a critical tool to advance trustworthy AI in this vital domain.
\end{abstract}

\section{Introduction}
\label{sec:introduction}

\begin{figure*}[!b]
  \centering
  \includegraphics[width=\textwidth]{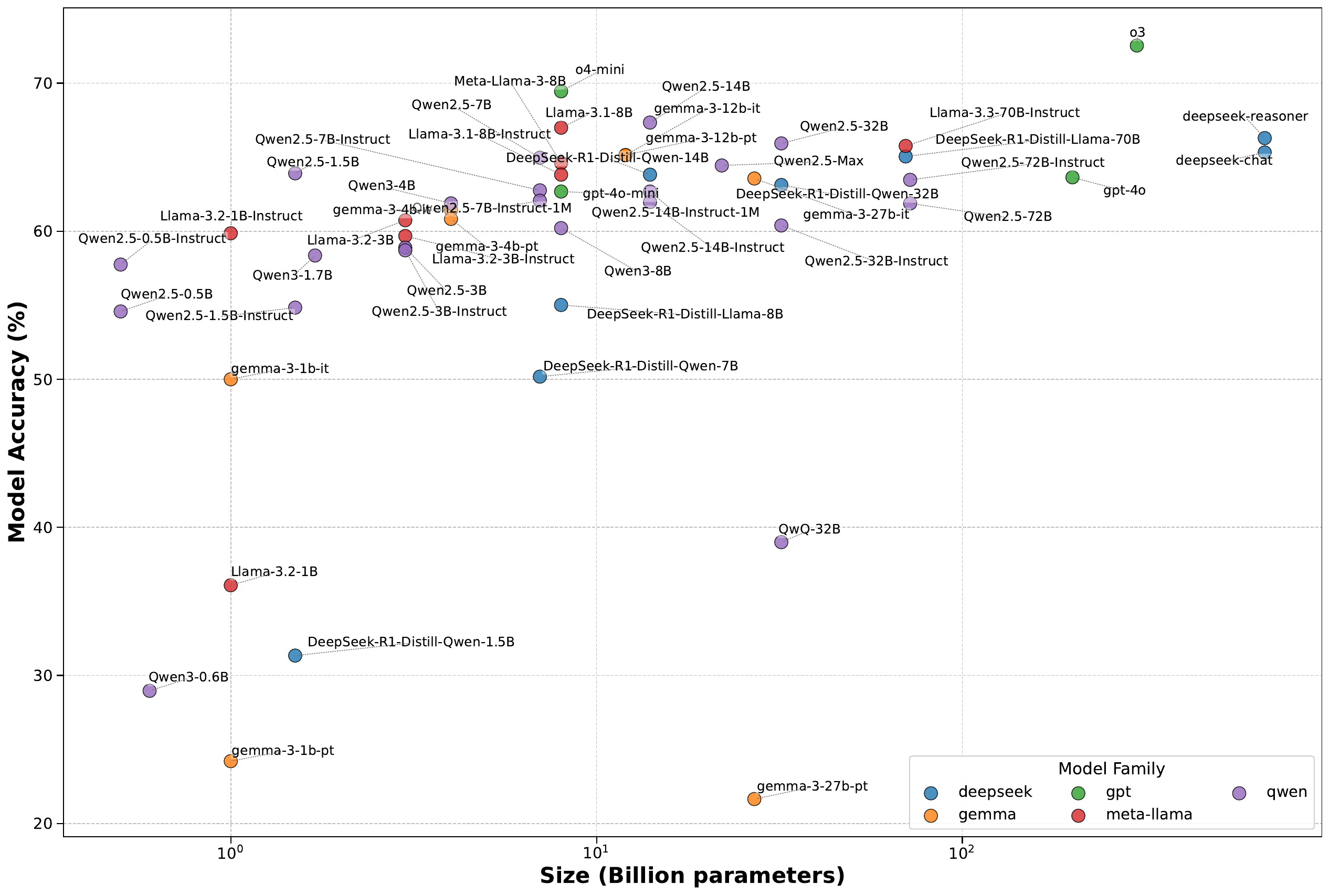}
  \caption{Relationship between model size and zero-shot accuracy across 50 LLMs evaluated on the ESGenius benchmark. Model sizes are plotted on a $\log_{10}$ scale (in billions of parameters), with accuracy shown as percentages. There is a moderate positive correlation between model size and performance, suggesting larger models generally perform better. Dot colors denote five model families (DeepSeek, Gemma, GPT, Meta-Llama, Qwen). For proprietary API models, parameter counts are based on industry estimates (See Table~\ref{tab:benchmark_results} for details).}
  \label{fig:model_size_vs_performance}
\end{figure*}

Environmental, Social, and Governance (ESG) knowledge encompasses a vast domain of sustainability and corporate responsibility information that Large Language Models (LLMs) must effectively process to serve emerging business needs \cite{zhou2024advancing, singh2025survey, ong2025towards, zou2025esgreveal, zhang2024greenrec, yu2025report}. This field spans critical areas from climate change and emissions tracking \cite{IPCC_AR6} to workplace safety and human rights \cite{GRI_Social}. Such knowledge is codified in technical frameworks and standards---major ones including GRI \cite{GRI_Standards}, SASB \cite{SASB_Standards}, TCFD \cite{TCFD_Recommendations}, ISSB \cite{IFRS_ISSB_Standards}, and CDP \cite{CDP_Framework}—all of which are constantly evolving.

While LLMs show promise in processing complex ESG documents and providing relevant answers to user queries, their capabilities in this interdisciplinary domain remain \textit{largely unevaluated}. Considering the high stakes involved, where incorrect responses about ESG requirements or metrics could lead to serious compliance violations or misguided sustainability initiatives, this assessment gap is particularly problematic. However, there is \textbf{currently no comprehensive question-answering (QA) benchmark specifically designed to evaluate how well LLMs understand and reason about ESG concepts}. Existing QA benchmarks either omit ESG topics entirely or address them only superficially. This gap leaves researchers and practitioners without a reliable way to measure and improve LLMs' ESG knowledge comprehension and question-answering abilities.

To close this gap, we present \textbf{ESGenius}, a curated benchmark that targets Multiple-Choice Questions (MCQs) answering as the core evaluation task for ESG and sustainability knowledge (Figure~\ref{fig:esgenius_pipeline}). Our contributions are fourfold: (1) \textbf{ESGenius} benchmark: A comprehensive evaluation framework comprising two integrated components: (i) \textbf{ESGenius-QA}, a collection of 1,136 MCQs across various ESG pillars and sustainability topics, generated using state-of-the-art (SOTA) LLM approaches and validated by domain experts. Each question is explicitly mapped to supporting evidence from authoritative source texts, enabling transparent evaluation and facilitating RAG applications; and (ii) \textbf{ESGenius-Corpus}, a collection of 231 ESG documents and frameworks enabling efficient knowledge retrieval from 7 major authoritative sources. (2) \textbf{Evaluation Protocol}: We implement a comprehensive two-stage evaluation protocol consisting of zero-shot testing and RAG to systematically assess LLM capabilities. This assessment framework provides valuable insights into the current limitations and future potential of LLMs in understanding ESG. (3) \textbf{Evaluation Analysis}: Testing across a diverse set of 50 LLMs (0.5B to 671B) reveals significant performance gaps in zero-shot settings, with most models achieving 55--70\% accuracy and the best model (\texttt{o3}) achieving a top score of 72.54\%, as shown in Figure~\ref{fig:model_size_vs_performance}. However, models demonstrate substantial potential for improvement through RAG approaches, with \texttt{DeepSeek-R1-Distill-Qwen-14B} improving from 63.82\% (zero-shot) to 80.46\%. (4) \textbf{Open Source Initiative}: To foster community engagement and collaborative advancement, we have made our complete benchmark suite publicly available at \url{https://github.com/ANGEL-NTU/ESGenius}. This includes comprehensive documentation, evaluation code, model implementations, and the full ESGenius dataset. We also maintain an interactive web portal at \url{https://angel-ntu.github.io/ESGenius} featuring a real-time leaderboard and detailed performance visualizations through heatmap and one of its cells (Figures~\ref{fig:zero_shot_heatmap} and~\ref{fig:heatmap_cell_info}), enabling researchers to track progress and identify areas for improvement in ESG-focused language models.

The remainder of this paper is organized as follows: \S\ref{sec:related_work} surveys prior efforts on ESG knowledge resources, question-answering benchmarks, and retrieval-augmented generation; \S\ref{sec:the_esgenius_benchmark} details the construction of the \textbf{ESGenius} benchmark, describing both the \textbf{ESGenius-QA} dataset and the \textbf{ESGenius-Corpus}; \S\ref{sec:experiments_and_results} outlines our experimental protocols—including zero-shot and RAG settings—and presents a comprehensive evaluation of 50 LLMs; \S\ref{sec:conclusion} concludes the paper and highlights directions for subsequent research; \S\ref{sec:limitations} discusses the limitations of our benchmark, while \S\ref{sec:ethical_considerations} reflects on ethical considerations.

\section{Related Work}
\label{sec:related_work}

Prior work at the intersection of LLMs and ESG has spanned diverse strands of research, from building domain-specific corpora and ontologies to designing evaluation benchmarks and retrieval-augmented generation methods. In this section, we review relevant literature across three key areas: (i) \emph{ESG and sustainability knowledge resources}, which provide the taxonomies, corpora, and specialized models needed to capture domain-specific terminology; (ii) \emph{QA and evaluation benchmarks}, which illustrate how standardized datasets have driven progress in measuring factual knowledge, reasoning, and domain expertise, while also exposing the absence of ESG-focused QA benchmarks; and (iii) \emph{RAG and knowledge grounding}, which highlight how retrieval mechanisms enhance interpretability and factual accuracy in specialized domains such as finance and climate science. 

\subsection{ESG \& Sustainability Knowledge Resources}
\label{sec:esg_and_sustainability_knowledge_resources}
ESG data have traditionally been guided by voluntary reporting frameworks such as the Global Reporting Initiative (GRI) and the Sustainability Accounting Standards Board (SASB). These frameworks define taxonomies of ESG topics (e.g., emissions, labor practices, board governance) and indicators, but the unstructured text in corporate sustainability reports poses significant challenges for computational use. To impose structure, researchers have proposed ontologies and knowledge bases for ESG. For instance, \citet{zhou2023ontosustain} introduced \textit{OntoSustain}, aligning GRI and EU standards (ESRS) to capture key sustainability indicators, though broad coverage and automated population remain difficult.

Several ESG-focused text corpora have emerged to support NLP research in this domain. One recent example is \textit{SusGen-3K}, a dataset of 30k instances spanning multiple financial NLP tasks (e.g., sentiment, Q\&A) and an ESG report generation task \cite{wu2025susgen}. Meanwhile, \citet{chang2024nlp} built a benchmark of more than 8,000 labeled sentences from more than 14,000 corporate sustainability reports covering 36 ESG topics for topic and quality classification. Other datasets focus on more specific areas, such as the benchmark from \citet{beck2025addressing} for extracting greenhouse gas emissions, or on non-English languages, like the ESG-Kor dataset for information extraction in Korean \cite{lee2024esg}. These datasets remain relatively restricted, especially compared to general-domain corpora, and focus primarily on classification or generation tasks rather than MCQs question answering, thus motivating the need for broader coverage like \textbf{ESGenius}.

Prior NLP work on sustainability texts has largely centered around information extraction and text classification, evolving from early methods that relied on topic models or lexicon-based approaches \cite{raghupathi2020identifying} but often fell short in capturing nuanced ESG terminology. More recent efforts leverage LLMs, either by developing specialized models or by applying general models in sophisticated pipelines. In the former category, researchers have introduced domain-specific models like \textit{ClimateBERT}, which continued BERT's pre-training on 1.6M climate-related paragraphs for more effective risk disclosure classification \cite{webersinke2021climatebert}, and \textit{E-BERT}, a set of models designed to optimize ESG reporting \cite{zhang2025optimizing}. In the latter category, approaches like \textit{ChatReport} use GPT-based summarization and semantic search to assess a company's report against TCFD recommendations \cite{ni2023chatreport}, while others have employed GPT-based pipelines to extract triples (\emph{company, action, ESG issue}) and populate knowledge graphs from sustainability disclosures \cite{bronzini2024glitter}. Collectively, these advanced approaches illustrate the promise of LLMs for ESG tasks but also highlight the need for standardized benchmarks for a thorough evaluation.

\subsection{QA \& Evaluation Benchmarks}
\label{sec:question_answering_and_evaluation_benchmarks}
Question answering (QA) benchmarks have driven many NLP advancements over the past decade, starting with the TREC QA tracks \cite{voorhees1999trec} and continuing through datasets like \textit{SQuAD} \cite{rajpurkar2016squad}. While these resources predominantly test general factual or reading comprehension, specialized QA datasets have emerged to measure higher-level reasoning and domain expertise. For example, \textit{OpenBookQA} \citet{mihaylov2018can} challenged models to combine a small ``open book'' of scientific facts with common sense, while \textit{BoolQ} \cite{clark2019boolq} featured yes/no queries requiring nuanced text understanding. Datasets such as \textit{DROP} \cite{dua2019drop} emphasize discrete reasoning skills, involving counting or arithmetic over paragraphs.

Further expansions have produced large-scale evaluation suites such as \textit{MMLU} (Massive Multitask Language Understanding) \cite{hendrycks2020measuring} and \textit{MMMU} (Massive Multi-Discipline Multimodal Understanding) \cite{yue2024mmmu}, which cover dozens of subjects at the college-exam level, along with domain-specific benchmarks like \textit{PubMedQA} \cite{jin2019pubmedqa} for biomedical research, \textit{JEC-QA} \cite{zhong2020jec} for legal question answering, and \textit{FinanceBench} \cite{islam2023financebench} for finance. Yet despite these advances, there is still a notable lack of QA benchmarks dedicated to ESG and sustainability. To address this gap, \textbf{ESGenius} introduces a large collection of expert-validated MCQs spanning environmental, social, and governance topics. Drawing from authoritative frameworks (e.g., IPCC, GRI, SASB), \textbf{ESGenius} provides a high-stakes, real-world testbed tailored to ESG.

\subsection{RAG and Knowledge Grounding}
\label{sec:rag_and_knowledge_grounding}
When tackling specialized or dynamically updated domains, RAG has emerged as a powerful method for accurate and interpretable QA. \citet{lewis2020retrieval} introduced \emph{RAG}, an approach that combines a parametric language model with a non-parametric memory of documents. This approach improved performance on knowledge-intensive QA tasks relative to purely parametric models, as it allows the LLM to ground responses in retrieved evidence.

Other retrieval-generation frameworks such as \emph{GraphRAG} \cite{edge2024local}, \emph{Fusion-in-Decoder} \cite{izacard2021leveraging} and \emph{REALM} \cite{guu2020retrieval} illustrated diverse strategies for fetching relevant text chunks to guide the model's reasoning. Recently, these methods have been applied to high-stakes sectors including finance and climate science. For instance, \citet{vaghefi2023chatclimate} integrated GPT-4 with a 3,000-page IPCC report to answer climate change questions more accurately. Similarly, \citet{ni2023chatreport} used semantic search to retrieve TCFD-related segments from a company's sustainability report, prompting an LLM to summarize potential climate risks. This grounding is critical in ESG contexts, where traceability and authoritative references are essential.

\section{The ESGenius Benchmark}
\label{sec:the_esgenius_benchmark}

Developing reliable AI tools for the complex ESG and sustainability domain requires high-quality, domain-specific data. Our \textbf{ESGenius} benchmark provides a comprehensive foundation to evaluate and enhance LLMs in this context, drawing from globally recognized standards and authoritative sources. As illustrated in Figure~\ref{fig:esgenius_pipeline}, the benchmark is built through a structured pipeline where ESG-related documents are digitized, LLMs generate candidate MCQs, and domain experts validate them to form \textbf{ESGenius-QA}, while the underlying source materials are systematically curated into the \textbf{ESGenius-Corpus}. Together, these two interconnected components ensure explicit linkages between questions and textual evidence, enabling transparent evaluation of ESG knowledge and supporting RAG methods. The detailed coverage of sources and statistics is presented in Appendix~\ref{sec:esgenius_corpus_details}.

\subsection{ESGenius-QA}
\label{sec:esgenius_qa}

\paragraph{QA Generation and Validation Principles.}\label{sec:qa_generation_and_validation_principles}
The \textbf{ESGenius-QA} dataset comprises 1,136 carefully curated MCQs with ground truth answers, validated by ESG domain experts. The questions systematically cover essential ESG topics across three core pillars including but not limited to: (i) \textbf{Environmental:} Climate change mitigation and adaptation, comprehensive carbon accounting (Scope 1, 2, 3 emissions), energy efficiency, water resource management, biodiversity conservation, waste reduction, and pollution control, etc; (ii) \textbf{Social:} Labor practices and standards, human rights protection, diversity and inclusion initiatives, workplace health and safety, community engagement and impact assessment, data privacy and protection, etc; (iii) \textbf{Governance:} Board structure and independence, executive compensation frameworks, business ethics and compliance, shareholder rights and engagement, enterprise risk management, and regulatory adherence, etc. The dataset is generated through our specialized pipeline (Appendix~\ref{sec:mcq_generation_pipeline}) and undergoes rigorous validation by domain experts to ensure accuracy and relevance (\S\ref{sec:qa_quality_control}).

\paragraph{Descriptions.}\label{sec:esgenius_qa_descriptions}
The \textbf{ESGenius-QA} dataset is structured to reflect the complexity and diversity of real-world ESG assessments through a standardized multiple-choice format. Each question contains four options (A-D) with a single correct answer, plus a dedicated ``Not sure'' option (Z) to capture model uncertainty. This design enables systematic evaluation of both performance and confidence. The dataset carefully balances conciseness with domain-specific precision. Questions and options employ precise ESG terminology while maintaining clarity and focus. Contextual prompts are provided where necessary for essential background. The specialized lexicon throughout authentically represents real-world ESG assessment scenarios. Table~\ref{tab:token_stats} presents a detailed token-level analysis of questions, options, and source text. Key terminology distributions are visualized through word clouds for questions (Figure~\ref{fig:esgenius_qa_question_wordcloud}), answer options (Figure~\ref{fig:esgenius_qa_option_wordcloud}) and the source text (Figure~\ref{fig:esgenius_corpus_wordcloud}).

\paragraph{Corpus-Question Mapping.}\label{sec:corpus_question_mapping}
Each question in \textbf{ESGenius-QA} is precisely mapped to relevant page(s) and text passages within the 7 authoritative source documents (GRI, SASB, IPCC, etc.) containing the information required for correct answers. ESG domain experts meticulously validate these mappings, ensuring both accuracy and relevance. This explicit linking between questions and their supporting evidence in source texts enables effective RAG approaches.

\paragraph{QA Quality Control.}\label{sec:qa_quality_control} 
To ensure the highest quality of questions and answers, we employ a multi-stage validation process (Appendix~\ref{sec:mcq_generation_pipeline}). While \texttt{Qwen-max-2025-01-25} was used in the initial phase to generate candidate QA pairs, the final benchmark was \textbf{entirely curated by human experts}: six independent reviewers with an average of five years’ experience in sustainability or NLP double-validated every question-answer pair, supported by three consulting industry experts with over ten years of relevant experience. Validators confirmed factual correctness against authoritative sources, clarity and unambiguity, plausibility yet incorrectness of distractors, and adequate context. Any question failing these criteria was refined or discarded, yielding a rejection rate of about 25\% (1,136 of 1,519 MCQs retained). Importantly, \texttt{Qwen-max-2025-01-25} was later evaluated only through its \textbf{official API} in a \textbf{fully isolated session}, ensuring no overlap or leakage between construction and evaluation. This expert-in-the-loop curation decisively eliminates circular dependency, and, consistent with standard practices in high-impact benchmarks, our method combines LLM-assisted generation with expert oversight to balance scale, quality, and domain relevance, while remaining open to validated alternatives to further reduce bias.

\subsection{ESGenius-Corpus}
\label{sec:esgenius_corpus}

The \textbf{ESGenius-Corpus} underpins every evaluation in this work, unifying authoritative frameworks, corporate questionnaires, and peer-reviewed scientific assessments that span the full breadth of environmental, social, and governance (ESG) concerns. A quantitative snapshot of the collection appears in Table~\ref{tab:ipcc_table} for IPCC climate-change assessments, Table~\ref{tab:gri_table} for the GRI Standards, Tables~\ref{tab:sasb_part1_table} and \ref{tab:sasb_part2_table} for SASB’s standards, Tables~\ref{tab:ifrsissb_part1_table} and \ref{tab:ifrsissb_part2_table} for IFRS/ISSB guidance, Table~\ref{tab:tcfd_table} for TCFD materials, Table~\ref{tab:cdp_table} for CDP questionnaires, and Table~\ref{tab:sdg_table} for the UN Sustainable Development Goals (SDGs). These statistics confirm (i) broad topical coverage across all three ESG pillars, (ii) deep sector-level granularity via GRI and SASB industry standards, and (iii) an expanding climate-finance focus through IFRS-ISSB, TCFD, and CDP additions.

\paragraph{Sourcing and Collection.}\label{sec:esgenius_corpus_sourcing_and_collection} The corpus integrates sources along three tiers: (1) \emph{Core reporting standards} include the Global Reporting Initiative (GRI) Standards~\cite{GRI_Standards}, Sustainability Accounting Standards Board (SASB) Standards~\cite{SASB_Standards}, and International Sustainability Standards Board (IFRS-ISSB) Standards~\cite{IFRS_ISSB_Standards}; (2) \emph{Specialized reports and frameworks} extend coverage with IPCC Assessment Reports~\cite{IPCC_AR6}, TCFD Guidelines~\cite{TCFD_Recommendations}, and CDP questionnaires~\cite{CDP_Framework}; and (3) \emph{Overarching global targets} are provided by the UN SDGs~\cite{UN_SDGs}.  

\paragraph{Extensibility.}\label{sec:esgenius_corpus_extensibility} 
The architecture of the ESGenius-Corpus is deliberately modular, enabling the rapid integration of new materials such as emerging regulations (e.g., EU CSRD, ISSB updates), sector-specific guidance (e.g., sustainable finance taxonomies), country-level policies, corporate ESG and sustainability reports, and certification frameworks (e.g., LEED for sustainable buildings). This forward-looking design ensures that the benchmark can evolve in step with the shifting ESG landscape, safeguarding its long-term relevance and maintaining ESGenius as a reliable yardstick for tracking LLM progress in sustainability understanding.

\paragraph{Copyrights and Privacy Handling.}\label{sec:esgenius_corpus_copyrights_and_privacy_handling} The ESGenius-Corpus comprises publicly available documents and open-access materials. For proprietary standards and frameworks (e.g., GRI, SASB, CDP), we provide references to their official source locations, allowing users to access them directly, with only brief excerpts referenced for academic purposes. The corpus documentation explicitly lists included sources and provides detailed guidance on accessing external materials, ensuring reproducible research while adhering to intellectual property rights. To protect privacy, the corpus excludes all personal and sensitive information through a rigorous manual auditing process, conducted alongside the source mapping step (\S\ref{sec:corpus_question_mapping}) and the quality control stage for dataset curation (\S\ref{sec:qa_quality_control}).

\section{Experiments and Results}
\label{sec:experiments_and_results}

Through \textbf{ESGenius}, we benchmark \textbf{50} LLMs spanning open-source checkpoints---DeepSeek~\cite{guo2025deepseek}, Meta-Llama 3~\cite{grattafiori2024llama}, Google Gemma 3~\cite{team2025gemma}, and Alibaba Qwen 2.5/3~\cite{yang2024qwen2, yang2025qwen3} from Hugging Face~\cite{huggingface_hub} and Transformers library~\cite{wolf-etal-2020-transformers}---as well as proprietary APIs (GPT, DeepSeek, Qwen).  
All experiments run on a DGX node (4 $\times$ 80 GB A100 GPUs) with fixed random seeds (\textsc{seed}=42) for reproducibility.  
Our evaluation suite comprises two progressively stronger settings: (1) \textbf{Zero-Shot Prompting} (\S\ref{sec:zero_shot_prompting_with_llms}), which probes the intrinsic ESG knowledge encoded during generic pre-training, and (2) \textbf{Long-Context RAG} (\S\ref{sec:long_context_rag}), a RAG baseline that prepends relevant evidence to each prompt as long context. The RAG experiments are conducted on 43 open-source models.

For each model-setting combination, we compute exact-match accuracy on single-answer questions. Because LLMs often generate raw outputs that may not directly match the required format, we apply response validation to ensure valid multiple-choice answers (Appendix~\ref{sec:llm_raw_response_validation}). The results are presented as a comprehensive leaderboard (Appendix~\ref{sec:main_experimental_results}, Table~\ref{tab:benchmark_results}) for both zero-shot and RAG settings. To complement these results, we also include a ranking bar plot in Figure~\ref{fig:esgenius_ranking}, providing a clearer visualization of LLM performance under the zero-shot setup.

\subsection{Evaluation Protocols}
\label{sec:evaluation}
To rigorously assess model performance on \textbf{ESGenius}, we design two complementary evaluation protocols. The first, \textbf{Zero-Shot Prompting}, measures how well models can answer ESG-related questions without external assistance, reflecting their intrinsic knowledge and reasoning ability. The second, \textbf{Long-Context RAG}, evaluates models in a retrieval-augmented setting by providing each question with its corresponding evidence passage from the ESGenius-Corpus. Together, these protocols enable a transparent comparison between raw model capabilities and knowledge-grounded reasoning, thereby establishing fair and reproducible baselines for future research.

\subsubsection{Zero-Shot Prompting}
\label{sec:zero_shot_prompting_with_llms}

\paragraph{Inference Protocol.} \label{sec:zero_shot_prompting_inference_protocol}
Each model is provided the zero-shot prompt template (Appendix~\ref{sec:zero_shot_prompt}) and an zero-shot question example is shown in Appendix~\ref{sec:mcq_zero_shot_example}. Response generation employs greedy decoding (\texttt{temperature}=0, \texttt{top\_p}=1, \texttt{top\_k}=0), with a maximum sequence length of 1024 tokens.

\paragraph{Implementation.} \label{sec:zero_shot_prompting_implementation}
Models are evaluated in half precision (FP16) using standard batched inference on GPUs. Results are logged incrementally with error handling to ensure robustness.

\subsubsection{Long-Context RAG}
\label{sec:long_context_rag}

\paragraph{Retrieval Protocol.}\label{sec:long_context_rag_retrieval_protocol} 
For each question, the pre-linked source passage from \textbf{ESGenius-QA} is retrieved from the ESG knowledge corpus. This passage is prepended to the prompt template (Appendix~\ref{sec:rag_prompt}), and an illustrative example is provided in Appendix~\ref{sec:mcq_rag_example}. This simple but effective retrieval ensures that models consistently access relevant domain knowledge during inference.  

\paragraph{Implementation.}\label{sec:long_context_rag_implementation} 
The evaluation pipeline mirrors the zero-shot setup, with the key difference being the augmented prompts. Context is injected by prepending the retrieved passage to each question prompt before feeding it to the model, using the same decoding parameters (\texttt{temperature}=0, \texttt{top\_p}=1) and evaluation metrics. Notably, even this straightforward lexical-matching long-context RAG method yielded measurable performance gains, underscoring the benchmark’s utility. While more sophisticated retrieval-augmented generation techniques could further improve performance, \textbf{our aim was to establish a transparent, fair, and reproducible RAG baseline}.

\subsection{Main Results}
\label{sec:main_results}

Drawing from the experimental results presented in Table~\ref{tab:benchmark_results}, we highlight \textbf{three} key findings and provide detailed analysis from various perspectives subsequently.
\subsubsection{Key Findings}
\label{sec:key_findings}

(1) \textbf{ESG Concept Understanding Remains Challenging for Current LLMs}. 
Our evaluation reveals that the top-performing zero-shot model, \texttt{o3}, attained an accuracy of 0.7254, whereas the majority of other models scored approximately 0.65. This performance level is notably comparable to, or lower than, that observed in benchmarks from other domains~\cite{guha2023legalbench}, underscoring the challenges current LLMs face in accurately comprehending and reasoning about complex ESG concepts. Further details are provided in \S\ref{sec:challenging_nature}.

(2) \textbf{Domain-Specific RAG Allows Smaller Models to Outperform Larger Zero-Shot Counterparts.}  
Performance evaluation demonstrates that applying RAG significantly enhances the capabilities of smaller models on ESG question answering. For example, the accuracy of \texttt{Qwen3-4B} increased from 0.6188 (zero-shot) to 0.7905 (RAG). Similarly, \texttt{Qwen2.5-1.5B} rose from 0.5484 (zero-shot) to 0.6972 (RAG). These improvements demonstrate that targeted retrieval is more effective than raw model scale for complex domain-specific knowledge. More details are described in \S\ref{sec:rag_performance}.

(3) \textbf{Reasoning Capabilities Enhance LLM Performance on ESG Understanding.}  
Models augmented with explicit reasoning capabilities demonstrate enhanced performance on ESG tasks compared to non-reasoning models of comparable scale. For instance, the reasoning variant of \texttt{o3} attained an accuracy of 0.7254, exceeding the scores of similarly sized models without such enhancements (Figure~\ref{fig:model_size_vs_performance}). This underscores the significant benefit derived from incorporating reasoning-focused mechanisms into LLMs for specialized domains. More details in \S\ref{sec:reasoning_models_analysis}.

\subsubsection{Challenging Nature of ESGenius}
\label{sec:challenging_nature}
Our comprehensive evaluation, detailed in Table~\ref{tab:benchmark_results}, reveals that the ESGenius-QA dataset is highly challenging for a broad range of contemporary LLMs. This includes proprietary APIs such as \texttt{GPT-4o} (zero-shot: 0.6364), \texttt{DeepSeek-R1} (zero-shot: 0.6629), \texttt{DeepSeek-V3} (zero-shot: 0.6532), and \texttt{Qwen2.5-Max} (zero-shot: 0.6444), as well as leading open-source families like Meta Llama (e.g., \texttt{Meta-Llama-3-8B} zero-shot: 0.6461), Google Gemma (e.g., \texttt{Gemma-3-12B-Instruct} zero-shot:0.6514, \texttt{Qwen2.5-14B} zero-shot: 0.6734). The best zero-shot performance in the table is 0.7254 from the proprietary \texttt{o3} model, while the highest RAG-enhanced score is 0.8336 with the open-source \texttt{Gemma-3-27B-Instruct} model. This gap underscores both the discriminative nature of ESGenius and the benchmark's heightened ESG reasoning demands.

\subsubsection{RAG Performance}
\label{sec:rag_performance}
Integrating ESG-specific knowledge through long-context RAG frequently yields substantial performance gains over zero-shot capabilities. For instance, the \texttt{DeepSeek-R1-Distill-Qwen} series demonstrates significant improvements across model sizes: the 1.5B model improves by 37.36\% (0.3134 to 0.4305), the 7B model by 29.63\% (0.5018 to 0.6505), and the 14B model by 26.07\% (0.6382 to 0.8046). Other notable examples include \texttt{Gemma-3-27B}, which achieves a remarkable 141.57\% improvement (0.2165 to 0.5229), and \texttt{QwQ-32B} with a substantial 95.23\% increase (0.3900 to 0.7614). Similarly, \texttt{Qwen3-4B} improves by 27.75\% (0.6188 to 0.7905), and the \texttt{Qwen2.5-1.5B-Instruct} model increases by 27.13\% (0.5484 to 0.6972). However, RAG does not universally enhance performance. Some models actually perform worse under the RAG configuration in this study---notably, \texttt{Qwen2.5-0.5B} shows a slight decline of 1.14\% (0.5458 to 0.5396), while its instruction-tuned counterpart experiences a more significant 7.48\% decrease (0.5775 to 0.5343). These results suggest that for some smaller architectures, additional context may function as noise or exceed their ability to leverage long sequences effectively. Overall, the effectiveness of RAG appears strongly dependent on model architecture and intrinsic capacity, yet it often allows many smaller models to outperform the zero-shot performance of much larger ones.

\subsubsection{Reasoning Models Analysis}
\label{sec:reasoning_models_analysis}
Table~\ref{tab:benchmark_results} flags ``Reasoning Focus'' (Rea: Yes) models. Those explicitly oriented or fine-tuned for reasoning, such as \texttt{DeepSeek-R1-Distill-Qwen} and \texttt{DeepSeek-R1-Distill-Llama} (Open Source) and proprietary offerings like \texttt{o3} (300B, Rea: Yes, zero-shot: 0.7254) or \texttt{o4-mini} (3B, Rea: Yes, zero-shot: 0.6945), frequently excel in zero-shot mode. For example, the reasoning-oriented \texttt{DeepSeek-R1-Distill-Qwen} series shows strong RAG performance and gains (14B model: 0.8046, 1.5B model improves by 37.36\%), while \texttt{QwQ-32B} (Rea: Yes) also achieves a 95.23\% improvement under RAG. Comparisons with non-reasoning peers reinforce this pattern. For instance, \texttt{o4-mini} (3B, Rea: Yes) reaches 0.6945 zero-shot, well above non-reasoning 3B models such as \texttt{Llama-3.2-3B} (zero-shot: 0.6074) or \texttt{Qwen2.5-3B} (zero-shot: 0.5889). These observations suggest that training or architectural choices emphasizing reasoning bolster ESG question-answering, especially for multi-step logical tasks.

\subsubsection{Instruct-Tuned Models Analysis}
\label{sec:instruct_tuned_models_analysis}
Our analysis of instruction-tuned models (marked ``I-T: Yes'' in Table~\ref{tab:benchmark_results}) reveals varied performance in zero-shot settings. Several models demonstrate clear benefits from instruction tuning: Google's \texttt{Gemma-3-1B} improves substantially from 0.2421 to 0.5000, while \texttt{Gemma-3-27B} shows an even more dramatic increase from 0.2165 to 0.6356. Similar positive trends appear in Alibaba's \texttt{Qwen2.5-0.5B} (0.5458 to 0.5775) and Meta's \texttt{Llama-3.2-1B} (0.3609 to 0.5986). However, instruction tuning can also lead to performance degradation in some cases: \texttt{Meta-Llama-3.1-8B} declines from 0.6699 to 0.6382, \texttt{Qwen2.5-1.5B} drops from 0.6391 to 0.5484, and \texttt{Qwen2.5-14B} decreases from 0.6734 to 0.6197. The impact of instruction tuning becomes particularly interesting when combined with RAG. In this context, instruction-tuned models frequently demonstrate superior improvement margins compared to their standard counterparts. For instance, the instruction-tuned version of \texttt{Gemma-3-12B} achieves 0.8380 (28.64\% improvement) while its non-instruction-tuned variant reaches only 0.6857 (5.26\% improvement). Similarly, the instruction-tuned \texttt{Llama-3.1-8B} attains 0.7993 (25.24\% improvement) compared to the standard model's 0.7650 (14.20\% improvement).

\subsubsection{Error Analysis}
\label{sec:error_analysis}
Sorting the heatmap (\S\ref{sec:interactive_visualization}) columns by failure rate exposes a long tail of adversarial-grade questions that no model answered correctly. Appendix~\ref{ex:mcq_low_accuracy_examples} reproduces three such zero-accuracy cases: Question 432 from the \textit{SDG 2024 Report} probes subtle regional disparities in per-capita renewable electricity capacity; Question 635, drawn from \textit{IPCC AR6 WG~III}, asks for the most direct barrier to ESCO adoption, where models consistently misattributed the bottleneck to regulation or awareness rather than the report’s emphasis on asymmetric information and split incentives; and Question 1006, based on the \textit{SASB Chemicals} standard, tests the definition of a recordable workplace incident. Across these and similar examples, several error patterns merit closer investigation: a reliance on surface keyword overlap between prompt and distractor, and a tendency to over-generalize to well-known policy frameworks when questions hinge on narrower ESG standards, suggesting that more fine-grained analyses are needed. Future work could therefore combine quantitative error clustering with qualitative case studies, cross-model comparisons, and expert-in-the-loop probing to develop a more comprehensive understanding of the root causes of such failures.

\subsubsection{Sub-topic Analysis}
\label{sec:subtopic_analysis}
A finer-grained sub-topic analysis provides additional insight into model strengths and weaknesses across ESG dimensions. While the current benchmark draws from seven ESG source collections to ensure broad coverage, disaggregating results by ESG pillar (Environmental, Social, Governance) and their sub-themes (\S\ref{sec:qa_generation_and_validation_principles}) offers more actionable perspectives. In practice, \textbf{Environmental} spans climate change, biodiversity, pollution control, circular economy, energy transition, etc; \textbf{Social} covers labor rights, diversity and inclusion, community impacts, human health and safety, equity in supply chains, etc; \textbf{Governance} addresses board accountability, anti-corruption, transparency, reporting compliance, risk management, etc; and \textbf{Cross-Cutting} themes such as just transition, Indigenous knowledge, digital governance further blurring boundaries, etc. This infrastructure lays the foundation for mapping each question to its corresponding ESG pillar or sub-theme. In future versions, we will expand this perspective by (i) reporting per-theme performance scores and (ii) discussing how this lens can guide targeted dataset expansion. As part of this process, we have begun annotating each question with an E, S, G, or Cross-subtopic label. For example, Question ID~1---``According to the IPCC AR6 Synthesis Report, which statement would most likely lead to increased vulnerability of biodiversity and ecosystem services while simultaneously reducing carbon uptake, based on the relationships described?''---is labeled \textbf{E} (Environmental, \emph{Biodiversity}). Question ID~578---``Which initiative best promotes workplace fairness and equal opportunities across different demographic groups within an organization?''---is labeled \textbf{S} (Social, \emph{Diversity and Inclusion}). In contrast, Question ID~1127---``When an organization determines that indirect economic impacts are a material topic, which of the following best describes the conditions under which it can omit specific disclosures without violating reporting requirements, according to GRI 203: Indirect Economic Impacts 2016?''---is labeled \textbf{G} (Governance, \emph{Reporting Compliance}). This annotation process is ongoing and will inform future analytical releases, enabling deeper performance breakdowns across ESG pillars and sub-themes.

\subsubsection{Interactive Visualization}
\label{sec:interactive_visualization}
Figure~\ref{fig:zero_shot_heatmap} presents an interactive heatmap visualization of model performance across 50 LLMs (ranging from 0.5 to 671 billion parameters) and all 1,136 MCQs in \textbf{ESGenius}. The visualization arranges models (rows) by their overall zero-shot accuracy and questions (columns) by difficulty, creating a clear left-to-right progression from predominantly red cells (incorrect/unanswered questions) to green cells (correct answers). Interactive features enhance analysis - hovering over any cell reveals detailed information including the model name and rank, question ID, difficulty score, complete prompt, answer options, ground truth, and the model's prediction. An example of this interactive tooltip is shown in Figure~\ref{fig:heatmap_cell_info}. This dynamic visualization transforms raw accuracy data into an intuitive diagnostic tool for analyzing model performance patterns and systematic weaknesses. The interactive visualization is publicly accessible at \url{https://angel-ntu.github.io/ESGenius}.

\section{Conclusion}
\label{sec:conclusion}
We presented \textbf{ESGenius}, the first comprehensive QA benchmark dedicated to probing LLMs on the breadth and depth of ESG and sustainability knowledge. It unifies two tightly integrated components: \textbf{ESGenius-QA}, a large-scale, expert-validated MCQs dataset, and the \textbf{ESGenius-Corpus}, a curated and extensible collection of authoritative ESG sources that anchors every question with verifiable evidence. Our two-stage evaluation protocol—\emph{Zero-Shot} followed by \emph{RAG}—shows that grounding responses in curated evidence boosts accuracy by 15–30 percentage points, with smaller models enhanced by RAG often surpassing much larger zero-shot models. This demonstrates that domain integration and transparent grounding are more decisive than sheer scale alone. By open-sourcing data, code, and a live and interactive leaderboard, \textbf{ESGenius} establishes a reproducible and continually updated yardstick for sustainability-aware AI. Our subsequent work will expand this mission to multi-modal ESG tasks \cite{zhang2025mmesgbench}, alongside expert-in-the-loop auditing, paving the way toward trustworthy ESG decision support.

\section{Limitations}
\label{sec:limitations}
While \textbf{ESGenius} aspires to provide a holistic evaluation of ESG understanding, several caveats must be acknowledged. 

\paragraph{Coverage trade-offs.} The benchmark currently includes only seven sources, which, while representing key frameworks and standards, cannot comprehensively encompass all ESG topics, industry-specific guidelines, regional regulations, or emerging sustainability frameworks. 

\paragraph{Expert dependency and scalability.} The creation of high-fidelity questions and the validation of answers are heavily dependent on scarce domain experts, making the process capacity-limited and prone to disciplinary bias. In future iterations, we plan to explore scalable strategies such as expert-in-the-loop active sampling or agreement-based adjudication to sustain long-term benchmark growth. 

\paragraph{Format constraints.} The reliance on standardized MCQs, although enabling large-scale scoring, compresses the nuanced reasoning and synthesis required in real-world ESG analysis and may therefore fall short of fully reflecting the complexity of sustainability decision-making. 

\paragraph{Standard volatility.} Frameworks such as ISSB guidance and IPCC assessments evolve rapidly, necessitating continual updates, corpus refreshes, and careful version control to maintain relevance. 

\paragraph{Language limitations.} The current focus on English-only corpora risks under-representing non-English ESG documents, local regulations, and region-specific sustainability practices. Expanding coverage to multilingual corpora is a key priority for improving global relevance. 

\paragraph{Exclusion of visual elements.} ESG reports often include key visual disclosures—charts, graphs, maps, and time-series plots. Our text-only design excludes multimodal reasoning. While text remains central, future versions will integrate visuals via OCR pipelines and vision–language models to better capture ESG reporting.

\paragraph{Metrics and evaluation scope.} Reliance on binary accuracy misses partial correctness and deeper ESG reasoning. We plan to add metrics such as explanation consistency, factual groundedness, and chain-of-thought plausibility. The MCQ format, while limited, offers scalability and comparability, as seen in MMLU and MMMU. We are exploring richer formats—open-ended answers, reasoning traces, and retrieval tasks—for better alignment with real-world ESG analysis.

\paragraph{Bias in LLM-generated questions.} LLMs are widely used in benchmarks for question generation and evaluation. Our pipeline adds expert verification to ensure factuality and relevance, but both models and humans can introduce bias. While we emphasize consistency and oversight, residual biases remain. We welcome validated alternatives to further reduce bias in future versions.

\paragraph{Limitations of the RAG implementation.} We use a simple RAG setup for reproducibility, deployment with smaller LLMs, and clear attribution of performance gains to the dataset and task design. Despite its simplicity, it already yields strong improvements. Future versions will integrate more advanced retrieval to enhance robustness and real-world alignment.

\paragraph{Human baseline performance.} Without human benchmarks, model scores lack context. We are developing a crowdsourcing interface for experts to provide evaluations, establishing baselines that will ground interpretation of results and guide future benchmarking.

\paragraph{Copyright limits.} Licensing restrictions constrain the inclusion of certain proprietary ESG standards, limiting exhaustive coverage and potentially omitting significant industry-specific frameworks.

\section{Ethical Considerations}
\label{sec:ethical_considerations}

\paragraph{Purpose and scope.}
\textbf{ESGenius} is released to advance \emph{trustworthy, evidence–grounded} research on LLMs Environmental, Social, and Governance (ESG) question answering.

\paragraph{Sources, copyright, and licensing.}
The \textbf{ESGenius-Corpus} aggregates \emph{authoritative but heterogeneous} materials (e.g., GRI, SASB/ISSB, IPCC, TCFD, CDP, SDGs in \S\ref{sec:esgenius_corpus}). Many of these works are copyrighted and/or distributed under specific use terms. We therefore: (i) include only publicly available documents or clearly reference official distribution pages; (ii) link every question to minimal excerpts sufficient for verification, avoiding wholesale redistribution; and (iii) ship documentation that records provenance and access routes for each source. Users of ESGenius must review and comply with the licenses and terms of the original sources; the benchmark is provided strictly for academic research and reproducibility.

\paragraph{LLM assistance and data leakage.}
LLMs were used only to \emph{propose} candidate MCQs; six domain experts double--validated all retained questions and three senior practitioners consulted the process (\S\ref{sec:qa_quality_control}). To reduce circularity, we (i) enforced strict evidence requirements per question, (ii) rejected or revised questions that were not unambiguously supported, and (iii) evaluated models---\emph{including the assisting model}---via API in an isolated session, with no access to construction traces. Each question is mapped to source passages (\S\ref{sec:corpus_question_mapping}), and a ``Not sure (Z)’’ option was added to discourage hallucination and reward calibrated abstention (\S\ref{sec:esgenius_qa_descriptions}). Beyond content generation, LLMs were occasionally employed as supportive tools for surface-level polishing, such as refining phrasing in writing or improving code readability and consistency. These uses were strictly non-substantive and did not influence the design, validation, or evaluation of ESGenius itself.

\paragraph{Annotation practice and labor.}
All validators were experienced in sustainability or NLP and gave informed consent to participate. The workflow emphasized factual verification, clarity, and distractor plausibility, with systematic rejection of ambiguous questions (\S\ref{sec:qa_quality_control}). We credit contributors as co–authors, and we encourage future adopters to follow fair labor practices when extending the dataset.

\paragraph{Fairness, representation, and epistemic balance.}
ESGenius currently emphasizes \emph{English–language, investor–oriented} frameworks and global standards (e.g., IFRS S1/S2, SASB) over national, local, Indigenous, or multilingual sources (\S\ref{sec:limitations}). This may reproduce an institutional or Global–North lens and under–represent region–specific norms or community priorities. We flag this as an ethical limitation and design the corpus to be \emph{modular and extensible} (\S\ref{sec:esgenius_corpus_extensibility}). Users should avoid over–generalizing conclusions beyond the covered sources and geographies.

\paragraph{Dual–use risks and misuse.}
Benchmarking can be gamed or misapplied. Examples include (i) training to the test (overfitting to known questions) and (ii) using ESG QA capability claims to enable persuasive \emph{but} ungrounded communication or greenwashing. To mitigate these risks we: (a) link every question to authoritative evidence and encourage evaluation settings that \emph{require} citations (RAG, \S\ref{sec:long_context_rag}); (b) provide versioning and clear changelogs; (c) retain the ``Z’’ option to penalize confident guessing; and (d) discourage using ESGenius scores as sole evidence of real–world compliance or as a checklist for circumventing disclosure obligations. We further recommend reporting retrieval quality and evidence coverage alongside accuracy.

\paragraph{Leaderboard and claims.}
Model rankings can invite outsized or decontextualized claims. We caution that (i) ESGenius is \emph{one} lens among many (\S\ref{sec:limitations}); (ii) zero–shot accuracy reflects parametric knowledge, whereas RAG performance reflects retrieval and grounding design (\S\ref{sec:main_results}); and (iii) differences of a few percentage points may not be practically meaningful without error and subtopic analyses (\S\ref{sec:subtopic_analysis}). Public results should disclose prompts, decoding settings, retrieval parameters, and version identifiers to support comparability and reproducibility.

\paragraph{Safety of model outputs.}
ESG advice or interpretations can carry legal, financial, or societal consequences. ESGenius is \emph{not} a substitute for professional guidance. We encourage deployers to (i) present grounded citations by default; (ii) surface uncertainty via calibrated abstention; and (iii) keep a expert-in-the-loop for consequential uses (e.g., risk management, regulatory filings). Our findings (e.g., stronger gains with RAG) underscore that \emph{evidence-linked} responses are ethically preferable to unsupported generations.

\paragraph{Environmental impact.}
Benchmarking LLMs consumes energy and incurs carbon costs. Our experiments used a single DGX A100 node (4 $\times$ 80GB) to limit overhead. We encourage (i) reporting hardware and token budgets, (ii) batching and caching retrieval, and (iii) preferring smaller models with strong RAG performance when feasible (\S\ref{sec:key_findings}).

\paragraph{Responsible release and governance.}
We release ESGenius with (i) documented data lineage and versioning; (ii) explicit usage guidelines discouraging compliance automation or deceptive communication; and (iii) maintenance plans for updates as standards evolve (\S\ref{sec:esgenius_corpus_extensibility}, \S\ref{sec:conclusion}). Community feedback on coverage gaps, bias, and error reports is welcome, and validated contributions will be incorporated through transparent review.

\section*{Acknowledgments}
This research is supported by the RIE2025 Industry Alignment Fund – Industry Collaboration Projects (IAF-ICP) (Award I2301E0026), administered by A*STAR, as well as supported by Alibaba Group and NTU Singapore through Alibaba-NTU Global e-Sustainability CorpLab (ANGEL).

% Use a standard bibliography style like acl_natbib
\bibliographystyle{acl_natbib} 
\bibliography{ref} % Assuming your bibliography file is named ref.bib

\appendix
\section*{Appendix}
\label{sec:appendix}

% --------------------------------------------------------

\section{ESGenius-QA}
\label{sec:esgenius_qa_appendix}

\subsection{Automated Preparation of Candidate MCQs}
\label{sec:mcq_generation_pipeline}

This appendix details the fully-automated preparation stage that converts raw knowledge sources into a pool of \emph{candidate} MCQs. (The complete algorithm is detailed in Algorithm~\ref{alg:mcq_generation_pipeline}.) 
These candidate questions subsequently undergo expert review, editing, and validation before potential inclusion in \textbf{ESGenius-QA}.
The goals of the preparation stage are twofold: (i) \emph{breadth}—to cover as many distinct passages as possible from the source corpus, and (ii) \emph{difficulty}—to generate questions that force LLMs to reason beyond surface-level facts.

\paragraph{Input corpus.} \label{sec:input_corpus}
Documents are first collected from authoritative sources in PDF format. Each PDF is then processed independently via the pipeline to enable precise tracking of corpus coverage. For encrypted PDFs, the pipeline attempts in-memory decryption where possible; if decryption fails, the file is skipped with a warning. Throughout processing, the system maintains detailed logs of extracted page counts and flags any instances of unsupported encryption formats.

\paragraph{Text extraction and chunking process.} \label{sec:text_extraction_process}
Using \texttt{PyPDF2}~\cite{pypdf2}, every page's text is cleaned (whitespace collapse and control-character stripping) and stored in a page map.  
A \emph{chunk selector} then samples passages that satisfy length constraints
\[
150 \le |{\rm chunk}| \le 4500 \text{ characters},
\]
optionally concatenating the next page when the combined length remains under the upper bound.  
Each chunk is tagged with its page span (e.g., \ ``12–13'') so that provenance is preserved.

\paragraph{Prompted multiple-choice question generation.} \label{sec:mcq_generation_process}
For every selected chunk the pipeline sends a carefully engineered prompt (\S\ref{app:prompt_generation}) to the \textbf{Qwen Max's DashScope} endpoint\footnote{\texttt{qwen-max-2025-01-25}, temperature 0, deterministic for reproducibility.} that requests \emph{exactly one} ``extremely challenging'' MCQ. Mandatory instructions enforce (i) deep reasoning across multiple sentences, (ii) near-miss distractors, (iii) answerability from the excerpt alone, and (iv) strict JSON output. A fixed seed (\texttt{42}) controls chunk sampling so that runs are reproducible.

\paragraph{Structural validation and metadata augmentation.} \label{sec:validation_and_augmentation}
The returned JSON is sanitised, parsed, and validated.  
Missing or malformed keys trigger rejection.  
The correct answer text is cross-checked with the declared key, and a fallback correction is applied if necessary.  
A universal \textbf{Z} option (``Not sure'') is appended to every question to enable abstention analysis.  
Each validated MCQ is wrapped in a metadata envelope containing
\texttt{\{id, source\_pdf, page\_range, difficulty, generation\_timestamp\}}.

\paragraph{Iterative difficulty refinement process.} \label{sec:refinement_process}
Because the pipeline aims for a target zero-shot accuracy of $\le 50\%$ on an external evaluator (we use Qwen Max as the evaluator), it performs up to 15 iterative refinement rounds to achieve this threshold:
\begin{enumerate}
  \item Test the current pool; label each MCQ with the model's chosen answer and correctness.
  \item Remove questions that the evaluator answers correctly.
  \item Replace the removed questions by sampling new chunks and repeating the generation step, maintaining the original pool size.
  \item Abort early if (a) accuracy falls below the target, (b) authentication errors reach the threshold (default 5), or (c) no new valid chunks remain.
\end{enumerate}

\paragraph{Consistency checking and question deduplication.} \label{sec:consistency_and_deduplication}
Beginning with round 2 the pipeline checks that an MCQ's evaluator result is stable across rounds; inconsistent questions are discarded.  
After refinement, a pairwise LLM-based similarity check removes near-duplicate questions that probe the same underlying concept.  
Only the first occurrence is retained.

\paragraph{Output artifacts generation and storage.} \label{sec:output_artifacts}
Two JSON files are written atomically after every processed PDF:  
\texttt{all\_candidates\_untested.json} stores \emph{every} syntactically valid question, whereas  
\texttt{final\_round\_tested.json} retains only the post-refinement pool together with evaluator metadata.  
Incremental saves ensure that partial progress survives crashes, and IDs are globally unique across sessions.

\paragraph{Transition to expert review process.} \label{sec:expert_review_process}
The resulting candidate pool is then passed to the expert review process detailed in \S\ref{sec:qa_quality_control}. This marks the transition from automated preparation to rigorous human validation and refinement of each question.

% --- Sole-page, full-width block in a two-column document ---
\onecolumn

\subsection*{Exact Prompt Template}
\label{app:prompt_generation}

\noindent
\begin{minipage}{\textwidth}
\begin{small}
\begin{verbatim}
Context: Based *only* on the following text excerpt from page(s) {page_range} of the document '{doc_name}':
'''
{text_excerpt}
'''
Task: Generate ONE **extremely challenging** multiple-choice question (MCQ) 
with a target difficulty level of '{difficulty}' specifically designed to be difficult for advanced LLMs like Qwen-Max. 
The question must rigorously test deep comprehension, critical reasoning, inference, and synthesis *strictly* based on 
the provided text excerpt. Avoid simple factual recall.

**Mandatory Instructions for Difficulty:**
1.  **Question Focus:** The question must target subtle details, implicit relationships, logical consequences, 
nuanced comparisons, or the underlying purpose/function of specific information within the text. It *must* require
synthesizing information from *multiple distinct sentences or points* within the excerpt.
2.  **Distractor Design (CRITICAL):** Create highly plausible but definitively incorrect distractors. These distractors 
should:
    * Be semantically very close to the correct answer or other concepts in the text.
    * Represent common misconceptions or logical fallacies related to the topic, *but be directly contradicted by 
    the provided text*.
    * Exploit potential ambiguities or require careful reading to identify subtle inaccuracies.
    * Avoid being trivially wrong or easily eliminated. All options (A, B, C, D) should seem plausible on a 
    superficial reading.
3.  **Avoid External Knowledge:** The question MUST be unambiguously answerable *solely* from the information 
contained within the provided text excerpt. The difficulty must come from the complexity of reasoning *within* the text, 
not from needing outside information.
4.  **Direct Questioning:** The text of the generated `question` itself must **not** contain phrases like 
"Based on the text excerpt", "According to the document...", "In the provided context...", etc. Frame the question 
directly about the content.
5.  **Advanced Strategies:** Employ techniques like identifying cause-effect relationships only implied, 
analyzing the necessity or sufficiency of conditions mentioned, or evaluating the scope/limitations 
of claims made in the text. Carefully consider using negation/exception questions ("Which of the following is NOT...") 
only if they significantly increase cognitive load beyond simple recall.

**Self-Correction Check (Internal Thought Process before outputting):** Does this question *genuinely* require 
multi-step reasoning or deep inference based *only* on the text? Are the distractors *truly* deceptive and require 
careful analysis of the excerpt to disprove? Is it significantly harder than a simple keyword search 
or basic comprehension question?

Output Format: Respond ONLY with a single, valid JSON object matching this exact structure:
{
  "question": "String: The extremely challenging question text (must not refer to the prompt or text source).",
  "options": { "A": "String: Highly plausible but incorrect 
  Option A.", "B": "String: Highly plausible but incorrect Option B.", "C": "String: The single, unambiguously correct 
  answer based *only* on the text.", "D": "String: Highly plausible but incorrect Option D." },
  "correct_answer_text": "String: The exact text of the correct answer option (must match one of the options exactly).",
  "correct_option_key": "String: The key ('A', 'B', or 'C', or 'D') corresponding to the correct answer."
}
**Important:** Ensure the `correct_option_key` points to the option that IS the correct answer. Randomize the position 
of the correct answer (A, B, C, or D). Do not include any introductory text, explanations, or markdown formatting like 
```json ... ``` before or after the JSON object. Output *only* the JSON.
\end{verbatim}
\end{small}
\end{minipage}

\twocolumn               % resume normal two-column layout
% =======================================================================
\begin{algorithm*}[t]
  \DontPrintSemicolon        % no end‐statements
  \SetAlgoLined              % ruled + vlined comes from the package options
  \SetKwInOut{Input}{Input}  % custom names
  \SetKwInOut{Output}{Output}
  \SetKwRepeat{Do}{do}{while} % for clarity (not used here, but handy)
  
  % ---------- hyper-parameters ----------
  \textbf{Hyper-parameters:}\\
  Chunk length $L\in[150,4500]$ chars;\
  \quad LLM = \texttt{qwen-max-2025-01-25}, $T{=}0$;\
  \quad seed $=42$;\
  \quad max rounds $R_{\max}=15$;\
  \quad target acc.~$\le50\%$;\
  \quad auth-error cap $E_{\max}=5$.\;
  
  % ---------- I/O ----------
  \Input{PDF folder \texttt{knowledge\_source}}
  \Output{\texttt{all\_candidates\_untested.json}, \texttt{final\_round\_tested.json}}
  
  % ---------- main loop ----------
  \ForEach(\tcp*[f]{independent handling}){PDF $d$ in folder}{
      decryptIfPossible$(d)$\;
      \lIf{unsupported encryption}{\textbf{continue}}
  
      pages $\leftarrow$ \textsc{ExtractPages}(d)\tcp*{cleaned text}
      chunks $\leftarrow$ \textsc{SelectChunks}(pages,$L$)\;
  
      \ForEach{chunk $c$ in chunks}{
          $\mathit{mcq} \leftarrow$ \textsc{PromptLLM}$(c)$\;
          \If{\textsc{Valid}$(\mathit{mcq})$}{
              append option $\mathbf{Z}$ (``Not~sure'')\;
              add metadata $\{\texttt{id},d,\texttt{page\_range},\texttt{difficulty},\texttt{timestamp}\}$\;
              save to \texttt{all\_candidates\_untested.json}\;
          }
      }
  
      % --- iterative difficulty refinement ---
      round $\leftarrow1$;\; acc $\leftarrow1$;\; errs $\leftarrow0$\;
      \While{$round\le R_{\max}$ \textbf{and} $acc>0.5$}{
          acc $\leftarrow$ \textsc{EvaluatePool}()\;
          remove correctly answered MCQs\;
          \lIf{$errs\ge E_{\max}$ \textbf{or} no new chunks}{\textbf{break}}
          replenish via \textsc{SelectChunks}$\rightarrow$\textsc{PromptLLM}\;
          $round\leftarrow round+1$\;
      }
  
      % --- post-processing ---
      drop unstable MCQs across rounds\;
      deduplicate with LLM similarity check (keep first)\;
      atomically write surviving pool $\rightarrow$ \texttt{final\_round\_tested.json}\;
  }
  
  \textbf{Expert review:} domain specialists verify facts, polish wording, rebalance difficulty, and accept/reject questions for release.\;
  
  \caption{Automated pipeline for generating candidate MCQs}
  \label{alg:mcq_generation_pipeline}
  \end{algorithm*}

\begin{table}[tb]
  \centering
  \begin{tabular}{@{}lr@{}}
    \toprule
    \textbf{Metric} & \textbf{Value} \\
    \midrule
    \multicolumn{2}{l}{\textit{Questions}} \\
    Entries/Tokens/Vocab & 1\,136 / 40\,983 / 2\,896 \\
    Mean (Median) & 36.08 (35) \\
    Q$_{1}$ -- Q$_{3}$ & 30 -- 40 \\
    Range & 18 -- 94 \\
    \midrule
    \multicolumn{2}{l}{\textit{Options}} \\
    Entries/Tokens/Vocab & 5\,680 / 88\,304 / 6\,810 \\
    Mean (Median) & 15.55 (17) \\
    Q$_{1}$ -- Q$_{3}$ & 12 -- 21 \\
    Range & 1 -- 54 \\
    \midrule
    \multicolumn{2}{l}{\textit{Source Text}} \\
    Entries/Tokens/Vocab & 1\,136 / 550\,200 / 18\,826 \\
    Mean (Median) & 484.33 (467) \\
    Q$_{1}$ -- Q$_{3}$ & 390 -- 586 \\
    Range & 33 -- 984 \\
    \bottomrule
  \end{tabular}
  \caption{Token--level profile of the \textbf{ESGenius-QA}. Source texts provide dense evidence (median 467 tokens) supporting higher-order reasoning.}
  \label{tab:token_stats}
\end{table}

\subsection{LLM's Raw Response Validation} \label{sec:llm_raw_response_validation}
Given an LLM's raw response, the routine first guards against \textsc{None} or empty inputs, then accepts ``Z'' directly as the special \emph{unsure} option. For other strings it proceeds in two stages:

\textbf{Direct acceptance:} If every character in the trimmed string is drawn from \texttt{\{A,B,C,D\}}, return the deduplicated, \emph{sorted} set (e.g., ``\texttt{DCB}'' $\rightarrow$ ``\texttt{BCD}'').

\textbf{Robust cleanup:} Otherwise the string is sanitized: (i)~replace all non-alphabetic symbols by spaces, (ii)~drop any word containing lowercase letters, (iii)~retain only words composed of \texttt{A--D,Z}. The result is deduplicated and sorted as above. If the cleaned string is empty, contains a lone ``Z'', or mixes ``Z'' with other letters, the function outputs \texttt{INVALID\_ANSWER\_MARKER}.

This design accepts the minimal valid alphabet while aggressively filtering free-form text, punctuation, and lower-case distractors that often appear in LLM generations, to guarantee that the response is a valid multiple-choice answer.

\subsection{ESGenius-QA Example Question Structures}
\label{sec:esgenius_qa_example_questions_appendix}

% ------------ Zero-shot MCQ Example (no preamble needed) ------------
\newenvironment{example}[1][\unskip]{\par\smallskip\noindent\textbf{#1}\par\smallskip}{\par\smallskip}
\subsubsection{Zero-Shot Evaluation Example}
\label{sec:mcq_zero_shot_example}
\begin{example}
\noindent \textbf{Question ID:} 1
  
\noindent \textbf{Question.} According to the \textit{IPCC AR6 Synthesis Report}, which statement would most likely \emph{increase the vulnerability of biodiversity and ecosystem services} while simultaneously \emph{reducing carbon uptake}, given the relationships described?
  
\noindent A. Implementing targeted management strategies for ocean ecosystems without addressing global warming.

\noindent B. Failing to rebuild overexploited fisheries while achieving significant reductions in fossil-fuel dependency.

\noindent C. Limiting global warming but neglecting land-restoration efforts and cooperative decision-making with Indigenous Peoples.

\noindent D. Prioritizing disaster-risk management and early-warning systems without integrating ecosystem considerations into climate policies.

\noindent Z. Not sure
  
\noindent \textbf{Correct Answer:} A
  
\end{example}
  % --------------------------------------------------------------------

% ----------------------------------------------------------------
% *** RAG evaluation example with full metadata and retrieved text ***
% ----------------------------------------------------------------
\subsubsection{RAG Evaluation Example}
\label{sec:mcq_rag_example}
\begin{example}
\noindent \textbf{Question ID:} 580
  
\noindent \textbf{Reference.} Page~213, \texttt{SUS Report.pdf}
  
\noindent \textbf{Question.} Which of the following conclusions about Fiji’s progress toward sustainable development can be most reasonably inferred from the data trends and gaps presented in the \textit{Sustainable Development Report 2024}?
  
\noindent A. Fiji has achieved near-universal access to clean water but faces significant challenges in reducing urban slum populations.\\
\noindent B. Fiji’s environmental sustainability efforts are hindered primarily by high deforestation rates and low biodiversity protection.\\
\noindent C. Fiji excels in reducing income inequality, as evidenced by a low Gini coefficient and minimal reliance on imports for nitrogen emissions.\\
\noindent D. Fiji demonstrates strong performance in internet accessibility but shows untracked research and development expenditure.\\
\noindent Z. Not sure
  
\noindent\textbf{Correct Answer:} D

\noindent\textbf{Retrieved Source Text.}
\small
Performance by Indicator5. Country Profiles \textit{Sustainable Development Report 2024 – The SDGs and the UN Summit of the Future} 201\\*
* Imputed data point,\quad ** Not applicable\quad NA = Data not available\\[4pt]
\textbf{FIJI}\\
\textbf{SDG 9 – Industry, Innovation and Infrastructure}\\
Rural population with access to all-season roads (\%) 96.7 (2024) ••\\
Population using the internet (\%) 85.2 (2022) •A\\
Mobile broadband subscriptions (per 100 population) 76.1 (2021) •A\\
Logistics Performance Index: Infrastructure score (worst 1–5 best) 2.2 (2023) •G\\
The Times Higher Education Universities Ranking: Average score of top 3 universities (worst 0–100 best) 30.5 (2024) ••\\
Articles published in academic journals (per 1 000 population) 0.6 (2022) •A\\
Expenditure on research and development (\% of GDP) \textbf{NA NA ••}\\[4pt]
\textbf{SDG 10 – Reduced Inequalities}\\
Gini coefficient 30.7 (2019) ••\\
Palma ratio 1.1 (2019) ••\\[4pt]
\textbf{SDG 11 – Sustainable Cities and Communities}\\
Proportion of urban population living in slums (\%) 9.4 (2020) •D\\
Annual mean concentration of $PM_{2.5}$ (µg/m³) 7.4 (2022) •D\\
Access to improved water source, piped (\% of urban population) 98.4 (2022) •A\\
Population with convenient access to public transport in cities (\%) 19.2 (2020) ••\\[4pt]
\textbf{SDG 12 – Responsible Consumption and Production}\\
Municipal solid waste (kg/capita/day) 0.6 (2011) ••\\
Electronic waste (kg/capita) 6.1 (2019) ••\\
Production-based air pollution (DALYs per 1 000 population) NA NA ••\\
Air pollution associated with imports (DALYs per 1 000 population) NA NA ••\\
Production-based nitrogen emissions (kg/capita) NA NA ••\\
Nitrogen emissions associated with imports (kg/capita) NA NA ••\\
Exports of plastic waste (kg/capita) 0.6 (2022) •A\\[4pt]
\textbf{SDG 13 – Climate Action}\\
$CO_2$ emissions from fossil-fuel combustion and cement production (t $CO_2$/capita) 1.2 (2022) •A\\
GHG emissions embodied in imports (t $CO_2$/capita) NA NA ••\\
$CO_2$ emissions embodied in fossil-fuel exports (kg/capita) 0.0 (2022) ••\\[4pt]
\textbf{SDG 14 – Life Below Water}\\
Mean area that is protected in marine sites important to biodiversity (\%) 16.5 (2023) •D\\
Ocean Health Index: Clean Waters score (worst 0–100 best) 74.1 (2023) •D\\
Fish caught from overexploited or collapsed stocks (\% of total catch) 9.0 (2018) •A\\
Fish caught by trawling or dredging (\%) 0.0 (2019) ••\\
Fish caught that are then discarded (\%) 7.3 (2019) •D\\
Marine biodiversity threats embodied in imports (per million population) 0.3 (2018) ••\\[4pt]
\textbf{SDG 15 – Life on Land}\\
Mean area that is protected in terrestrial sites important to biodiversity (\%) 11.2 (2023) •D\\
Mean area that is protected in freshwater sites important to biodiversity (\%) 0.1 (2023) •D\\
Red List Index of species survival (worst 0–1 best) 0.69 (2024) •G\\
Permanent deforestation (\% of forest area, 3-year average) 0.1 (2022) •A\\
Imported deforestation (m²/capita) NA NA ••\\[4pt]
\textbf{SDG 16 – Peace, Justice and Strong Institutions}\\
Homicides (per 100 000 population) 2.2 (2020) •D\\
Crime is effectively controlled (worst 0–1 best) NA NA ••\\
Unsentenced detainees (\% of prison population) 19.9 (2021) •A\\
Birth registrations with civil authority (\% of children under 5) 86.6 (2021) ••\\
Corruption Perceptions Index (worst 0–1 best) 52.0 (2023) ••\\
Children involved in child labor (\%) 16.7 (2021) ••\\
Exports of major conventional weapons (TIV constant million USD per 100 000 population)\,* 0.0 (2023) ••\\
Press Freedom Index (worst 0–1 best) 71.2 (2024) •A\\
Access to and affordability of justice (worst 0–1 best) NA NA ••\\
Timeliness of administrative proceedings (worst 0–1 best) NA NA ••\\
Expropriations are lawful and adequately compensated (worst 0–1 best) NA NA ••\\[4pt]
\textbf{SDG 17 – Partnerships for the Goals}\\
Government spending on health and education (\% of GDP) 9.3 (2021) •A\\
Government revenue excluding grants (\% of GDP) 19.0 (2021) •G\\
Corporate Tax Haven score (best 0–100 worst)\,* 0 (2021) ••\\
Statistical Performance Index (worst 0–100 best) 63.2 (2022) •S\\
Index of countries' support to UN-based multilateralism (worst 0–100 best) 88.3 (2023) ••\\[4pt]
\textbf{SDG 1 – No Poverty}\\
Poverty headcount ratio at \$2.15/day (2017 PPP, \%) 1.6 (2024) •D\\
Poverty headcount ratio at \$3.65/day (2017 PPP, \%) 7.3 (2024) •D\\[4pt]
\textbf{SDG 2 – Zero Hunger}\\
Prevalence of undernourishment (\%) 6.6 (2021) •A\\
Prevalence of stunting in children under 5 years of age (\%) 7.2 (2021) ••\\
Prevalence of wasting in children under 5 years of age (\%) 4.6 (2021) ••\\
Prevalence of obesity, BMI \(\geq\) 30 (\% of adult population) 33.8 (2022) •G\\
Human Trophic Level (best 2–3 worst) 2.2 (2021) •D\\
Cereal yield (tonnes per hectare of harvested land) 4.1 (202\textellipsis)
\normalsize
\end{example}

\section{Prompt Template}
\label{sec:prompt_template}
For MCQs (4 options + 1 Not sure with 1 answer), we use the following prompt templates:

\subsection{Zero-Shot Prompt}
\label{sec:zero_shot_prompt}

\noindent You are an expert in ESG (Environmental, Social, Governance) and Sustainability related topics. Answer the question with a single letter based on authoritative knowledge. Each option content is case-sensitive.

\noindent Question: [Question text]

\noindent Options:\\[2pt]
A: [Option A text]\\[2pt]
B: [Option B text]\\[2pt]
C: [Option C text]\\[2pt]
D: [Option D text]\\[2pt]
Z: Not sure\\[2pt]

\noindent Answer: <Model's response goes here>

\subsection{RAG Prompt}
\label{sec:rag_prompt}

\noindent Context: [source text]

\noindent You are an expert in ESG (Environmental, Social, Governance) and Sustainability related topics. Answer the question with a single letter based on authoritative knowledge. Each option content is case-sensitive.

\noindent Question: [Question text]

\noindent Options:\\[2pt]
A: [Option A text]\\[2pt]
B: [Option B text]\\[2pt]
C: [Option C text]\\[2pt]
D: [Option D text]\\[2pt]
Z: Not sure\\[2pt]

\noindent Answer: <Model's response goes here>

\section{Main Experimental Results Table}
\label{sec:main_experimental_results}
\begin{table*}[tb]
  \centering
  \resizebox{\textwidth}{!}{%
  \begin{tabular}{l l l r l r r | r r r}
  \hline
  \textbf{Type} & \textbf{Family} & \textbf{Model} & \textbf{Size} & \textbf{S.G.} & \textbf{I-T} & \textbf{Rea} & \textbf{Zero-Shot} & \textbf{RAG} & \textbf{Improvement} \\
  \hline
  \multirow{43}{*}{Open Source} & \multirow{6}{*}{DeepSeek}
                     & DeepSeek-R1-Distill-Qwen  & 1.5B  & M  & No  & Yes & 0.3134 & 0.4305 & 37.36\% \\
                     &                      & DeepSeek-R1-Distill-Qwen  & 7B    & L  & No  & Yes & 0.5018 & 0.6505 & 29.63\% \\
                     &                      & DeepSeek-R1-Distill-Qwen  & 14B   & L  & No  & Yes & 0.6382 & 0.8046 & 26.07\% \\
                     &                      & DeepSeek-R1-Distill-Qwen  & 32B   & XL & No  & Yes & 0.6312 & 0.8143 & 29.01\% \\
                     &                      & DeepSeek-R1-Distill-Llama & 8B    & L  & No  & Yes & 0.5502 & 0.6250 & 13.60\% \\
                     &                      & DeepSeek-R1-Distill-Llama & 70B   & XL & No  & Yes & 0.6505 & 0.8170 & 25.60\% \\
  \cline{2-10}
                     & \multirow{8}{*}{Google Gemma}
                     & Gemma-3 \cite{team2025gemma}      & 1B    & M & No  & No  & 0.2421 & 0.2526 & 4.33\% \\
                     &                      & Gemma-3                   & 1B    & M & Yes & No  & 0.5000 & 0.5977 & 19.54\% \\
                     &                      & Gemma-3                   & 4B    & M & No  & No  & 0.6083 & 0.6860 & 12.77\% \\
                     &                      & Gemma-3                   & 4B    & M & Yes & No  & 0.6144 & 0.7518 & 22.36\% \\
                     &                      & Gemma-3                   & 12B   & L & No  & No  & 0.6514 & 0.6857 & 5.26\% \\
                     &                      & Gemma-3                   & 12B   & L & Yes & No  & 0.6514 & 0.8380 & 28.64\% \\
                     &                      & Gemma-3                   & 27B   & L & No  & No  & 0.2165 & 0.5229 & 141.57\% \\
                     &                      & Gemma-3                   & 27B   & L & Yes & No  & 0.6356 & 0.8336 & 31.15\% \\
  \cline{2-10}
                     & \multirow{8}{*}{Meta Llama}
                     & Meta-Llama-3 \cite{grattafiori2024llama} & 8B & L & No  & No  & 0.6461 & 0.7324 & 13.36\% \\
                     &                      & Llama-3.1                 & 8B    & L & No  & No  & 0.6699 & 0.7650 & 14.20\% \\
                     &                      & Llama-3.1                 & 8B    & L & Yes & No  & 0.6382 & 0.7993 & 25.24\% \\
                     &                      & Llama-3.2                 & 1B    & M & No  & No  & 0.3609 & 0.3680 & 2.00\% \\
                     &                      & Llama-3.2                 & 1B    & M & Yes & No  & 0.5986 & 0.6452 & 7.79\% \\
                     &                      & Llama-3.2                 & 3B    & M & No  & No  & 0.6074 & 0.6831 & 12.48\% \\
                     &                      & Llama-3.2                 & 3B    & M & Yes & No  & 0.5968 & 0.7218 & 20.95\% \\
                     &                      & Llama-3.3                 & 70B   & XL & Yes & No  & 0.6576 & 0.7887 & 20.00\% \\
  \cline{2-10}
                     & \multirow{21}{*}{Alibaba Qwen}
                     & Qwen2.5 \cite{yang2024qwen2}   & 0.5B  & S & No  & No  & 0.5458 & 0.5396 & -1.14\% \\
                     &                      & Qwen2.5                   & 0.5B  & S & Yes & No  & 0.5775 & 0.5343 & -7.48\% \\
                     &                      & Qwen2.5                   & 1.5B  & M & No  & No  & 0.6391 & 0.6928 &  8.40\% \\
                     &                      & Qwen2.5                   & 1.5B  & M & Yes & No  & 0.5484 & 0.6972 & 27.13\% \\
                     &                      & Qwen2.5                   & 3B    & M & No  & No  & 0.5889 & 0.7632 & 29.60\% \\
                     &                      & Qwen2.5                   & 3B    & M & Yes & No  & 0.5871 & 0.5211 & -11.24\% \\
                     &                      & Qwen2.5                   & 7B    & L & No  & No  & 0.6496 & 0.8055 & 23.99\% \\
                     &                      & Qwen2.5                   & 7B    & L & Yes & No  & 0.6276 & 0.7967 & 27.27\% \\
                     &                      & Qwen2.5                   & 14B   & L & No  & No  & 0.6734 & 0.8231 & 22.22\% \\
                     &                      & Qwen2.5                   & 14B   & L & Yes & No  & 0.6197 & 0.8231 & 32.83\% \\
                     &                      & Qwen2.5                   & 32B   & XL & No  & No  & 0.6593 & 0.8081 & 22.55\% \\
                     &                      & Qwen2.5                   & 32B   & XL & Yes & No  & 0.6039 & 0.8247 & 36.57\% \\
                     &                      & Qwen2.5                   & 72B   & XL & No  & No  & 0.6188 & 0.7201 & 16.39\% \\
                     &                      & Qwen2.5                   & 72B   & XL & Yes & No  & 0.6347 & 0.8257 & 29.78\% \\
                     &                      & Qwen2.5-1M \cite{yang2025qwen2} & 7B    & L & Yes & No  & 0.6206 & 0.8063 & 29.92\% \\
                     &                      & Qwen2.5-1M                & 14B   & L & Yes & No  & 0.6268 & 0.8222 & 28.01\% \\
                     &                      & QwQ \cite{qwq32b}              & 32B   & XL & No  & Yes & 0.3900 & 0.7614 & 95.23\% \\
                     &                      & Qwen3 \cite{yang2025qwen3}            & 0.6B   & S & No  & No  & 0.2896 & 0.0942 & -67.47\% \\
                     &                      & Qwen3             & 1.7B   & M & No  & No  & 0.5836 & 0.6937 & 18.87\% \\
                     &                      & Qwen3             & 4B   & M & No  & No  & 0.6188 & 0.7905 & 27.75\% \\
                     &                      & Qwen3             & 8B   & L & No  & No  & 0.6021 & 0.6708 & 11.41\% \\
  \hline
  \multirow{7}{*}{Proprietary API} & \multirow{2}{*}{DeepSeek}
                     & DeepSeek-R1 \cite{guo2025deepseek} & 671B  & XXL & No  & Yes & 0.6629 & - & - \\
                     &                      & DeepSeek-V3 \cite{liu2024deepseek} & 671B  & XXL & No  & No  & 0.6532 & - & - \\
  \cline{2-10}
                     & Alibaba Qwen
                     & Qwen2.5-Max (qwen-max-2025-01-25)               & A22B†(MoE 325B)  & XL & No  & Yes & 0.6444 & - &  - \\
  \cline{2-10}
                     & \multirow{4}{*}{OpenAI GPT}
                     & GPT-4o-mini               & 8B†    & L & No  & No  & 0.6268 & - &  - \\
                     &                      & GPT-4o \cite{hurst2024gpt}     & 200B†  & XXL & No & No & 0.6364 & - & - \\
                     &                      & o4-mini              & 3B†    & M & No  & Yes & 0.6945 & - &  - \\
                     &                      & o3           & 300B†    & M & No  & Yes & 0.7254 & - & - \\
  \hline
  \end{tabular}%
  }
  \caption{Comprehensive ESGenius results showing LLM performance under Zero-Shot and RAG settings. S.G. denotes Size Group (S: Small ($\leq$1B), M: Medium (1–7B), L: Large (7–30B), XL: Extra Large (30–100B), XXL: Extra Extra Large ($>$100B)); I-T: Instruction Tuning; Rea: Reasoning Focus. Improvement (\%): $\displaystyle \frac{\text{RAG} - \text{Zero-Shot}}{\text{Zero-Shot}} \times 100$. ``--'' indicates scores currently unavailable due to technical constraints. † indicates industry size estimates. A ranking bar chart of zero-shot performance is shown in Figure~\ref{fig:esgenius_ranking}.}
  \label{tab:benchmark_results}
\end{table*}

\begin{figure*}[h!]
  \centering
  \includegraphics[width=\textwidth]{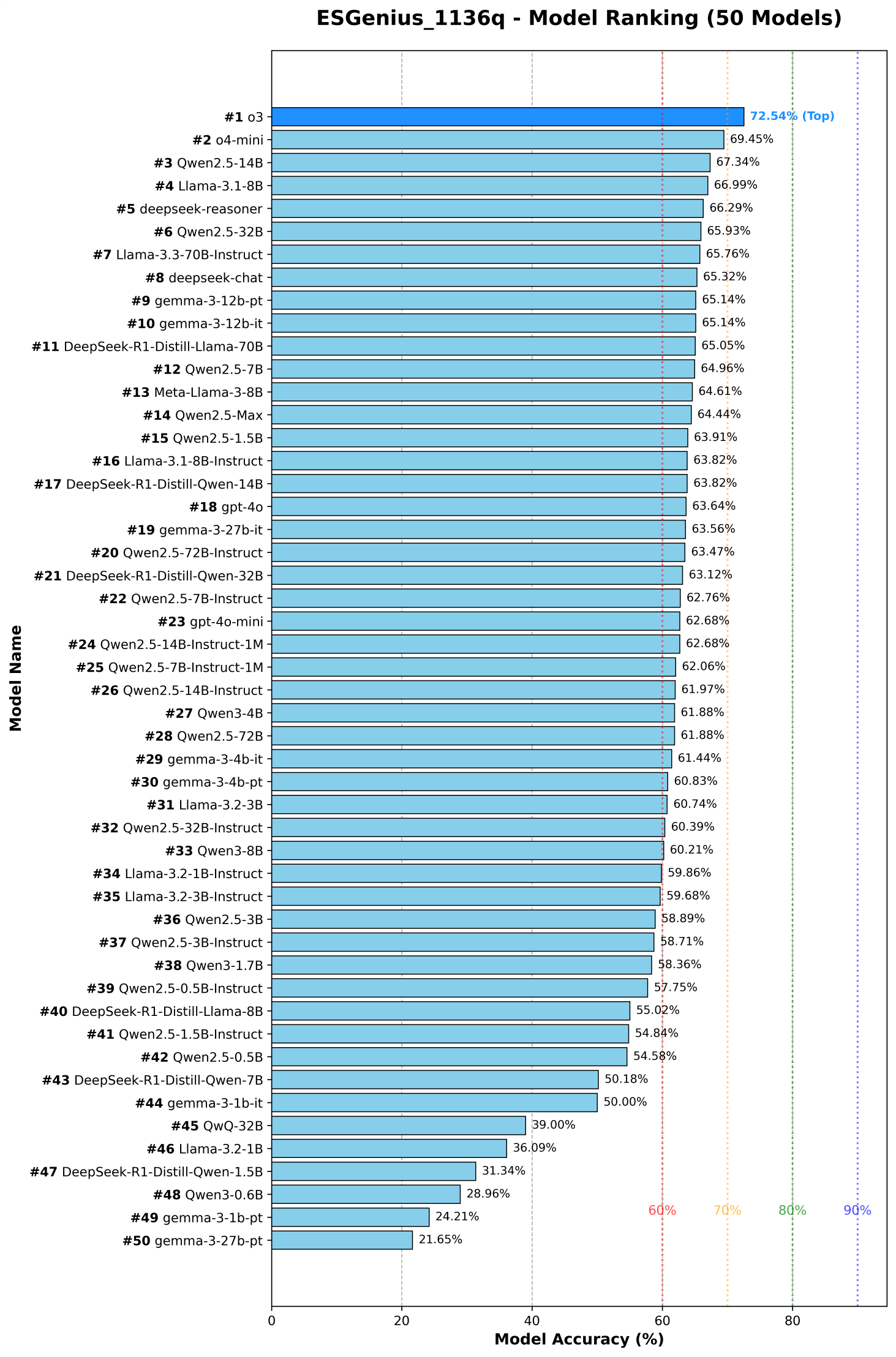}
  \caption{Zero-Shot model ranking on ESGenius (1136 questions). The chart shows model accuracies across 50 LLMs, with \texttt{o3} achieving the highest score of 72.54\%.}
  \label{fig:esgenius_ranking}
\end{figure*}

\subsection{3 Examples with Very Low Accuracy Across All Models}
\label{ex:mcq_low_accuracy_examples}

\vspace{0.5em} % Add a little space before the first question block

\noindent % Prevent indentation for the first line
\textbf{Question ID:} 432 \\
\textbf{Question.} Which statement accurately reflects the relationship between renewable energy adoption and regional disparities as indicated in \textit{The Sustainable Development Goals Report 2024}?

\noindent Options:

\noindent A: Developing countries are projected to surpass developed countries in per capita renewable electricity capacity within the next decade based on current growth rates.

\noindent B: The disparity in renewable energy capacity between least developed countries (LDCs) and developing countries is expected to close within 15 years if LDCs maintain their current growth trajectory.

\noindent C: The installed renewable electricity capacity in least developed countries (LDCs) grew at a faster compound annual growth rate than in developed countries over the past seven years.

\noindent D: Landlocked developing States have achieved a higher per capita renewable electricity capacity than small island developing countries but still lag significantly behind developing countries overall.

\noindent Z: Not sure

\noindent
\textbf{Correct Answer:} D \\
\textbf{Accuracy:} 0\%

\vspace{0.5em}
\hrule
\vspace{1em}

\noindent
\textbf{Question ID:} 635 \\
\textbf{Question.} Which factor is most directly responsible for the limited adoption of ESCO business models despite their potential to mitigate financial risks and provide expertise in energy efficiency projects, according to the \textit{Climate Change 2022: Mitigation of Climate Change. Working Group III Contribution to the Sixth Assessment Report of the Intergovernmental Panel on Climate Change}?

\noindent Options:

\noindent A: The absence of stringent regulatory frameworks governing energy codes.

\noindent B: The insufficient involvement of the public sector in transportation infrastructure projects.

\noindent C: The lack of awareness among financial institutions about energy efficiency benefits.

\noindent D: The prevalence of asymmetric information and split incentives within firms.

\noindent Z: Not sure

\noindent
\textbf{Correct Answer:} C \\
\textbf{Accuracy:} 0\%

\vspace{0.5em}
\hrule
\vspace{1em}

\noindent
\textbf{Question ID:} 1006 \\
\textbf{Question.} Which of the following best describes a necessary condition for an injury or illness to be classified as a recordable incident under the entity's disclosure requirements in the \textit{Chemicals – Sustainability Accounting Standard}?

\noindent Options:

\noindent A: The injury or illness must result in at least one day away from work or require medical treatment beyond first aid.

\noindent B: The injury or illness must be diagnosed by a licensed healthcare professional, regardless of its impact on the employee's work status.

\noindent C: The injury or illness must occur within the establishment but does not need to involve exposure to harmful substances or heavy machinery.

\noindent D: The injury or illness must lead to restricted work, job transfer, or loss of consciousness, even if it is not diagnosed by a physician.

\noindent Z: Not sure

\noindent
\textbf{Correct Answer:} B \\
\textbf{Accuracy:} 0\%

\section{ESGenius-Corpus}
\label{sec:esgenius_corpus_details}

\begin{figure*}[tb]
  \centering
  \includegraphics[width=1\textwidth]{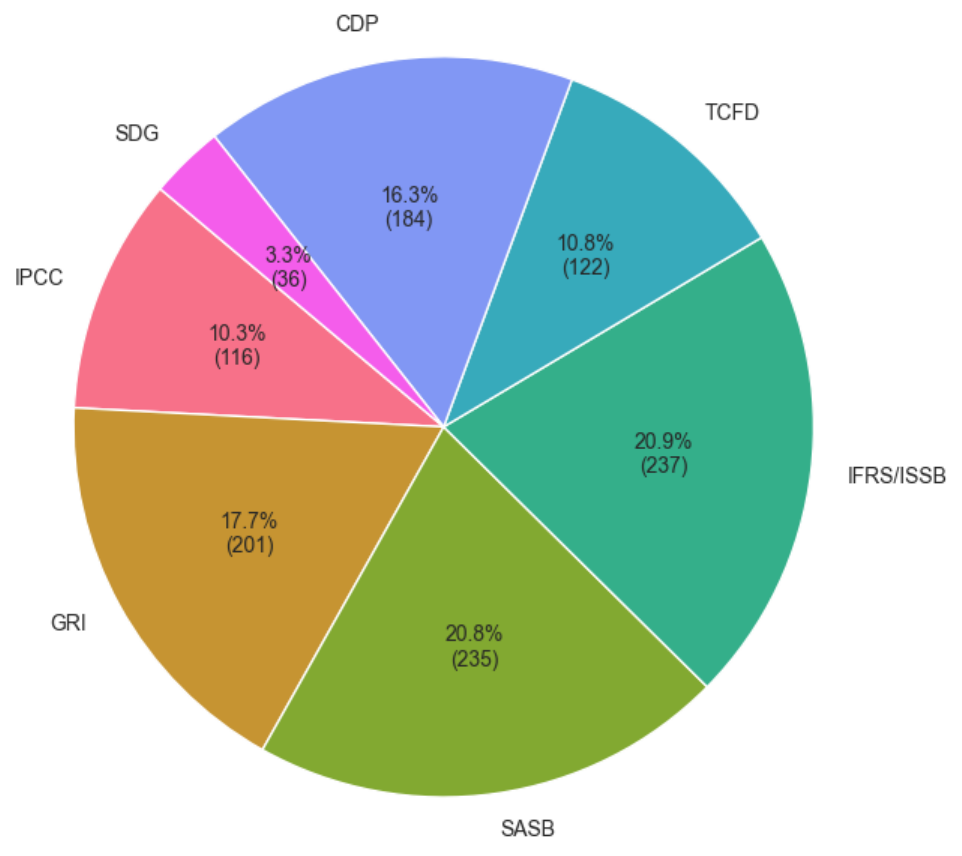}
  \caption{Relative \textbf{question distribution} (benchmark MCQs) derived from each source family. In contrast to the page distribution, questions are more evenly spread: the largest contributors are IFRS/ISSB (\textbf{20.9\%}, 237 Qs) and SASB (\textbf{20.8\%}, 235 Qs), followed by GRI (\textbf{17.7\%}, 201 Qs) and CDP (\textbf{16.3\%}, 184 Qs).  TCFD (\textbf{10.8\%}, 122 Qs) and IPCC (\textbf{10.3\%}, 116 Qs) provide focused climate-risk and climate-science coverage, while the SDG set supplies a compact but essential sustainability anchor (\textbf{3.3\%}, 36 Qs).}

  \label{fig:num_questions_pie}
\end{figure*}

\begin{figure*}[tb]
  \centering
  \includegraphics[width=1\textwidth]{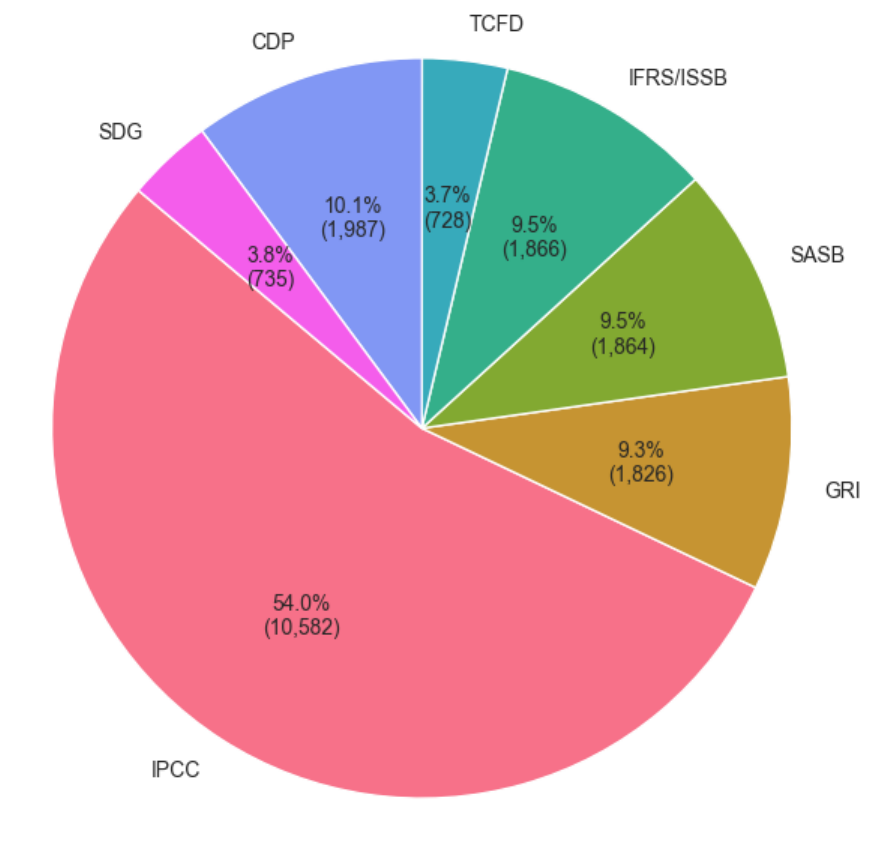}
  \caption{Relative \textbf{page‐count distribution} of the \textbf{ESGenius-Corpus} across its 7 source families. The Intergovernmental Panel on Climate Change (IPCC) alone accounts for a majority of the material—\textbf{54\%}, or \textbf{10,582 pages}—reflecting the encyclopaedic scope of its assessment reports. Standards and disclosure frameworks contribute smaller but still substantive shares: CDP (\textbf{10.1\%}, 1,987 pp.), IFRS/ISSB (\textbf{9.5\%}, 1,866 pp.), SASB (\textbf{9.5\%}, 1,864 pp.), and GRI (\textbf{9.3\%}, 1,826 pp.).  Guidance‐oriented sources such as TCFD (\textbf{3.7\%}, 728 pp.) and the UN SDGs corpus (\textbf{3.8\%}, 735 pp.) round out the collection.}
  \label{fig:pages_pie}
\end{figure*}

The corpus integrates authoritative frameworks, corporate questionnaires, and scientific assessments that comprehensively cover environmental, social, and governance (ESG) considerations. A detailed quantitative overview of the collection is presented across multiple tables: Table~\ref{tab:ipcc_table} summarizes IPCC climate-science assessments, Table~\ref{tab:gri_table} covers GRI Standards, Tables~\ref{tab:sasb_part1_table}–\ref{tab:sasb_part2_table} detail SASB's industry-specific disclosures, Tables~\ref{tab:ifrsissb_part1_table}–\ref{tab:ifrsissb_part2_table} outline IFRS/ISSB standards and guidance, Table~\ref{tab:tcfd_table} catalogs TCFD materials, Table~\ref{tab:cdp_table} documents CDP questionnaires, and Table~\ref{tab:sdg_table} captures UN Sustainable Development Goals (SDGs) content. The distribution of pages and questions across these sources is visualized in Figures~\ref{fig:num_questions_pie} and \ref{fig:pages_pie}, respectively. From this comprehensive dataset, several key insights emerge:

\begin{enumerate}
\item \textbf{Breadth versus depth.}
Altogether, these seven sources comprise \textbf{231 distinct documents} spanning approximately \textbf{19,600 pages}. While bodies such as SASB and IFRS/ISSB each contribute dozens of relatively concise files, the Intergovernmental Panel on Climate Change (IPCC) anchors the corpus with just seven reports that total over 10,000 pages.

\item \textbf{Standards-driven core.}
Internationally recognized ESG standards and disclosure frameworks—GRI, SASB, IFRS/ISSB, TCFD, and CDP—form the backbone of \textbf{ESGenius}. They reflect the most widely adopted practices for sustainability reporting, management, and climate-risk disclosure, offering a robust foundation for both quantitative and qualitative benchmarking.

\item \textbf{Disclosure and reporting frameworks.}
Market-facing reporting initiatives (SASB, GRI, TCFD, CDP) collectively emphasize implementation guides and sector- or topic-specific questionnaires, resulting in many individual questions but fewer pages per document. This modular structure facilitates domain-specific data collection and comparability across diverse industries.

\item \textbf{Science-heavy climate assessments.}
IPCC assessment reports present the opposite pattern: fewer individual documents but extremely high page counts. This science-heavy text ensures deep coverage of climate-change fundamentals, impacts, and mitigation pathways—an essential knowledge base informing the standards and regulations in the broader ESG ecosystem.

\item \textbf{Sustainable development anchor.}
The UN’s SDGs underpin cross-sector and cross-country sustainability objectives. Although comparatively compact in page count, these seminal UN publications situate corporate ESG strategies within the global 2030 Agenda, ensuring broader alignment with international development priorities.

\item \textbf{Imbalanced density highlights practical challenges.}
Marked disparities between the distribution of documents and the distribution of pages underscore the varied scope of ESG sources: some (e.g., \ IPCC) are exhaustive scientific compendiums, while others (e.g., \ GRI, SASB, IFRS) comprise slimmer but more numerous reference standards. For researchers and practitioners alike, tasks ranging from large-scale summarization to specialized technical queries must navigate this imbalance of question count versus depth.
\end{enumerate}

Taken together, these characteristics demonstrate that \textbf{ESGenius} provides both the \emph{breadth} (multiple standards, guidance, and scientific anchors) and the \emph{depth} (tens of thousands of pages) required for evaluating advanced language models on ESG-focused reasoning, retrieval, and generation tasks. Future expansions will incorporate emerging disclosure rules and further national or sectoral guidelines, maintaining the corpus’s comprehensive coverage over time.

\begin{table*}[tb]
\centering
  \scriptsize
  \setlength\tabcolsep{3pt}
  \begin{tabular}{p{9cm} c c c c}
      \toprule
      \textbf{Original Document Title} &
      \textbf{Year} & \textbf{Size} &
      \textbf{Pages} & \textbf{No. Qs} \\ 
      \midrule
      \href{https://www.ipcc.ch/report/ar6/syr/downloads/report/IPCC_AR6_SYR_FullVolume.pdf}{\emph{Climate Change 2023 — Synthesis Report}} \cite{IPCC2023SYR} & 2023 & 4.9 MB & 186 & 18 \\[2pt]
      \href{https://www.ipcc.ch/report/ar6/wg3/downloads/report/IPCC_AR6_WGIII_FullReport.pdf}{\emph{Climate Change 2022: Mitigation of Climate Change. Working Group III Contribution to the Sixth Assessment Report of the Intergovernmental Panel on Climate Change}} \cite{IPCC2022_WG3} & 2022 & 74.2 MB & 2\,042 & 19 \\[2pt]
      \href{https://report.ipcc.ch/ar6/wg2/IPCC_AR6_WGII_FullReport.pdf}{\emph{Climate Change 2022: Impacts, Adaptation and Vulnerability. Working Group II Contribution to the IPCC Sixth Assessment Report} \cite{IPCC2022_WG2}} & 2022 & 378 MB & 3\,675 & 14 \\[2pt]
      \href{https://www.ipcc.ch/report/ar6/wg1/downloads/report/IPCC_AR6_WGI_FullReport_small.pdf}{\emph{Climate Change 2021: The Physical Science Basis. Working Group I Contribution to the IPCC Sixth Assessment Report} \cite{IPCC2021_WG1}} & 2021 & 275 MB & 2\,409 & 14 \\[2pt]
      \href{https://www.ipcc.ch/site/assets/uploads/2019/11/SRCCL-Full-Report-Compiled-191128.pdf}{\emph{Climate Change and Land: An IPCC Special Report on climate change, desertification, land degradation, sustainable land management, food security, and greenhouse gas fluxes in terrestrial ecosystems}} \cite{IPCC2019SRCCL} & 2019 & 28 MB & 874 & 16 \\[2pt]
      \href{https://www.ipcc.ch/site/assets/uploads/sites/3/2019/12/SROCC_FullReport_FINAL.pdf}{\emph{The Ocean and Cryosphere in a Changing Climate: A Special Report of the Intergovernmental Panel on Climate Change}} \cite{IPCC2019_SROCC}& 2019 & 59.4 MB & 765 & 17 \\[2pt]
      \href{https://www.ipcc.ch/site/assets/uploads/2022/06/SR15_Full_Report_HR.pdf}{\emph{Global Warming of 1.5 °C: An IPCC Special Report on the impacts of global warming of 1.5°C above pre-industrial levels and related global greenhouse gas emission pathways, in the context of strengthening the global response to the threat of climate change, sustainable development, and efforts to eradicate poverty}} \cite{IPCC2018_SR15} & 2018 & 65 MB & 631 & 19 \\[2pt]
      \midrule
      Total & - & 884.5MB & 10\,582 & 117 \\[2pt]
      \bottomrule
  \end{tabular}
  \caption{Comprehensive metadata for the \textbf{seven IPCC reports} curated in the \textbf{ESGenius}. This collection represents the complete Sixth Assessment Report (AR6) cycle and key Special Reports from 2018-2023. The AR6 materials include the 2023 Synthesis Report and three Working Group contributions covering physical science (WG1), impacts \& adaptation (WG2), and mitigation (WG3). Three thematic Special Reports address land use (SRCCL), oceans \& ice (SROCC), and 1.5°C warming pathways (SR15). Totaling \textbf{10,582 pages} and \textbf{117 evaluation questions}, these authoritative climate science assessments form a crucial knowledge foundation for ESG analysis. Document sizes range from 4.9MB to 378MB. The collection provides comprehensive coverage of climate science, impacts, and policy responses that inform modern ESG frameworks like TCFD, CSRD, and ISSB standards. All documents are sourced directly from IPCC (\url{https://www.ipcc.ch}) and represent peer-reviewed, UN-mandated scientific assessments.}
  \label{tab:ipcc_table}
\end{table*}
\clearpage

\begin{table*}[tb]
    \centering
    \scriptsize
    \setlength\tabcolsep{3pt}
    \begin{tabular}{p{7.5cm} c c c c}
        \toprule
        \textbf{Original Document Title} &
        \textbf{Year} & \textbf{Size} &
        \textbf{Pages} & \textbf{No. Qs} \\
        \midrule
        \href{https://www.amauni.org/wp-content/uploads/2022/03/Set-of-GRI-Stnds-2021.pdf}{\emph{Consolidated Set of the GRI Standards 2021}} \cite{gri_consolidated_2021} & 2021 & 19 MB & 677 & 46 \\[2pt]        
        \href{https://www.globalreporting.org/how-to-use-the-gri-standards/gri-standards-english-language/}{\emph{GRI 1: Foundation 2021}} \cite{gri1_2021} & 2021 & 1.2 MB & 39 & 6 \\[2pt]        
        \href{https://www.globalreporting.org/how-to-use-the-gri-standards/gri-standards-english-language/}{\emph{GRI 2: General Disclosures 2021}} \cite{gri2_2021} & 2021 & 1.2 MB & 58 & 11 \\[2pt]        
        \href{https://www.globalreporting.org/how-to-use-the-gri-standards/gri-standards-english-language/}{\emph{GRI 3: Material Topics 2021}} \cite{gri3_2021} & 2021 & 1.1 MB & 30 & 5 \\[2pt]        
        \href{https://www.globalreporting.org/how-to-use-the-gri-standards/gri-standards-english-language/}{\emph{GRI 11: Oil and Gas Sector 2021}} \cite{gri11_2021} & 2021 & 2.2 MB & 93 & 14 \\[2pt]        
        \href{https://www.globalreporting.org/how-to-use-the-gri-standards/gri-standards-english-language/}{\emph{GRI 12: Coal Sector 2022}} \cite{gri12_2022} & 2022 & 2.1 MB & 86 & 16 \\[2pt]        
        \href{https://www.globalreporting.org/how-to-use-the-gri-standards/gri-standards-english-language/}{\emph{GRI 13: Agriculture, Aquaculture and Fishing Sectors 2022}} \cite{gri13_2022} & 2022 & 2.5 MB & 95 & 17 \\[2pt]        
        \href{https://www.globalreporting.org/how-to-use-the-gri-standards/gri-standards-english-language/}{\emph{GRI 14: Mining Sector 2024}} \cite{gri14_2024} & 2024 & 2.6 MB & 100 & 15 \\[2pt]        
        \href{https://www.globalreporting.org/how-to-use-the-gri-standards/gri-standards-english-language/}{\emph{GRI 101: Biodiversity 2024}} \cite{gri101_2024} & 2024 & 1.3 MB & 50 & 10 \\[2pt]        
        \href{https://www.globalreporting.org/how-to-use-the-gri-standards/gri-standards-english-language/}{\emph{GRI 201: Economic Performance 2016}} \cite{gri201_2016} & 2016 & 862 KB & 16 & 2 \\[2pt]       
        \href{https://www.globalreporting.org/how-to-use-the-gri-standards/gri-standards-english-language/}{\emph{GRI 202: Market Presence 2016}} \cite{gri202_2016} & 2016 & 834 KB & 15 & 1 \\[2pt]
        \href{https://www.globalreporting.org/how-to-use-the-gri-standards/gri-standards-english-language/}{\emph{GRI 203: Indirect Economic Impacts 2016}} \cite{gri203_2016} & 2016 & 817 KB & 11 & 3 \\[2pt]        
        \href{https://www.globalreporting.org/how-to-use-the-gri-standards/gri-standards-english-language/}{\emph{GRI 204: Procurement Practices 2016}} \cite{gri204_2016} & 2016 & 821 KB & 11 & 3 \\[2pt]       
        \href{https://www.globalreporting.org/how-to-use-the-gri-standards/gri-standards-english-language/}{\emph{GRI 205: Anti-corruption 2016}} \cite{gri205_2016} & 2016 & 855 KB & 16 & 1 \\[2pt]       
        \href{https://www.globalreporting.org/how-to-use-the-gri-standards/gri-standards-english-language/}{\emph{GRI 206: Anti-competitive Behavior 2016}} \cite{gri206_2016} & 2016 & 829 KB & 13 & 1 \\[2pt]       
        \href{https://www.globalreporting.org/how-to-use-the-gri-standards/gri-standards-english-language/}{\emph{GRI 207: Tax 2019}} \cite{gri207_2019} & 2019 & 947 KB & 21 & 3 \\[2pt]
        \href{https://www.globalreporting.org/how-to-use-the-gri-standards/gri-standards-english-language/}{\emph{GRI 301: Materials 2016}} \cite{gri301_2016} & 2016 & 831 KB & 13 & 1 \\[2pt]
        \href{https://www.globalreporting.org/how-to-use-the-gri-standards/gri-standards-english-language/}{\emph{GRI 302: Energy 2016}} \cite{gri302_2016} & 2016 & 859 KB & 19 & 1 \\[2pt]       
        \href{https://www.globalreporting.org/how-to-use-the-gri-standards/gri-standards-english-language/}{\emph{GRI 303: Water and Effluents 2018}} \cite{gri303_2018} & 2018 & 1 MB & 28 & 4 \\[2pt]
        \href{https://www.globalreporting.org/how-to-use-the-gri-standards/gri-standards-english-language/}{\emph{GRI 304: Biodiversity 2016}} \cite{gri304_2016} & 2016 & 845 KB & 15 & 1 \\[2pt]        
        \href{https://www.globalreporting.org/how-to-use-the-gri-standards/gri-standards-english-language/}{\emph{GRI 305: Emissions 2016}} \cite{gri305_2016} & 2016 & 936 KB & 26 & 4 \\[2pt]        
        \href{https://www.globalreporting.org/how-to-use-the-gri-standards/gri-standards-english-language/}{\emph{GRI 306: Effluents and Waste 2016}} \cite{gri306_effluents_2016} & 2016 & 640 KB & 15 & 1 \\[2pt]        
        \href{https://www.globalreporting.org/how-to-use-the-gri-standards/gri-standards-english-language/}{\emph{GRI 306: Waste 2020}} \cite{gri306_waste_2020} & 2020 & 1.7 MB & 30 & 1 \\[2pt]       
        \href{https://www.globalreporting.org/how-to-use-the-gri-standards/gri-standards-english-language/}{\emph{GRI 308: Supplier Environmental Assessment 2016}} \cite{gri308_2016} & 2016 & 851 KB & 14 & 1 \\[2pt]      
        \href{https://www.globalreporting.org/how-to-use-the-gri-standards/gri-standards-english-language/}{\emph{GRI 401: Employment 2016}} \cite{gri401_2016} & 2016 & 860 KB & 16 & 2 \\[2pt]   
        \href{https://www.globalreporting.org/how-to-use-the-gri-standards/gri-standards-english-language/}{\emph{GRI 402: Labor/Management Relations 2016}} \cite{gri402_2016} & 2016 & 855 KB & 13 & 1 \\[2pt]
        \href{https://www.globalreporting.org/how-to-use-the-gri-standards/gri-standards-english-language/}{\emph{GRI 403: Occupational Health and Safety 2018}} \cite{gri403_2018} & 2018 & 1.1 MB & 35 & 6 \\[2pt]        
        \href{https://www.globalreporting.org/how-to-use-the-gri-standards/gri-standards-english-language/}{\emph{GRI 404: Training and Education 2016}} \cite{gri404_2016} & 2016 & 837 KB & 15 & 0 \\[2pt]        
        \href{https://www.globalreporting.org/how-to-use-the-gri-standards/gri-standards-english-language/}{\emph{GRI 405: Diversity and Equal Opportunity 2016}} \cite{gri405_2016} & 2016 & 856 KB & 15 & 1 \\[2pt]         
        \href{https://www.globalreporting.org/how-to-use-the-gri-standards/gri-standards-english-language/}{\emph{GRI 406: Non-discrimination 2016}} \cite{gri406_2016} & 2016 & 853 KB & 12 & 1 \\[2pt]       
        \href{https://www.globalreporting.org/how-to-use-the-gri-standards/gri-standards-english-language/}{\emph{GRI 407: Freedom of Association and Collective Bargaining 2016}} \cite{gri407_2016} & 2016 & 870 KB & 13 & 1 \\[2pt]
        \href{https://www.globalreporting.org/how-to-use-the-gri-standards/gri-standards-english-language/}{\emph{GRI 408: Child Labor 2016}} \cite{gri408_2016} & 2016 & 868 KB & 14 & 0 \\[2pt]
        \href{https://www.globalreporting.org/how-to-use-the-gri-standards/gri-standards-english-language/}{\emph{GRI 409: Forced or Compulsory Labor 2016}} \cite{gri409_2016} & 2016 & 912 KB & 13 & 1 \\[2pt]        
        \href{https://www.globalreporting.org/how-to-use-the-gri-standards/gri-standards-english-language/}{\emph{GRI 410: Security Practices 2016}} \cite{gri410_2016} & 2016 & 841 KB & 11 & 4 \\[2pt]        
        \href{https://www.globalreporting.org/how-to-use-the-gri-standards/gri-standards-english-language/}{\emph{GRI 411: Rights of Indigenous Peoples 2016}} \cite{gri411_2016} & 2016 & 863 KB & 14 & 1 \\[2pt]       
        \href{https://www.globalreporting.org/how-to-use-the-gri-standards/gri-standards-english-language/}{\emph{GRI 413: Local Communities 2016}} \cite{gri413_2016} & 2016 & 885 KB & 16 & 2 \\[2pt]
        \href{https://www.globalreporting.org/how-to-use-the-gri-standards/gri-standards-english-language/}{\emph{GRI 414: Supplier Social Assessment 2016}} \cite{gri414_2016} & 2016 & 853 KB & 14 & 1 \\[2pt] 
        \href{https://www.globalreporting.org/how-to-use-the-gri-standards/gri-standards-english-language/}{\emph{GRI 415: Public Policy 2016}} \cite{gri415_2016} & 2016 & 816 KB & 12 & 1 \\[2pt]
        \href{https://www.globalreporting.org/how-to-use-the-gri-standards/gri-standards-english-language/}{\emph{GRI 416: Customer Health and Safety 2016}} \cite{gri416_2016} & 2016 & 825 KB & 12 & 1 \\[2pt]       
        \href{https://www.globalreporting.org/how-to-use-the-gri-standards/gri-standards-english-language/}{\emph{GRI 417: Marketing and Labeling 2016}} \cite{gri417_2016} & 2016 & 837 KB & 14 & 1 \\[2pt]       
        \href{https://www.globalreporting.org/how-to-use-the-gri-standards/gri-standards-english-language/}{\emph{GRI 418: Customer Privacy 2016}} \cite{gri418_2016} & 2016 & 825 KB & 12 & 1 \\[2pt]        
        \href{https://www.globalreporting.org/how-to-use-the-gri-standards/gri-standards-english-language/}{\emph{GRI Standards Glossary 2022}} \cite{gri_glossary_2022} & 2022 & 680 KB & 23 & 2 \\[2pt]        
        \href{https://www.globalreporting.org/media/mlkjpn1i/gri-sasb-joint-publication-april-2021.pdf}{\emph{A Practical Guide to Sustainability Reporting Using GRI and SASB Standards}} \cite{gri_practical_guide_2021} & 2021 & 1.5 MB & 42 & 4 \\[2pt]
        \href{https://www.globalreporting.org/media/nmmnwfsm/gri-policymakers-guide.pdf}{\emph{The GRI Standards — A Guide for Policy Makers}} \cite{gri_policy_makers_2021} & 2021 & 7.1 MB & 19 & 2 \\[2pt]
        \midrule
        Total & - & 70.863MB & 1,826 & 201 \\
        \bottomrule
    \end{tabular}
    \caption{Metadata for the \textbf{GRI Standards} collection in \textbf{ESGenius}. This comprehensive collection spans 2016-2024 and comprises: (1) Universal Standards (GRI 1-3) establishing core reporting principles, (2) Sector Standards (GRI 11-14) for high-impact industries, and (3) Topic Standards covering economic (200 series), environmental (300 series), and social (400 series) aspects. Key features include: \textbf{Coverage:} 1,826 pages across 40+ standards documents; \textbf{Evaluation:} 201 domain-specific questions; \textbf{Scope:} Comprehensive ESG disclosure requirements spanning corporate governance, environmental impact, and social responsibility. All standards sourced from GRI (\url{https://www.globalreporting.org/}), the leading authority in sustainability reporting frameworks.}
    \label{tab:gri_table}
\end{table*}
\clearpage

\begin{table*}[tb]
    \centering
    \scriptsize
    \setlength\tabcolsep{3pt}
    \begin{tabular}{p{11cm} c c c c}
      \toprule
      \textbf{Original Document Title} &
      \textbf{Year} & \textbf{Size} & \textbf{Pages} & \textbf{No. Qs} \\
      \midrule
      \href{https://d3flraxduht3gu.cloudfront.net/latest_standards/apparel-accessories-and-footwear-standard_en-gb.pdf}{\emph{Apparel, Accessories \& Footwear - SUSTAINABILITY ACCOUNTING STANDARD}} \cite{sasb_apparel_accessories_footwear_2023} & 2023 & 410 KB & 21 & 4 \\[2pt] 
      \href{https://d3flraxduht3gu.cloudfront.net/latest_standards/appliance-manufacturing-standard_en-gb.pdf}{\emph{Appliance Manufacturing - SUSTAINABILITY ACCOUNTING STANDARD}} \cite{sasb_appliance_manufacturing_2023} & 2023 & 340 KB & 13 & 1 \\[2pt]  
      \href{https://d3flraxduht3gu.cloudfront.net/latest_standards/building-products-and-furnishings-standard_en-gb.pdf}{\emph{Building Products \& Furnishings - SUSTAINABILITY ACCOUNTING STANDARD}} \cite{sasb_building_products_furnishings_2023} & 2023 & 369 KB & 18 & 1 \\[2pt]
      \href{https://d3flraxduht3gu.cloudfront.net/latest_standards/e-commerce-standard_en-gb.pdf}{\emph{E-commerce - SUSTAINABILITY ACCOUNTING STANDARD}} \cite{sasb_e_commerce_2023} & 2023 & 402 KB & 24 & 3 \\[2pt] 
      \href{https://d3flraxduht3gu.cloudfront.net/latest_standards/household-and-personal-products-standard_en-gb.pdf}{\emph{Household \& Personal Products - SUSTAINABILITY ACCOUNTING STANDARD}} \cite{sasb_household_personal_products_2023} & 2023 & 371 KB & 19 & 1 \\[2pt] 
      \href{https://d3flraxduht3gu.cloudfront.net/latest_standards/multiline-and-specialty-retailers-and-distributors-standard_en-gb.pdf}{\emph{Multiline \& Specialty Retailers \& Distributors - SUSTAINABILITY ACCOUNTING STANDARD}} \cite{sasb_multiline_specialty_retailers_distributors_2023} & 2023 & 399 KB & 24 & 3 \\[2pt] 
      \href{https://d3flraxduht3gu.cloudfront.net/latest_standards/toys-and-sporting-goods-standard_en-gb.pdf}{\emph{Toys \& Sporting Goods - SUSTAINABILITY ACCOUNTING STANDARD}} \cite{sasb_toys_sporting_goods_2023} & 2023 & 346 KB & 13 & 0 \\[2pt] 
      \href{https://d3flraxduht3gu.cloudfront.net/latest_standards/coal-operations-standard_en-gb.pdf}{\emph{Coal Operations - SUSTAINABILITY ACCOUNTING STANDARD}} \cite{sasb_coal_operations_2023} & 2023 & 490 KB & 41 & 8 \\[2pt]  
      \href{https://d3flraxduht3gu.cloudfront.net/latest_standards/construction-materials-standard_en-gb.pdf}{\emph{Construction Materials - SUSTAINABILITY ACCOUNTING STANDARD}} \cite{sasb_construction_materials_2023} & 2023 & 414 KB & 27 & 3 \\[2pt]
      \href{https://d3flraxduht3gu.cloudfront.net/latest_standards/iron-and-steel-producers-standard_en-gb.pdf}{\emph{Iron \& Steel Producers - SUSTAINABILITY ACCOUNTING STANDARD}} \cite{sasb_iron_steel_producers_2023} & 2023 & 392 KB & 24 & 4 \\[2pt]
      \href{https://d3flraxduht3gu.cloudfront.net/latest_standards/metals-and-mining-standard_en-gb.pdf}{\emph{Metals \& Mining - SUSTAINABILITY ACCOUNTING STANDARD}} \cite{sasb_metals_mining_2023} & 2023 & 521 KB & 47 & 8 \\[2pt]
      \href{https://d3flraxduht3gu.cloudfront.net/latest_standards/oil-and-gas-exploration-and-production-standard_en-gb.pdf}{\emph{Oil \& Gas – Exploration \& Production - SUSTAINABILITY ACCOUNTING STANDARD}} \cite{sasb_oil_gas_exploration_production_2023} & 2023 & 511 KB & 47 & 6 \\[2pt]
      \href{https://d3flraxduht3gu.cloudfront.net/latest_standards/oil-and-gas-midstream-standard_en-gb.pdf}{\emph{Oil \& Gas – Midstream - SUSTAINABILITY ACCOUNTING STANDARD}} \cite{sasb_oil_gas_midstream_2023} & 2023 & 407 KB & 26 & 4 \\[2pt]
      \href{https://d3flraxduht3gu.cloudfront.net/latest_standards/oil-and-gas-refining-and-marketing-standard_en-gb.pdf}{\emph{Oil \& Gas – Refining \& Marketing - SUSTAINABILITY ACCOUNTING STANDARD}} \cite{sasb_oil_gas_refining_marketing_2023} & 2023 & 433 KB & 31 & 5 \\[2pt]
      \href{https://d3flraxduht3gu.cloudfront.net/latest_standards/oil-and-gas-services-standard_en-gb.pdf}{\emph{Oil \& Gas – Services - SUSTAINABILITY ACCOUNTING STANDARD}} \cite{sasb_oil_gas_services_2023} & 2023 & 413 KB & 28 & 4 \\[2pt]
      \href{https://d3flraxduht3gu.cloudfront.net/latest_standards/asset-management-and-custody-activities-standard_en-gb.pdf}{\emph{Asset Management \& Custody Activities - SUSTAINABILITY ACCOUNTING STANDARD}} \cite{sasb_asset_management_custody_activities_2023} & 2023 & 404 KB & 24 & 3 \\[2pt]
      \href{https://d3flraxduht3gu.cloudfront.net/latest_standards/commercial-banks-standard_en-gb.pdf}{\emph{Commercial Banks - SUSTAINABILITY ACCOUNTING STANDARD}} \cite{sasb_commercial_banks_2023} & 2023 & 394 KB & 23 & 3 \\[2pt]  
      \href{https://d3flraxduht3gu.cloudfront.net/latest_standards/consumer-finance-standard_en-gb.pdf}{\emph{Consumer Finance - SUSTAINABILITY ACCOUNTING STANDARD}} \cite{sasb_consumer_finance_2023} & 2023 & 372 KB & 20 & 3 \\[2pt]  
      \href{https://d3flraxduht3gu.cloudfront.net/latest_standards/insurance-standard_en-gb.pdf}{\emph{Insurance - SUSTAINABILITY ACCOUNTING STANDARD}} \cite{sasb_insurance_2023} & 2023 & 408 KB & 26 & 3 \\[2pt]  
      \href{https://d3flraxduht3gu.cloudfront.net/latest_standards/investment-banking-and-brokerage-standard_en-gb.pdf}{\emph{Investment Banking \& Brokerage - SUSTAINABILITY ACCOUNTING STANDARD}} \cite{sasb_investment_banking_brokerage_2023} & 2023 & 420 KB & 28 & 4 \\[2pt]
      \href{https://d3flraxduht3gu.cloudfront.net/latest_standards/mortgage-finance-standard_en-gb.pdf}{\emph{Mortgage Finance - SUSTAINABILITY ACCOUNTING STANDARD}} \cite{sasb_mortgage_finance_2023} & 2023 & 366 KB & 17 & 2 \\[2pt]  
      \href{https://d3flraxduht3gu.cloudfront.net/latest_standards/security-and-commodity-exchanges-standard_en-gb.pdf}{\emph{Security \& Commodity Exchanges - SUSTAINABILITY ACCOUNTING STANDARD}} \cite{sasb_security_commodity_exchanges_2023} & 2023 & 362 KB & 17 & 1 \\[2pt]  
      \href{https://d3flraxduht3gu.cloudfront.net/latest_standards/agricultural-products-standard_en-gb.pdf}{\emph{Agricultural Products - SUSTAINABILITY ACCOUNTING STANDARD}} \cite{sasb_agricultural_products_2023} & 2023 & 426 KB & 30 & 3 \\[2pt] 
      \href{https://d3flraxduht3gu.cloudfront.net/latest_standards/alcoholic-beverages-standard_en-gb.pdf}{\emph{Alcoholic Beverages - SUSTAINABILITY ACCOUNTING STANDARD}} \cite{sasb_alcoholic_beverages_2023} & 2023 & 400 KB & 26 & 4 \\[2pt]
      \href{https://d3flraxduht3gu.cloudfront.net/latest_standards/food-retailers-and-distributors-standard_en-gb.pdf}{\emph{Food Retailers \& Distributors - SUSTAINABILITY ACCOUNTING STANDARD}} \cite{sasb_food_retailers_distributors_2023} & 2023 & 452 KB & 37 & 5 \\[2pt]
      \href{https://d3flraxduht3gu.cloudfront.net/latest_standards/meat-poultry-and-dairy-standard_en-gb.pdf}{\emph{Meat, Poultry \& Dairy - SUSTAINABILITY ACCOUNTING STANDARD}} \cite{sasb_meat_poultry_dairy_2023} & 2023 & 442 KB & 34 & 6 \\[2pt] 
      \href{https://d3flraxduht3gu.cloudfront.net/latest_standards/non-alcoholic-beverages-standard_en-gb.pdf}{\emph{Non-Alcoholic Beverages - SUSTAINABILITY ACCOUNTING STANDARD}} \cite{sasb_non_alcoholic_beverages_2023} & 2023 & 419 KB & 30 & 3 \\[2pt] 
      \href{https://d3flraxduht3gu.cloudfront.net/latest_standards/processed-foods-standard_en-gb.pdf}{\emph{Processed Foods - SUSTAINABILITY ACCOUNTING STANDARD}} \cite{sasb_processed_foods_2023} & 2023 & 434 KB & 32 & 5 \\[2pt]
      \href{https://d3flraxduht3gu.cloudfront.net/latest_standards/restaurants-standard_en-gb.pdf}{\emph{Restaurants - SUSTAINABILITY ACCOUNTING STANDARD}} \cite{sasb_restaurants_2023} & 2023 & 418 KB & 30 & 5 \\[2pt]  
      \href{https://d3flraxduht3gu.cloudfront.net/latest_standards/tobacco-standard_en-gb.pdf}{\emph{Tobacco - SUSTAINABILITY ACCOUNTING STANDARD}} \cite{sasb_tobacco_2023} & 2023 & 337 KB & 12 & 1 \\[2pt]  
      \href{https://d3flraxduht3gu.cloudfront.net/latest_standards/biotechnology-and-pharmaceuticals-standard_en-gb.pdf}{\emph{Biotechnology \& Pharmaceuticals - SUSTAINABILITY ACCOUNTING STANDARD}} \cite{sasb_biotechnology_pharmaceuticals_2023} & 2023 & 415 KB & 29 & 4 \\[2pt]
      \href{https://d3flraxduht3gu.cloudfront.net/latest_standards/drug-retailers-standard_en-gb.pdf}{\emph{Drug Retailers - SUSTAINABILITY ACCOUNTING STANDARD}} \cite{sasb_drug_retailers_2023} & 2023 & 378 KB & 21 & 3 \\[2pt]  
      \href{https://d3flraxduht3gu.cloudfront.net/latest_standards/health-care-delivery-standard_en-gb.pdf}{\emph{Health Care Delivery - SUSTAINABILITY ACCOUNTING STANDARD}} \cite{sasb_health_care_delivery_2023} & 2023 & 432 KB & 33 & 6 \\[2pt] 
      \href{https://d3flraxduht3gu.cloudfront.net/latest_standards/health-care-distributors-standard_en-gb.pdf}{\emph{Health Care Distributors - SUSTAINABILITY ACCOUNTING STANDARD}} \cite{sasb_health_care_distributors_2023} & 2023 & 358 KB & 17 & 1 \\[2pt] 
      \href{https://d3flraxduht3gu.cloudfront.net/latest_standards/managed-care-standard_en-gb.pdf}{\emph{Managed Care - SUSTAINABILITY ACCOUNTING STANDARD}} \cite{sasb_managed_care_2023} & 2023 & 371 KB & 20 & 3 \\[2pt]  
      \href{https://d3flraxduht3gu.cloudfront.net/latest_standards/medical-equipment-and-supplies-standard_en-gb.pdf}{\emph{Medical Equipment \& Supplies - SUSTAINABILITY ACCOUNTING STANDARD}} \cite{sasb_medical_equipment_supplies_2023} & 2023 & 380 KB & 22 & 2 \\[2pt]
      \href{https://d3flraxduht3gu.cloudfront.net/latest_standards/electric-utilities-and-power-generators-standard_en-gb.pdf}{\emph{Electric Utilities \& Power Generators - SUSTAINABILITY ACCOUNTING STANDARD}} \cite{sasb_electric_utilities_power_generators_2023} & 2023 & 458 KB & 35 & 4 \\[2pt]  
      \href{https://d3flraxduht3gu.cloudfront.net/latest_standards/engineering-and-construction-services-standard_en-gb.pdf}{\emph{Engineering \& Construction Service - SUSTAINABILITY ACCOUNTING STANDARD}} \cite{sasb_engineering_construction_services_2023} & 2023 & 402 KB & 26 & 4 \\[2pt]
      \bottomrule
    \end{tabular}
    \caption{Metadata for SASB industry-specific disclosure standards — \textbf{Part I}.}
    \label{tab:sasb_part1_table}
  \end{table*}
  \clearpage
  \begin{table*}[tb]
    \centering
    \scriptsize
    \setlength\tabcolsep{3pt}
    \begin{tabular}{p{12cm} c c c c}
      \toprule
      \textbf{Original Document Title} &
      \textbf{Year} & \textbf{Size} & \textbf{Pages} & \textbf{No. Qs} \\
      \midrule
      \href{https://d3flraxduht3gu.cloudfront.net/latest_standards/gas-utilities-and-distributors-standard_en-gb.pdf}{\emph{Gas Utilities \& Distributors - SUSTAINABILITY ACCOUNTING STANDARD}} \cite{sasb_gas_utilities_distributors_2023} & 2023 & 372 KB & 19 & 0 \\[2pt] 
      \href{https://d3flraxduht3gu.cloudfront.net/latest_standards/home-builders-standard_en-gb.pdf}{\emph{Home Builders - SUSTAINABILITY ACCOUNTING STANDARD}} \cite{sasb_home_builders_2023} & 2023 & 383 KB & 21 & 1 \\[2pt]  
      \href{https://d3flraxduht3gu.cloudfront.net/latest_standards/real-estate-standard_en-gb.pdf}{\emph{Real Estate - SUSTAINABILITY ACCOUNTING STANDARD}} \cite{sasb_real_estate_2023} & 2023 & 285 KB & 38 & 5 \\[2pt]
      \href{https://d3flraxduht3gu.cloudfront.net/latest_standards/real-estate-services-standard_en-gb.pdf}{\emph{Real Estate Services - SUSTAINABILITY ACCOUNTING STANDARD}} \cite{sasb_real_estate_services_2023} & 2023 & 349 KB & 15 & 0 \\[2pt]   
      \href{https://d3flraxduht3gu.cloudfront.net/latest_standards/waste-management-standard_en-gb.pdf}{\emph{Waste Management - SUSTAINABILITY ACCOUNTING STANDARD}} \cite{sasb_waste_management_2023} & 2023 & 430 KB & 31 & 5 \\[2pt] 
      \href{https://d3flraxduht3gu.cloudfront.net/latest_standards/water-utilities-and-services-standard_en-gb.pdf}{\emph{Water Utilities \& Services - SUSTAINABILITY ACCOUNTING STANDARD}}  \cite{sasb_water_utilities_services_2023} & 2023 & 436 KB & 32 & 4 \\[2pt]  
      \href{https://d3flraxduht3gu.cloudfront.net/latest_standards/biofuels-standard_en-gb.pdf}{\emph{Biofuels - SUSTAINABILITY ACCOUNTING STANDARD}} \cite{sasb_biofuels_2023} & 2023 & 380 KB & 20 & 1 \\[2pt] 
      \href{https://d3flraxduht3gu.cloudfront.net/latest_standards/forestry-management-standard_en-gb.pdf}{\emph{Forestry Management - SUSTAINABILITY ACCOUNTING STANDARD}} \cite{sasb_forestry_management_2023} & 2023 & 363 KB & 18 & 2 \\[2pt] 
      \href{https://d3flraxduht3gu.cloudfront.net/latest_standards/fuel-cells-and-industrial-batteries-standard_en-gb.pdf}{\emph{Fuel Cells \& Industrial Batteries - SUSTAINABILITY ACCOUNTING STANDARD}} \cite{sasb_fuel_cells_industrial_batteries_2023} & 2023 & 381 KB & 21 & 3 \\[2pt] 
      \href{https://d3flraxduht3gu.cloudfront.net/latest_standards/pulp-and-paper-products-standard_en-gb.pdf}{\emph{Pulp \& Paper Products - SUSTAINABILITY ACCOUNTING STANDARD}} \cite{sasb_pulp_paper_products_2023} & 2023 & 390 KB & 24 & 2 \\[2pt] 
      \href{https://d3flraxduht3gu.cloudfront.net/latest_standards/solar-technology-and-project-developers-standard_en-gb.pdf}{\emph{Solar Technology \& Project Developers - SUSTAINABILITY ACCOUNTING STANDARD}} \cite{sasb_solar_technology_project_developers_2023} & 2023 & 410 KB & 28 & 5 \\[2pt] 
      \href{https://d3flraxduht3gu.cloudfront.net/latest_standards/wind-technology-and-project-developers-standard_en-gb.pdf}{\emph{Wind Technology \& Project Developers - SUSTAINABILITY ACCOUNTING STANDARD}} \cite{sasb_wind_technology_project_developers_2023} & 2023 & 358 KB & 17 & 2 \\[2pt] 
      \href{https://d3flraxduht3gu.cloudfront.net/latest_standards/aerospace-and-defence-standard_en-gb.pdf}{\emph{Aerospace \& Defence - SUSTAINABILITY ACCOUNTING STANDARD}} \cite{sasb_aerospace_defense_2023} & 2023 & 439 KB & 27 & 4 \\[2pt]  
      \href{https://d3flraxduht3gu.cloudfront.net/latest_standards/chemicals-standard_en-gb.pdf}{\emph{Chemicals - SUSTAINABILITY ACCOUNTING STANDARD}} \cite{sasb_chemicals_2023} & 2023 & 447 KB & 36 & 7 \\[2pt] 
      \href{https://d3flraxduht3gu.cloudfront.net/latest_standards/containers-and-packaging-standard_en-gb.pdf}{\emph{Containers \& Packaging - SUSTAINABILITY ACCOUNTING STANDARD}} \cite{sasb_containers_packaging_2023} & 2023 & 422 KB & 30 & 5 \\[2pt] 
      \href{https://d3flraxduht3gu.cloudfront.net/latest_standards/electrical-and-electronic-equipment-standard_en-gb.pdf}{\emph{Electrical \& Electronic Equipment - SUSTAINABILITY ACCOUNTING STANDARD}} \cite{sasb_electrical_electronic_equipment_2023} & 2023 & 385 KB & 23 & 3 \\[2pt]
      \href{https://d3flraxduht3gu.cloudfront.net/latest_standards/industrial-machinery-and-goods-standard_en-gb.pdf}{\emph{Industrial Machinery \& Goods - SUSTAINABILITY ACCOUNTING STANDARD}} \cite{sasb_industrial_machinery_goods_2023} & 2023 & 361 KB & 17 & 2 \\[2pt]
      \href{https://d3flraxduht3gu.cloudfront.net/latest_standards/advertising-and-marketing-standard_en-gb.pdf}{\emph{Advertising \& Marketing - SUSTAINABILITY ACCOUNTING STANDARD}} \cite{sasb_advertising_marketing_2023} & 2023 & 368 KB & 18 & 1 \\[2pt] 
      \href{https://d3flraxduht3gu.cloudfront.net/latest_standards/casinos-and-gaming-standard_en-gb.pdf}{\emph{Casinos \& Gaming - SUSTAINABILITY ACCOUNTING STANDARD}} \cite{sasb_casinos_gaming_2023} & 2023 & 353 KB & 15 & 1 \\[2pt] 
      \href{https://d3flraxduht3gu.cloudfront.net/latest_standards/education-standard_en-gb.pdf}{\emph{Education - SUSTAINABILITY ACCOUNTING STANDARD}} \cite{sasb_education_2023} & 2023 & 366 KB & 18 & 1 \\[2pt] 
      \href{https://d3flraxduht3gu.cloudfront.net/latest_standards/hotels-and-lodging-standard_en-gb.pdf}{\emph{Hotels \& Lodging - SUSTAINABILITY ACCOUNTING STANDARD}} \cite{sasb_hotels_lodging_2023} & 2023 & 371 KB & 19 & 2 \\[2pt]  
      \href{https://d3flraxduht3gu.cloudfront.net/latest_standards/leisure-facilities-standard_en-gb.pdf}{\emph{Leisure Facilities - SUSTAINABILITY ACCOUNTING STANDARD}} \cite{sasb_leisure_facilities_2023} & 2023 & 343 KB & 13 & 1 \\[2pt]  
      \href{https://d3flraxduht3gu.cloudfront.net/latest_standards/media-and-entertainment-standard_en-gb.pdf}{\emph{Media \& Entertainment - SUSTAINABILITY ACCOUNTING STANDARD}} \cite{sasb_media_entertainment_2023} & 2023 & 357 KB & 15 & 1 \\[2pt]
      \href{https://d3flraxduht3gu.cloudfront.net/latest_standards/professional-and-commercial-services-standard_en-gb.pdf}{\emph{Professional \& Commercial Services - SUSTAINABILITY ACCOUNTING STANDARD}} \cite{sasb_professional_commercial_services_2023} & 2023 & 371 KB & 18 & 2 \\[2pt] 
      \href{https://d3flraxduht3gu.cloudfront.net/latest_standards/electronic-manufacturing-services-and-original-design-manufacturing-standard_en-gb.pdf}{\emph{Electronic Manufacturing Services \& Original Design Manufacturing - SUSTAINABILITY ACCOUNTING STANDARD}} \cite{sasb_electronic_manufacturing_services_odm_2023} & 2023 & 379 KB & 22 & 3 \\[2pt] 
      \href{https://d3flraxduht3gu.cloudfront.net/latest_standards/hardware-standard_en-gb.pdf}{\emph{Hardware - SUSTAINABILITY ACCOUNTING STANDARD}} \cite{sasb_hardware_2023} & 2023 & 392 KB & 24 & 3 \\[2pt]  
      \href{https://d3flraxduht3gu.cloudfront.net/latest_standards/internet-media-and-services-standard_en-gb.pdf}{\emph{Internet Media \& Services - SUSTAINABILITY ACCOUNTING STANDARD}} \cite{sasb_internet_media_services_2023} & 2023 & 417 KB & 27 & 4 \\[2pt] 
      \href{https://d3flraxduht3gu.cloudfront.net/latest_standards/semiconductors-standard_en-gb.pdf}{\emph{Semiconductors - SUSTAINABILITY ACCOUNTING STANDARD}} \cite{sasb_semiconductors_2023} & 2023 & 409 KB & 27 & 4 \\[2pt]  
      \href{https://d3flraxduht3gu.cloudfront.net/latest_standards/software-and-it-services-standard_en-gb.pdf}{\emph{Software \& IT Services - SUSTAINABILITY ACCOUNTING STANDARD}} \cite{sasb_software_it_services_2023} & 2023 & 426 KB & 29 & 4 \\[2pt] 
      \href{https://d3flraxduht3gu.cloudfront.net/latest_standards/telecommunication-services-standard_en-gb.pdf}{\emph{Telecommunication Services - SUSTAINABILITY ACCOUNTING STANDARD}} \cite{sasb_telecommunication_services_2023} & 2023 & 409 KB & 26 & 2 \\[2pt]  
      \href{https://d3flraxduht3gu.cloudfront.net/latest_standards/air-freight-and-logistics-standard_en-gb.pdf}{\emph{Air Freight \& Logistics - SUSTAINABILITY ACCOUNTING STANDARD}} \cite{sasb_air_freight_logistics_2023} & 2023 & 394 KB & 24 & 4 \\[2pt] 
      \href{https://d3flraxduht3gu.cloudfront.net/latest_standards/airlines-standard_en-gb.pdf}{\emph{Airlines - SUSTAINABILITY ACCOUNTING STANDARD}} \cite{sasb_airlines_2023} & 2023 & 371 KB & 19 & 2 \\[2pt]
      \href{https://d3flraxduht3gu.cloudfront.net/latest_standards/auto-parts-standard_en-gb.pdf}{\emph{Auto Parts - SUSTAINABILITY ACCOUNTING STANDARD}} \cite{sasb_auto_parts_2023} & 2023 & 373 KB & 20 & 2 \\[2pt] 
      \href{https://d3flraxduht3gu.cloudfront.net/latest_standards/automobiles-standard_en-gb.pdf}{\emph{Automobiles - SUSTAINABILITY ACCOUNTING STANDARD}} \cite{sasb_automobiles_2023} & 2023 & 378 KB & 22 & 3 \\[2pt]  
      \href{https://d3flraxduht3gu.cloudfront.net/latest_standards/car-rental-and-leasing-standard_en-gb.pdf}{\emph{Car Rental \& Leasing - SUSTAINABILITY ACCOUNTING STANDARD}} \cite{sasb_car_rental_leasing_2023} & 2023 & 334 KB & 11 & 3 \\[2pt]  
      \href{https://d3flraxduht3gu.cloudfront.net/latest_standards/cruise-lines-standard_en-gb.pdf}{\emph{Cruise Lines - SUSTAINABILITY ACCOUNTING STANDARD}} \cite{sasb_cruise_lines_2023} & 2023 & 414 KB & 26 & 3 \\[2pt]  
      \href{https://d3flraxduht3gu.cloudfront.net/latest_standards/marine-transportation-standard_en-gb.pdf}{\emph{Marine Transportation - SUSTAINABILITY ACCOUNTING STANDARD}} \cite{sasb_marine_transportation_2023} & 2023 & 400 KB & 24 & 2 \\[2pt]
      \href{https://d3flraxduht3gu.cloudfront.net/latest_standards/rail-transportation-standard_en-gb.pdf}{\emph{Rail Transportation - SUSTAINABILITY ACCOUNTING STANDARD}} \cite{sasb_rail_transportation_2023} & 2023 & 381 KB & 21 & 3 \\[2pt]
      \href{https://d3flraxduht3gu.cloudfront.net/latest_standards/road-transportation-standard_en-gb.pdf}{\emph{Road Transportation - SUSTAINABILITY ACCOUNTING STANDARD}} \cite{sasb_road_transportation_2023} & 2023 & 368 KB & 17 & 1 \\[2pt]
      \midrule
      \textbf{Total} & & 30.431MB & 1,864 & 236 \\[2pt]
      \bottomrule
    \end{tabular}
    \caption{Comprehensive metadata for all \textbf{77 industry-specific SASB Sustainability Accounting Standards} issued in 2023 and now stewarded by the IFRS Foundation's ISSB. This corpus, distilled for our \textbf{ESGenius} system, comprises 77 documents totalling \textbf{30.431 MB}, \textbf{1,864 pages}, and \textbf{236 MCQs}. The standards span the full economy—from \emph{Apparel} to \emph{Transportation}—and cover financially material sustainability topics including greenhouse gas emissions, water management, data security, and workforce diversity. These standards provide investors with decision-relevant ESG information for valuation, risk assessment, and stewardship. The complete collection is freely accessible at \url{https://www.sasb.org/}.}
    \label{tab:sasb_part2_table}
  \end{table*}
  \clearpage

\begin{table*}[tb]
    \centering
    \scriptsize
    \setlength\tabcolsep{3pt}
    \begin{tabular}{p{12cm} c c c c}
      \toprule
      \textbf{Original Document Title} & \textbf{Year} & \textbf{Size} & \textbf{Pages} & \textbf{No. Qs} \\
      \midrule
      % ---------- core standards / explanatory material ----------
      \href{https://www.ifrs.org/content/dam/ifrs/publications/pdf-standards-issb/english/2023/issued/part-a/issb-2023-a-ifrs-s1-general-requirements-for-disclosure-of-sustainability-related-financial-information.pdf}{\emph{IFRS S1}} \cite{ifrs_s1_2023} & 2023 & 307 KB & 48 & 8 \\[2pt]
      \href{https://www.ifrs.org/content/dam/ifrs/publications/pdf-standards-issb/english/2023/issued/part-a/issb-2023-a-ifrs-s2-climate-related-disclosures.pdf?bypass=on}{\emph{IFRS S2}} \cite{ifrs_s2_2023} & 2023 & 297 KB & 46 & 8 \\[2pt]
      \href{https://www.ifrs.org/content/dam/ifrs/supporting-implementation/issb-standards/progress-climate-related-disclosures-2024.pdf}{\emph{Progress on Corporate Climate-related Disclosures — 2024 Report}} \cite{ifrs_progress_climate_disclosures_2024} & 2024 & 1.3 MB & 164 & 27 \\
      \href{https://www.ifrs.org/content/dam/ifrs/publications/amendments/english/2023/issb-2023-c-basis-for-conclusions-on-ifrs-s1-general-requirements-for-disclosure-of-sustainability-related-financial-information-part-c.pdf}{\emph{IFRS S1 Basis for Conclusions on General Requirements for Disclosure of Sustainability-related Financial Information}} \cite{ifrs_s1_basis_conclusions_2023} & 2023 & 341 KB & 57 & 10 \\[2pt]
      \href{https://www.ifrs.org/content/dam/ifrs/publications/amendments/english/2023/issb-2023-c-basis-for-conclusions-on-ifrs-s2-climate-related-disclosures-part-c.pdf}{\emph{IFRS S2 Basis for Conclusions on Climate-related Disclosures}} \cite{ifrs_s2_basis_conclusions_2023} & 2023 & 337 KB & 55 & 11 \\[2pt]
      \href{https://www.ifrs.org/content/dam/ifrs/publications/pdf-standards-issb/english/2023/issued/part-b/ifrs-s2-ibg.pdf}{\emph{IFRS S2 Industry-based Guidance on implementing Climate-related Disclosures}} \cite{ifrs_s2_industry_guidance_master_2023} & 2023 & 2.3 MB & 538 & 33 \\[2pt]
      \href{https://www.ifrs.org/content/dam/ifrs/publications/pdf-standards-issb/english/2023/issued/part-b/issb-2023-b-ifrs-s1-general-requirements-for-disclosure-of-sustainability-related-financial-information-accompanying-guidance-part-b.pdf}{\emph{IFRS S1 Accompanying Guidance on General Requirements for Disclosure of Sustainability-related Financial Information}} \cite{ifrs_s1_accompanying_guidance_2023} & 2023 & 201 KB & 17 & 2 \\[2pt]
      \href{https://www.ifrs.org/content/dam/ifrs/project/amendments-greenhouse-gas-s2/issb-ed-2025-1-greenhouse-gas-s2.pdf}{\emph{Exposure Draft Amendments to Greenhouse Gas Emissions Disclosures Proposed amendments to IFRS S2 Comments to be received by 27 June 2025}} \cite{ifrs_ed_ghg_amendments_2025} & 2025 & 239 KB & 24 & 4 \\[2pt]
      \href{https://www.ifrs.org/content/dam/ifrs/project/amendments-greenhouse-gas-s2/issb-ed-2025-1-greenhouse-gas-s2-bc.pdf}{\emph{Exposure Draft Basis for Conclusions on Amendments to Greenhouse Gas Emissions Disclosures Proposed amendments to IFRS S2 Comments to be received by 27 June 2025}} \cite{ifrs_ed_basis_conclusions_ghg_amendments_2025} & 2025 & 238 KB & 25 & 4 \\[2pt]
      \href{https://www.ifrs.org/content/dam/ifrs/supporting-implementation/ifrs-s2/ifrs-s2-comparison-tcfd.pdf}{\emph{Comparison IFRS S2 Climate-related Disclosures with the TCFD recommendations}} \cite{ifrs_s2_vs_tcfd_comparison_2023} & 2024 & 105 KB & 12 & 1 \\[2pt] 
      \href{https://www.ifrs.org/content/dam/ifrs/publications/amendments/english/2024/issb-tu-2024-1-ifrs-sustainability-disclosure-taxonomy.pdf}{\emph{IFRS Taxonomy IFRS Sustainability Disclosure Taxonomy 2024 IFRS S1 General Requirements for Disclosure of Sustainability-related Financial Information and IFRS S2 Climate-related Disclosures}} \cite{ifrs_sustainability_taxonomy_2024} & 2024 & 1.1 MB & 76 & 10 \\[2pt]  
      \midrule
      % ---------- IFRS S2 industry-based guidance  Volumes 1 – 30 ----------
      \href{https://www.ifrs.org/content/dam/ifrs/publications/pdf-standards-issb/english/2023/issued/part-b/ifrs-s2-ibg-volume-1-apparel-accessories-and-footwear-part-b.pdf}{\emph{IFRS S2 Industry-based Guidance on implementing Climate-related Disclosures Vol 1 — Apparel, Accessories \& Footwear}} \cite{ifrs_s2_vol01_apparel_accessories_footwear_2023} & 2023 & 194 KB & 10 & 4 \\[2pt]
      \href{https://www.ifrs.org/content/dam/ifrs/publications/pdf-standards-issb/english/2023/issued/part-b/ifrs-s2-ibg-volume-2-appliance-manufacturing-part-b.pdf}{\emph{IFRS S2 Industry-based Guidance on implementing Climate-related Disclosures Vol 2 — Appliance Manufacturing}} \cite{ifrs_s2_vol02_appliance_manufacturing_2023} & 2023 & 179 KB & 7 & 3 \\[2pt]
      \href{https://www.ifrs.org/content/dam/ifrs/publications/pdf-standards-issb/english/2023/issued/part-b/ifrs-s2-ibg-volume-3-building-products-and-furnishings-part-b.pdf}{\emph{IFRS S2 Industry-based Guidance on implementing Climate-related Disclosures Vol 3 — Building Products \& Furnishings}} \cite{ifrs_s2_vol03_building_products_furnishings_2023} & 2023 & 203 KB & 13 & 1 \\[2pt]
      \href{https://www.ifrs.org/content/dam/ifrs/publications/pdf-standards-issb/english/2023/issued/part-b/ifrs-s2-ibg-volume-4-e-commerce-and-furnishings-part-b.pdf}{\emph{IFRS S2 Industry-based Guidance on implementing Climate-related Disclosures Vol 4 — E-Commerce}} \cite{ifrs_s2_vol04_e_commerce_2023} & 2023 & 199 KB & 11 & 2 \\[2pt]
      \href{https://www.ifrs.org/content/dam/ifrs/publications/pdf-standards-issb/english/2023/issued/part-b/ifrs-s2-ibg-volume-5-households-and-personal-products-part-b.pdf}{\emph{IFRS S2 Industry-based Guidance on implementing Climate-related Disclosures Vol 5 — Household \& Personal Products}} \cite{ifrs_s2_vol05_household_personal_products_2023} & 2023 & 191 KB & 9 & 2 \\[2pt]
      \href{https://www.ifrs.org/content/dam/ifrs/publications/pdf-standards-issb/english/2023/issued/part-b/ifrs-s2-ibg-volume-6-multiline-and-specialty-retailers-and-distributors-part-b.pdf}{\emph{IFRS S2 Industry-based Guidance on implementing Climate-related Disclosures Vol 6 — Multiline and Specialty Retailers \& Distributors}} \cite{ifrs_s2_vol06_multiline_specialty_retail_2023} & 2023 & 180 KB & 7 & 0 \\[2pt]
      \href{https://www.ifrs.org/content/dam/ifrs/publications/pdf-standards-issb/english/2023/issued/part-b/ifrs-s2-ibg-volume-7-coal-operations-part-b.pdf}{\emph{IFRS S2 Industry-based Guidance on implementing Climate-related Disclosures Vol 7 — Coal Operations}} \cite{ifrs_s2_vol07_coal_operations_2023} & 2023 & 216 KB & 14 & 1 \\[2pt]
      \href{https://www.ifrs.org/content/dam/ifrs/publications/pdf-standards-issb/english/2023/issued/part-b/ifrs-s2-ibg-volume-8-construction-materials-part-b.pdf}{\emph{IFRS S2 Industry-based Guidance on implementing Climate-related Disclosures Vol 8 — Construction Materials}} \cite{ifrs_s2_vol08_construction_materials_2023} & 2023 & 232 KB & 17 & 1 \\[2pt]
      \href{https://www.ifrs.org/content/dam/ifrs/publications/pdf-standards-issb/english/2023/issued/part-b/ifrs-s2-ibg-volume-9-iron-and-steel-producers-part-b.pdf}{\emph{IFRS S2 Industry-based Guidance on implementing Climate-related Disclosures Vol 9 — Iron \& Steel Producers}} \cite{ifrs_s2_vol09_iron_steel_producers_2023} & 2023 & 210 KB & 14 & 1 \\[2pt]
      \href{https://www.ifrs.org/content/dam/ifrs/publications/pdf-standards-issb/english/2023/issued/part-b/ifrs-s2-ibg-volume-10-metals-and-mining-part-b.pdf}{\emph{IFRS S2 Industry-based Guidance on implementing Climate-related Disclosures Vol 10 — Metals \& Mining}} \cite{ifrs_s2_vol10_metals_mining_2023} & 2023 & 204 KB & 12 & 1 \\[2pt]
      \href{https://www.ifrs.org/content/dam/ifrs/publications/pdf-standards-issb/english/2023/issued/part-b/ifrs-s2-ibg-volume-11-oil-and-gas-exploration-and-production-part-b.pdf}{\emph{IFRS S2 Industry-based Guidance on implementing Climate-related Disclosures Vol 11 — Oil \& Gas – Exploration \& Production}} \cite{ifrs_s2_vol11_oil_gas_ep_2023} & 2023 & 244 KB & 20 & 2 \\[2pt]
      \href{https://www.ifrs.org/content/dam/ifrs/publications/pdf-standards-issb/english/2023/issued/part-b/ifrs-s2-ibg-volume-12-oil-and-gas-midstream-part-b.pdf}{\emph{IFRS S2 Industry-based Guidance on implementing Climate-related Disclosures Vol 12 — Oil \& Gas – Midstream}} \cite{ifrs_s2_vol12_oil_gas_midstream_2023} & 2023 & 196 KB & 10 & 3 \\[2pt]
      \href{https://www.ifrs.org/content/dam/ifrs/publications/pdf-standards-issb/english/2023/issued/part-b/ifrs-s2-ibg-volume-13-oil-and-gas-refining-and-marketing-part-b.pdf}{\emph{IFRS S2 Industry-based Guidance on implementing Climate-related Disclosures Vol 13 — Oil \& Gas – Refining \& Marketing}} \cite{ifrs_s2_vol13_oil_gas_refining_marketing_2023} & 2023 & 212 KB & 13 & 1 \\[2pt]
      \href{https://www.ifrs.org/content/dam/ifrs/publications/pdf-standards-issb/english/2023/issued/part-b/ifrs-s2-ibg-volume-14-oil-and-gas-services-part-b.pdf}{\emph{IFRS S2 Industry-based Guidance on implementing Climate-related Disclosures Vol 14 — Oil \& Gas – Services}} \cite{ifrs_s2_vol14_oil_gas_services_2023} & 2023 & 202 KB & 11 & 1 \\[2pt]
      \href{https://www.ifrs.org/content/dam/ifrs/publications/pdf-standards-issb/english/2023/issued/part-b/ifrs-s2-ibg-volume-15-asset-management-and-custody-activities-part-b.pdf}{\emph{IFRS S2 Industry-based Guidance on implementing Climate-related Disclosures Vol 15 — Asset Management \& Custody Activities}} \cite{ifrs_s2_vol15_asset_mgmt_custody_2023} & 2023 & 202 KB & 11 & 4 \\[2pt]
      \href{https://www.ifrs.org/content/dam/ifrs/publications/pdf-standards-issb/english/2023/issued/part-b/ifrs-s2-ibg-volume-16-commercial-banks-part-b.pdf}{\emph{IFRS S2 Industry-based Guidance on implementing Climate-related Disclosures Vol 16 — Commercial Banks}} \cite{ifrs_s2_vol16_commercial_banks_2023} & 2023 & 189 KB & 8 & 1 \\[2pt]
      \href{https://www.ifrs.org/content/dam/ifrs/publications/pdf-standards-issb/english/2023/issued/part-b/ifrs-s2-ibg-volume-17-insurance-part-b.pdf}{\emph{IFRS S2 Industry-based Guidance on implementing Climate-related Disclosures Vol 17 — Insurance}} \cite{ifrs_s2_vol17_insurance_2023} & 2023 & 223 KB & 15 & 1 \\[2pt]
      \href{https://www.ifrs.org/content/dam/ifrs/publications/pdf-standards-issb/english/2023/issued/part-b/ifrs-s2-ibg-volume-18-investment-banking-and-brokerage-part-b.pdf}{\emph{IFRS S2 Industry-based Guidance on implementing Climate-related Disclosures Vol 18 — Investment Banking \& Brokerage}} \cite{ifrs_s2_vol18_investment_banking_brokerage_2023} & 2023 & 201 KB & 10 & 2 \\[2pt]
      \href{https://www.ifrs.org/content/dam/ifrs/publications/pdf-standards-issb/english/2023/issued/part-b/ifrs-s2-ibg-volume-19-mortgage-finance-part-b.pdf}{\emph{IFRS S2 Industry-based Guidance on implementing Climate-related Disclosures Vol 19 — Mortgage Finance}} \cite{ifrs_s2_vol19_mortgage_finance_2023} & 2023 & 188 KB & 7 & 1 \\[2pt]
      \href{https://www.ifrs.org/content/dam/ifrs/publications/pdf-standards-issb/english/2023/issued/part-b/ifrs-s2-ibg-volume-20-agricultural-products-part-b.pdf}{\emph{IFRS S2 Industry-based Guidance on implementing Climate-related Disclosures Vol 20 — Agricultural Products}} \cite{ifrs_s2_vol20_agricultural_products_2023} & 2023 & 223 KB & 17 & 1 \\[2pt]
      \href{https://www.ifrs.org/content/dam/ifrs/publications/pdf-standards-issb/english/2023/issued/part-b/ifrs-s2-ibg-volume-21-alcoholic-beverages-part-b.pdf}{\emph{IFRS S2 Industry-based Guidance on implementing Climate-related Disclosures Vol 21 — Alcoholic Beverages}} \cite{ifrs_s2_vol21_alcoholic_beverages_2023} & 2023 & 209 KB & 13 & 1 \\[2pt]
      \href{https://www.ifrs.org/content/dam/ifrs/publications/pdf-standards-issb/english/2023/issued/part-b/ifrs-s2-ibg-volume-22-food-retailers-and-distributors-part-b.pdf}{\emph{IFRS S2 Industry-based Guidance on implementing Climate-related Disclosures Vol 22 — Food Retailers \& Distributors}} \cite{ifrs_s2_vol22_food_retailers_distributors_2023} & 2023 & 219 KB & 15 & 1 \\[2pt]
      \href{https://www.ifrs.org/content/dam/ifrs/publications/pdf-standards-issb/english/2023/issued/part-b/ifrs-s2-ibg-volume-23-meat-poultry-and-dairy-part-b.pdf}{\emph{IFRS S2 Industry-based Guidance on implementing Climate-related Disclosures Vol 23 — Meat, Poultry \& Dairy}} \cite{ifrs_s2_vol23_meat_poultry_dairy_2023} & 2023 & 235 KB & 19 & 1 \\[2pt]
      \href{https://www.ifrs.org/content/dam/ifrs/publications/pdf-standards-issb/english/2023/issued/part-b/ifrs-s2-ibg-volume-24-non-alcoholic-beverages-part-b.pdf}{\emph{IFRS S2 Industry-based Guidance on implementing Climate-related Disclosures Vol 24 — Non-Alcoholic Beverages}} \cite{ifrs_s2_vol24_non_alcoholic_beverages_2023} & 2023 & 217 KB & 15 & 1 \\[2pt]
      \href{https://www.ifrs.org/content/dam/ifrs/publications/pdf-standards-issb/english/2023/issued/part-b/ifrs-s2-ibg-volume-25-processed-foods-part-b.pdf}{\emph{IFRS S2 Industry-based Guidance on implementing Climate-related Disclosures Vol 25 — Processed Foods}} \cite{ifrs_s2_vol25_processed_foods_2023} & 2023 & 219 KB & 15 & 0 \\[2pt]
      \href{https://www.ifrs.org/content/dam/ifrs/publications/pdf-standards-issb/english/2023/issued/part-b/ifrs-s2-ibg-volume-26-restaurants-part-b.pdf}{\emph{IFRS S2 Industry-based Guidance on implementing Climate-related Disclosures Vol 26  - Restaurants}} \cite{ifrs_s2_vol26_restaurants_2023} & 2023 & 202 KB & 11 & 3 \\[2pt]
      \href{https://www.ifrs.org/content/dam/ifrs/publications/pdf-standards-issb/english/2023/issued/part-b/ifrs-s2-ibg-volume-27-drug-retailers-part-b.pdf}{\emph{IFRS S2 Industry-based Guidance on implementing Climate-related Disclosures Vol 27 — Drug Retailers}} \cite{ifrs_s2_vol27_drug_retailers_2023} & 2023 & 186 KB & 7 & 2 \\[2pt]
      \href{https://www.ifrs.org/content/dam/ifrs/publications/pdf-standards-issb/english/2023/issued/part-b/ifrs-s2-ibg-volume-28-health-care-delivery-part-b.pdf}{\emph{IFRS S2 Industry-based Guidance on implementing Climate-related Disclosures Vol 28 — Health Care Delivery}} \cite{ifrs_s2_vol28_health_care_delivery_2023} & 2023 & 201 KB & 11 & 2 \\[2pt]
      \href{https://www.ifrs.org/content/dam/ifrs/publications/pdf-standards-issb/english/2023/issued/part-b/ifrs-s2-ibg-volume-29-health-care-distributors-part-b.pdf}{\emph{IFRS S2 Industry-based Guidance on implementing Climate-related Disclosures Vol 29 — Health Care Distributors}} \cite{ifrs_s2_vol29_health_care_distributors_2023} & 2023 & 180 KB & 6 & 2 \\[2pt]
      \href{https://www.ifrs.org/content/dam/ifrs/publications/pdf-standards-issb/english/2023/issued/part-b/ifrs-s2-ibg-volume-30-managed-care-part-b.pdf}{\emph{IFRS S2 Industry-based Guidance on implementing Climate-related Disclosures Vol 30 — Managed Care}} \cite{ifrs_s2_vol30_managed_care_2023} & 2023 & 174 KB & 6 & 1 \\
      \bottomrule
    \end{tabular}
    \caption{IFRS / ISSB sustainability disclosure materials — \textbf{Part I}.}
    \label{tab:ifrsissb_part1_table}
  \end{table*}
  \clearpage
  
  \begin{table*}[tb]
    \centering
    \scriptsize
    \setlength\tabcolsep{3pt}
    \begin{tabular}{p{12cm} c c c c}
      \toprule
      \textbf{Original Document Title} &
      \textbf{Year} & \textbf{Size} & \textbf{Pages} & \textbf{No. Qs} \\
      \midrule
      % ---------- IFRS S2 industry-based guidance Vols 31 – 68 ----------
      \href{https://www.ifrs.org/content/dam/ifrs/publications/pdf-standards-issb/english/2023/issued/part-b/ifrs-s2-ibg-volume-31-medical-equipment-and-supplies-part-b.pdf}{\emph{IFRS S2 Industry-based Guidance on implementing Climate-related Disclosures Vol 31 — Medical Equipment \& Supplies}} \cite{ifrs_s2_vol31_medical_equipment_supplies_2023} & 2023 & 183 KB & 7 & 2 \\[2pt]
      \href{https://www.ifrs.org/content/dam/ifrs/publications/pdf-standards-issb/english/2023/issued/part-b/ifrs-s2-ibg-volume-32-electric-utilities-and-power-generators-part-b.pdf}{\emph{IFRS S2 Industry-based Guidance on implementing Climate-related Disclosures Vol 32 — Electric Utilities \& Power Generators}} \cite{ifrs_s2_vol32_electric_utilities_power_generators_2023} & 2023 & 257 KB & 24 & 3 \\[2pt]
      \href{https://www.ifrs.org/content/dam/ifrs/publications/pdf-standards-issb/english/2023/issued/part-b/ifrs-s2-ibg-volume-33-engineering-and-construction-services-part-b.pdf}{\emph{IFRS S2 Industry-based Guidance on implementing Climate-related Disclosures Vol 33 — Engineering \& Construction Services}} \cite{ifrs_s2_vol33_engineering_construction_services_2023} & 2023 & 222 KB & 16 & 1 \\[2pt]
      \href{https://www.ifrs.org/content/dam/ifrs/publications/pdf-standards-issb/english/2023/issued/part-b/ifrs-s2-ibg-volume-34-gas-utilities-and-distributors-part-b.pdf}{\emph{IFRS S2 Industry-based Guidance on implementing Climate-related Disclosures Vol 34 — Gas Utilities \& Distributors}} \cite{ifrs_s2_vol34_gas_utilities_distributors_2023} & 2023 & 210 KB & 13 & 1 \\[2pt]
      \href{https://www.ifrs.org/content/dam/ifrs/publications/pdf-standards-issb/english/2023/issued/part-b/ifrs-s2-ibg-volume-35-home-builders-part-b.pdf}{\emph{IFRS S2 Industry-based Guidance on implementing Climate-related Disclosures Vol 35 — Home Builders}} \cite{ifrs_s2_vol35_home_builders_2023} & 2023 & 214 KB & 13 & 1 \\[2pt]
      \href{https://www.ifrs.org/content/dam/ifrs/publications/pdf-standards-issb/english/2023/issued/part-b/ifrs-s2-ibg-volume-36-real-estate-part-b.pdf}{\emph{IFRS S2 Industry-based Guidance on implementing Climate-related Disclosures Vol 36 — Real Estate}} \cite{ifrs_s2_vol36_real_estate_2023} & 2023 & 285 KB & 34 & 3 \\[2pt]
      \href{https://www.ifrs.org/content/dam/ifrs/publications/pdf-standards-issb/english/2023/issued/part-b/ifrs-s2-ibg-volume-37-real-estate-services-part-b.pdf}{\emph{IFRS S2 Industry-based Guidance on implementing Climate-related Disclosures Vol 37 — Real Estate Services}} \cite{ifrs_s2_vol37_real_estate_services_2023} & 2023 & 197 KB & 9 & 2 \\[2pt]
      \href{https://www.ifrs.org/content/dam/ifrs/publications/pdf-standards-issb/english/2023/issued/part-b/ifrs-s2-ibg-volume-38-waste-management-part-b.pdf}{\emph{IFRS S2 Industry-based Guidance on implementing Climate-related Disclosures Vol 38 — Waste Management}} \cite{ifrs_s2_vol38_waste_management_2023} & 2023 & 215 KB & 12 & 1 \\[2pt]
      \href{https://www.ifrs.org/content/dam/ifrs/publications/pdf-standards-issb/english/2023/issued/part-b/ifrs-s2-ibg-volume-39-water-utilities-and-services-part-b.pdf}{\emph{IFRS S2 Industry-based Guidance on implementing Climate-related Disclosures Vol 39 — Water Utilities \& Services}} \cite{ifrs_s2_vol39_water_utilities_services_2023} & 2023 & 241 KB & 19 & 1 \\[2pt]
      \href{https://www.ifrs.org/content/dam/ifrs/publications/pdf-standards-issb/english/2023/issued/part-b/ifrs-s2-ibg-volume-40-biofuels-part-b.pdf}{\emph{IFRS S2 Industry-based Guidance on implementing Climate-related Disclosures Vol 40 — Biofuels}} \cite{ifrs_s2_vol40_biofuels_2023} & 2023 & 208 KB & 14 & 0 \\[2pt]
      \href{https://www.ifrs.org/content/dam/ifrs/publications/pdf-standards-issb/english/2023/issued/part-b/ifrs-s2-ibg-volume-41-forestry-management-part-b.pdf}{\emph{IFRS S2 Industry-based Guidance on implementing Climate-related Disclosures Vol 41 — Forestry Management}} \cite{ifrs_s2_vol41_forestry_management_2023} & 2023 & 209 KB & 12 & 1 \\[2pt]
      \href{https://www.ifrs.org/content/dam/ifrs/publications/pdf-standards-issb/english/2023/issued/part-b/ifrs-s2-ibg-volume-42-fuel-cells-and-industrial-batteries-part-b.pdf}{\emph{IFRS S2 Industry-based Guidance on implementing Climate-related Disclosures Vol 42 — Fuel Cells \& Industrial Batteries}} \cite{ifrs_s2_vol42_fuel_cells_industrial_batteries_2023} & 2023 & 202 KB & 10 & 3 \\[2pt]
      \href{https://www.ifrs.org/content/dam/ifrs/publications/pdf-standards-issb/english/2023/issued/part-b/ifrs-s2-ibg-volume-43-pulp-and-paper-products-part-b.pdf}{\emph{IFRS S2 Industry-based Guidance on implementing Climate-related Disclosures Vol 43 — Pulp \& Paper Products}} \cite{ifrs_s2_vol43_pulp_paper_products_2023} & 2023 & 202 KB & 18 & 1 \\[2pt]
      \href{https://www.ifrs.org/content/dam/ifrs/publications/pdf-standards-issb/english/2023/issued/part-b/ifrs-s2-ibg-volume-44-solar-technology-and-project-developers-part-b.pdf}{\emph{IFRS S2 Industry-based Guidance on implementing Climate-related Disclosures Vol 44 — Solar Technology \& Project Developers}} \cite{ifrs_s2_vol44_solar_tech_project_developers_2023} & 2023 & 216 KB & 14 & 1 \\[2pt]
      \href{https://www.ifrs.org/content/dam/ifrs/publications/pdf-standards-issb/english/2023/issued/part-b/ifrs-s2-ibg-volume-45-wind-technology-and-project-developers-part-b.pdf}{\emph{IFRS S2 Industry-based Guidance on implementing Climate-related Disclosures Vol 45 — Wind Technology \& Project Developers}} \cite{ifrs_s2_vol45_wind_tech_project_developers_2023} & 2023 & 195 KB & 8 & 2 \\[2pt]
      \href{https://www.ifrs.org/content/dam/ifrs/publications/pdf-standards-issb/english/2023/issued/part-b/ifrs-s2-ibg-volume-46-aerospace-and-defence-part-b.pdf}{\emph{IFRS S2 Industry-based Guidance on implementing Climate-related Disclosures Vol 46 — Aerospace \& Defence}} \cite{ifrs_s2_vol46_aerospace_defense_2023} & 2023 & 204 KB & 9 & 2 \\[2pt]
      \href{https://www.ifrs.org/content/dam/ifrs/publications/pdf-standards-issb/english/2023/issued/part-b/ifrs-s2-ibg-volume-47-chemicals-part-b.pdf}{\emph{IFRS S2 Industry-based Guidance on implementing Climate-related Disclosures Vol 47 — Chemicals}} \cite{ifrs_s2_vol47_chemicals_2023} & 2023 & 224 KB & 16 & 2 \\[2pt]
      \href{https://www.ifrs.org/content/dam/ifrs/publications/pdf-standards-issb/english/2023/issued/part-b/ifrs-s2-ibg-volume-48-containers-and-packaging-part-b.pdf}{\emph{IFRS S2 Industry-based Guidance on implementing Climate-related Disclosures Vol 48 — Containers \& Packaging}} \cite{ifrs_s2_vol48_containers_packaging_2023} & 2023 & 232 KB & 17 & 2 \\[2pt]
      \href{https://www.ifrs.org/content/dam/ifrs/publications/pdf-standards-issb/english/2023/issued/part-b/ifrs-s2-ibg-volume-49-electrical-and-electronic-equipment-part-b.pdf}{\emph{IFRS S2 Industry-based Guidance on implementing Climate-related Disclosures Vol 49 — Electrical \& Electronic Equipment}} \cite{ifrs_s2_vol49_electrical_electronic_equipment_2023} & 2023 & 202 KB & 10 & 2 \\[2pt]
      \href{https://www.ifrs.org/content/dam/ifrs/publications/pdf-standards-issb/english/2023/issued/part-b/ifrs-s2-ibg-volume-50-industrial-machinery-and-goods-part-b.pdf}{\emph{IFRS S2 Industry-based Guidance on implementing Climate-related Disclosures Vol 50 — Industrial Machinery \& Goods}} \cite{ifrs_s2_vol50_industrial_machinery_goods_2023} & 2023 & 196 KB & 9 & 3 \\[2pt]
      \href{https://www.ifrs.org/content/dam/ifrs/publications/pdf-standards-issb/english/2023/issued/part-b/ifrs-s2-ibg-volume-51-casinos-and-gaming-part-b.pdf}{\emph{IFRS S2 Industry-based Guidance on implementing Climate-related Disclosures Vol 51 — Casinos \& Gaming}} \cite{ifrs_s2_vol51_casinos_gaming_2023} & 2023 & 189 KB & 7 & 3 \\[2pt]
      \href{https://www.ifrs.org/content/dam/ifrs/publications/pdf-standards-issb/english/2023/issued/part-b/ifrs-s2-ibg-volume-52-hotels-and-lodging-part-b.pdf}{\emph{IFRS S2 Industry-based Guidance on implementing Climate-related Disclosures Vol 52 — Hotels \& Lodging}} \cite{ifrs_s2_vol52_hotels_lodging_2023} & 2023 & 201 KB & 9 & 4 \\[2pt]
      \href{https://www.ifrs.org/content/dam/ifrs/publications/pdf-standards-issb/english/2023/issued/part-b/ifrs-s2-ibg-volume-53-leisure-facilities-part-b.pdf}{\emph{IFRS S2 Industry-based Guidance on implementing Climate-related Disclosures Vol 53 — Leisure Facilities}} \cite{ifrs_s2_vol53_leisure_facilities_2023} & 2023 & 188 KB & 7 & 1 \\[2pt]
      \href{https://www.ifrs.org/content/dam/ifrs/publications/pdf-standards-issb/english/2023/issued/part-b/ifrs-s2-ibg-volume-54-electronic-manufacturing-services-and-original-design-manufacturing-part-b.pdf}{\emph{IFRS S2 Industry-based Guidance on implementing Climate-related Disclosures Vol 54 — Electronic Mfg Services \& ODM}} \cite{ifrs_s2_vol54_ems_odm_2023} & 2023 & 196 KB & 8 & 1 \\[2pt]
      \href{https://www.ifrs.org/content/dam/ifrs/publications/pdf-standards-issb/english/2023/issued/part-b/ifrs-s2-ibg-volume-55-hardware-part-b.pdf}{\emph{IFRS S2 Industry-based Guidance on implementing Climate-related Disclosures Vol 55 — Hardware}} \cite{ifrs_s2_vol55_hardware_2023} & 2023 & 200 KB & 10 & 2 \\[2pt]
      \href{https://www.ifrs.org/content/dam/ifrs/publications/pdf-standards-issb/english/2023/issued/part-b/ifrs-s2-ibg-volume-56-internet-media-and-services-part-b.pdf}{\emph{IFRS S2 Industry-based Guidance on implementing Climate-related Disclosures Vol 56 — Internet Media \& Services}} \cite{ifrs_s2_vol56_internet_media_services_2023} & 2023 & 200 KB & 9 & 2 \\[2pt]
      \href{https://www.ifrs.org/content/dam/ifrs/publications/pdf-standards-issb/english/2023/issued/part-b/ifrs-s2-ibg-volume-57-semiconductors-part-b.pdf}{\emph{IFRS S2 Industry-based Guidance on implementing Climate-related Disclosures Vol 57 — Semiconductors}} \cite{ifrs_s2_vol57_semiconductors_2023} & 2023 & 222 KB & 14 & 0 \\[2pt]
      \href{https://www.ifrs.org/content/dam/ifrs/publications/pdf-standards-issb/english/2023/issued/part-b/ifrs-s2-ibg-volume-58-software-and-it-services-part-b.pdf}{\emph{IFRS S2 Industry-based Guidance on implementing Climate-related Disclosures Vol 58 — Software \& IT Services}} \cite{ifrs_s2_vol58_software_it_services_2023} & 2023 & 208 KB & 11 & 3 \\[2pt]
      \href{https://www.ifrs.org/content/dam/ifrs/publications/pdf-standards-issb/english/2023/issued/part-b/ifrs-s2-ibg-volume-59-telecommunication-services-part-b.pdf}{\emph{IFRS S2 Industry-based Guidance on implementing Climate-related Disclosures Vol 59 — Telecommunication Services}} \cite{ifrs_s2_vol59_telecommunication_services_2023} & 2023 & 205 KB & 10 & 1 \\[2pt]
      \href{https://www.ifrs.org/content/dam/ifrs/publications/pdf-standards-issb/english/2023/issued/part-b/ifrs-s2-ibg-volume-60-air-freight-and-logistics-part-b.pdf}{\emph{IFRS S2 Industry-based Guidance on implementing Climate-related Disclosures Vol 60 — Air Freight \& Logistics}} \cite{ifrs_s2_vol60_air_freight_logistics_2023} & 2023 & 202 KB & 11 & 2 \\[2pt]
      \href{https://www.ifrs.org/content/dam/ifrs/publications/pdf-standards-issb/english/2023/issued/part-b/ifrs-s2-ibg-volume-61-airlines-part-b.pdf}{\emph{IFRS S2 Industry-based Guidance on implementing Climate-related Disclosures Vol 61 — Airlines}} \cite{ifrs_s2_vol61_airlines_2023} & 2023 & 201 KB & 9 & 3 \\[2pt]
      \href{https://www.ifrs.org/content/dam/ifrs/publications/pdf-standards-issb/english/2023/issued/part-b/ifrs-s2-ibg-volume-62-auto-parts-part-b.pdf}{\emph{IFRS S2 Industry-based Guidance on implementing Climate-related Disclosures Vol 62 — Auto Parts}} \cite{ifrs_s2_vol62_auto_parts_2023} & 2023 & 192 KB & 8 & 3 \\[2pt]
      \href{https://www.ifrs.org/content/dam/ifrs/publications/pdf-standards-issb/english/2023/issued/part-b/ifrs-s2-ibg-volume-63-automobiles-part-b.pdf}{\emph{IFRS S2 Industry-based Guidance on implementing Climate-related Disclosures Vol 63 — Automobiles}} \cite{ifrs_s2_vol63_automobiles_2023}  & 2023 & 193 KB & 8 & 3 \\[2pt]
      \href{https://www.ifrs.org/content/dam/ifrs/publications/pdf-standards-issb/english/2023/issued/part-b/ifrs-s2-ibg-volume-64-car-rental-and-leasing-part-b.pdf}{\emph{IFRS S2 Industry-based Guidance on implementing Climate-related Disclosures Vol 64 — Car Rental \& Leasing}} \cite{ifrs_s2_vol64_car_rental_leasing_2023} & 2023 & 189 KB & 7 & 2 \\[2pt]
      \href{https://www.ifrs.org/content/dam/ifrs/publications/pdf-standards-issb/english/2023/issued/part-b/ifrs-s2-ibg-volume-65-cruise-lines-part-b.pdf}{\emph{IFRS S2 Industry-based Guidance on implementing Climate-related Disclosures Vol 65 — Cruise Lines}} \cite{ifrs_s2_vol65_cruise_lines_2023} & 2023 & 206 KB & 10 & 3 \\[2pt]
      \href{https://www.ifrs.org/content/dam/ifrs/publications/pdf-standards-issb/english/2023/issued/part-b/ifrs-s2-ibg-volume-66-marine-transportation-part-b.pdf}{\emph{IFRS S2 Industry-based Guidance on implementing Climate-related Disclosures Vol 66 — Marine Transportation}} \cite{ifrs_s2_vol66_marine_transportation_2023} & 2023 & 208 KB & 10 & 2 \\[2pt]
      \href{https://www.ifrs.org/content/dam/ifrs/publications/pdf-standards-issb/english/2023/issued/part-b/ifrs-s2-ibg-volume-67-rail-transportation-part-b.pdf}{\emph{IFRS S2 Industry-based Guidance on implementing Climate-related Disclosures Vol 67 — Rail Transportation}} \cite{ifrs_s2_vol67_rail_transportation_2023} & 2023 & 201 KB & 9 & 1 \\[2pt]
      \href{https://www.ifrs.org/content/dam/ifrs/publications/pdf-standards-issb/english/2023/issued/part-b/ifrs-s2-ibg-volume-68-road-transportation-part-b.pdf}{\emph{IFRS S2 Industry-based Guidance on implementing Climate-related Disclosures Vol 68 — Road Transportation}} \cite{ifrs_s2_vol68_road_transportation_2023} & 2023 & 200 KB & 9 & 3 \\
      \midrule
      Total & - & 20.81MB & 1,866 & 238 \\
      \bottomrule
    \end{tabular}
    \caption{Comprehensive metadata for the entire \textbf{IFRS / ISSB sustainability-disclosure corpus} (2023–2025): it comprises the universal core standards \emph{IFRS S1} (general sustainability-related financial disclosure) and \emph{IFRS S2} (climate-related disclosure), their bases for conclusions, accompanying guidance, the 2024 climate-progress report, the 2024 Sustainability Disclosure Taxonomy, the 2025 greenhouse-gas exposure draft (with basis), a comparison of \emph{IFRS S2} with the TCFD recommendations, and the 68-volume \emph{IFRS S2 Industry-based Guidance} that maps SASB’s sector-specific materiality into the new framework. Together the 77 PDFs, total \textbf{20.81 MB}, span \textbf{1,866 pages}—well over two thousand when including ancillary matter—and yield \textbf{238} benchmark MCQs for \textbf{ESGenius}. \emph{IFRS S1} sets the universal disclosure baseline, while \emph{IFRS S2} details climate-specific metrics, mirroring the TCFD architecture and enriched with SASB’s sectoral depth; the accompanying materials establish a globally consistent, investor-focused baseline that links decision-relevant sustainability information—such as greenhouse-gas emissions, transition plans, climate resilience, data security, and workforce diversity—directly to financial statements. All documents are freely available at \url{https://www.ifrs.org/}.}
    \label{tab:ifrsissb_part2_table}
  \end{table*}
  \clearpage
  
\begin{table*}[tb]
    \centering
    \scriptsize
    \setlength\tabcolsep{3pt}
    \begin{tabular}{p{12cm} c c c c}
      \toprule
      \textbf{Original Document Title} &
      \textbf{Year} & \textbf{Size} & \textbf{Pages} & \textbf{No. Qs} \\
      \midrule
      \href{https://assets.bbhub.io/company/sites/60/2022/12/tcfd-2022-overview-booklet.pdf}{\emph{Task Force on Climate-related Financial Disclosures Overview}} \cite{tcfd_overview_2022} & 2022 & 11.1 MB & 25 & 2 \\[2pt]
      \href{https://assets.bbhub.io/company/sites/60/2021/07/2021-Metrics_Targets_Guidance-1.pdf}{\emph{Task Force on Climate-related Financial Disclosures Guidance on Metrics, Targets, and Transition Plans}} \cite{tcfd_metrics_targets_transition_2021} & 2021 & 12 MB & 79 & 11 \\[2pt]
      \href{https://assets.bbhub.io/company/sites/60/2021/07/2021-TCFD-Implementing_Guidance.pdf}{\emph{Task Force on Climate-related Financial Disclosures Implementing the Recommendations of the Task Force on Climate-related Financial Disclosures}} \cite{tcfd_implementing_recommendations_2021} & 2021 & 1.1 MB & 88 & 18 \\[2pt]
      \href{https://assets.bbhub.io/company/sites/60/2022/02/TCFD-Fundamentals-Workshop.pdf}{\emph{TCFD Workshop – Session 1: Fundamentals and Overview of TCFD}} \cite{tcfd_workshop_session1_2022}& 2022 & 3.2 MB & 40 & 7 \\[2pt]
      \href{https://assets.bbhub.io/company/sites/60/2022/02/TCFD-Governance-Workshop.pdf}{\emph{TCFD Workshop – Session 2: Governance}} \cite{tcfd_workshop_session2_2022} & 2022 & 1.4 MB & 20 & 1 \\[2pt]
      \href{https://assets.bbhub.io/company/sites/60/2022/02/TCFD-Strategy-Workshop.pdf}{\emph{TCFD Workshop – Session 3: Strategy}} \cite{tcfd_workshop_session3_2022}& 2022 & 2.7 MB & 43 & 8 \\[2pt]
      \href{https://assets.bbhub.io/company/sites/60/2022/02/TCFD-Risk-Management-Workshop.pdf}{\emph{TCFD Workshop – Session 4: Risk Management}} \cite{tcfd_workshop_session4_2022} & 2022 & 1.4 MB & 34 & 4 \\[2pt]
      \href{https://assets.bbhub.io/company/sites/60/2022/02/Metrics-and-Targets-Workshop.pdf}{\emph{TCFD Workshop – Session 5: Metrics and Targets}} \cite{tcfd_workshop_session5_2022} & 2022 & 3 MB & 53 & 9 \\[2pt]
      \href{https://assets.bbhub.io/company/sites/60/2020/09/2020-TCFD_Guidance-Scenario-Analysis-Guidance.pdf}{\emph{Task Force on Climate-related Financial Disclosures Guidance on Scenario Analysis for Non-Financial Companies}} \cite{tcfd_scenario_analysis_guidance_2020} & 2020 & 3.7 MB & 133 & 22 \\[2pt]
      \href{https://assets.bbhub.io/company/sites/60/2020/09/2020-TCFD_Guidance-Risk-Management-Integration-and-Disclosure.pdf}{\emph{Task Force on Climate-related Financial Disclosures Guidance on Risk Management Integration and Disclosure}} \cite{tcfd_risk_mgmt_integration_guidance_2020} & 2020 & 5.2 MB & 52 & 9 \\[2pt]
      \href{https://assets.bbhub.io/company/sites/60/2023/09/2023-Status-Report.pdf}{\emph{Task Force on Climate-related Financial Disclosures 2023 Status Report}} \cite{tcfd_status_report_2023} & 2023 & 19.3 MB & 161 & 32 \\[2pt]
      \midrule
      Total & - & 60 MB & 728 & 123 \\
      \bottomrule
    \end{tabular}
    \caption{Comprehensive metadata for the \textbf{Task Force on Climate-related Financial Disclosures (TCFD)} corpus included in \textbf{ESGenius}. The collection spans 2020--2023 and comprises 11 documents totaling \textbf{60 MB} across \textbf{728 pages}, yielding \textbf{123} benchmark questions. The materials include core guidance documents on implementation, metrics, targets, scenario analysis and risk management, a complete five-part workshop series covering fundamentals through metrics, and the 2023 status report. These documents establish the foundation for global climate-risk disclosure practices and form the conceptual framework adopted by ISSB's IFRS S2. All materials are accessible at \url{https://www.fsb-tcfd.org/}.}
    \label{tab:tcfd_table}
  \end{table*}
  \clearpage

\begin{table*}[tb]
    \centering
    \scriptsize
    \setlength\tabcolsep{3pt}
    \begin{tabular}{p{10cm} c c c c}
      \toprule
      \textbf{Original Document Title} &
      \textbf{Year} & \textbf{Size} & \textbf{Pages} & \textbf{No. Qs} \\
      \midrule
      \href{https://assets.ctfassets.net/v7uy4j80khf8/2vwnUxFROogGypKMp6P5wq/39a028b0f0385eef77e1fbe5253a16fa/Full_Corporate_Questionnaire_Modules_1-6.pdf}{\emph{CDP Full Corporate Questionnaire April 2025 – Modules 1–6}} \cite{cdp_full_corp_mod1_6_2025} & 2025 & 5.1 MB & 447 & 29 \\[2pt]
      \href{https://assets.ctfassets.net/v7uy4j80khf8/32GtEECo2X1qWmaOdN2Luy/6909cdbbb2fad3aeddce8944109081ae/Full_Corporate_Questionnaire_Module_7.pdf}{\emph{CDP Full Corporate Questionnaire April 2025 – Module 7}} \cite{cdp_full_corp_mod7_2025}& 2025 & 4.9 MB & 482 & 41 \\[2pt]
      \href{https://assets.ctfassets.net/v7uy4j80khf8/3od5x6nkcnaOx9psMV2YBu/c64b1a8a8d5c5f86c9ff630d7f28bd35/Full_Corporate_Questionnaire_Modules_8-13.pdf}{\emph{CDP Full Corporate Questionnaire April 2025 – Modules 8–13}} \cite{cdp_full_corp_mod8_13_2025} & 2025 & 4.7 MB & 435 & 44 \\[2pt]
      \href{https://assets.ctfassets.net/v7uy4j80khf8/5jaubP76UEYtQfASBd9JgK/b443421754a817ec84e2a45584037f03/SME_Questionnaire_Modules_14-21.pdf}{\emph{CDP SME Questionnaire April 2025 – Modules 14–21}} \cite{cdp_sme_mod14_21_2025} & 2025 & 2.1 MB & 174 & 38 \\[2pt]      
      \href{https://assets.ctfassets.net/v7uy4j80khf8/g2pV5HJDE4LFwsSOjYvU2/8bb279ccccea3528671cd2e2bb41550a/2025_CDP-ICLEI_Track_Questionnaire_and_Guidance.pdf}{\emph{2025 CDP–ICLEI Track and States \& Regions Questionnaire and Guidance}} \cite{cdp_iclei_track_sr_quest_2025} & 2025 & 2.5 MB & 430 & 31 \\[2pt]     
      \href{https://assets.ctfassets.net/v7uy4j80khf8/6Q9DPfzvSaFJpBX1zNV0Vi/577510c8e20852cf1eda8ef39f4f3ab7/CDP_Full_Corporate_Scoring_Introduction_2024.pdf}{\emph{CDP Full Corporate Scoring Introduction 2024}} \cite{cdp_scoring_intro_2024} & 2024 & 293 KB & 19 & 2 \\[2pt] 
      \midrule
      Total & - & 19.6 MB & 1,987 & 185 \\
      \bottomrule
    \end{tabular}
    \caption{Comprehensive metadata for the \textbf{Carbon Disclosure Project (CDP)} knowledge base curated in \textbf{ESGenius}: the collection comprises six hyper-linked PDFs including the \emph{2025 Full Corporate Questionnaire} split across three modules (114 questions, 1,364 pages), the \emph{2025 SME Questionnaire} (38 questions, 174 pages), the \emph{2025 CDP–ICLEI Track and States \& Regions Questionnaire} (31 questions, 430 pages), and the \emph{2024 Full Corporate Scoring Introduction} (2 scoring criteria, 19 pages). Together the corpus totals \textbf{19.6 MB} and spans \textbf{1,987 pages}, delivering \textbf{185 unique, standardised questions} that power the world's largest voluntary platform for climate, water-security, and deforestation disclosure. All files are directly accessible via the linked titles, with consolidated resources available at \url{https://www.cdp.net/}.}
    \label{tab:cdp_table}
  \end{table*}
  \clearpage

\begin{table*}[tb]
    \centering
    \scriptsize
    \setlength\tabcolsep{3pt}
    \begin{tabular}{p{12cm} c c c c}
      \toprule
      \textbf{Original Document Title} &
      \textbf{Year} & \textbf{Size} & \textbf{Pages} & \textbf{No. Qs} \\
      \midrule
      \href{https://sustainabledevelopment.un.org/content/documents/21252030%20Agenda%20for%20Sustainable%20Development%20web.pdf}{\emph{The 2030 Agenda for Sustainable Development's 17 Sustainable Development Goals (SDGs)}} \cite{sdg_17_goals_2015} & 2015 & 424 KB & 19 & 1 \\[2pt] 
      \href{https://sdgs.un.org/sites/default/files/publications/21252030%20Agenda%20for%20Sustainable%20Development%20web.pdf}{\emph{Transforming Our World: The 2030 Agenda for Sustainable Development}} \cite{transforming_our_world_2015} & 2015 & 378 KB & 41 & 4 \\[2pt] 
      \href{https://unstats.un.org/sdgs/files/report/2024/SG-SDG-Progress-Report-2024-advanced-unedited-version.pdf}{\emph{Progress towards the Sustainable Development Goals — Report of the Secretary-General}} \cite{sdg_progress_report_2024} & 2024 & 518 KB & 26 & 2 \\[2pt] 
      \href{https://unstats.un.org/sdgs/report/2024/The-Sustainable-Development-Goals-Report-2024.pdf}{\emph{The Sustainable Development Goals Report 2024}} \cite{sdg_report_2024} & 2024 & 8.6 MB & 51 & 6 \\[2pt]  
      \href{https://unpartnerships.un.org/sites/default/files/publications/2024-01/SDG%20Briefing%20Book_2023.pdf}{\emph{Sustainable Development Goals – Briefing Book 2023 (UN Office for Partnerships)}} \cite{sdg_briefing_book_2023} & 2023 & 4.6 MB & 35 & 7 \\[2pt] 
      \href{https://www.undrr.org/media/88718}{\emph{GAR Special Report 2023 — Mapping Resilience for the Sustainable Development Goals}} \cite{gar_special_report_2023} & 2023 & 19.5 MB & 51 & 5 \\[2pt] 
      \href{https://s3.amazonaws.com/sustainabledevelopment.report/2024/sustainable-development-report-2024.pdf}{\emph{Sustainable Development Report 2024 — The SDGs and the UN Summit of the Future (includes the SDG Index and Dashboards)}} \cite{sdr_global_2024} & 2024 & 39.3 MB & 512 & 12 \\
      \midrule
      Total & - & 73.33 MB & 735 & 37 \\
      \bottomrule
    \end{tabular}
    \caption{Comprehensive metadata for the \textbf{United Nations Sustainable Development Goals (SDGs)} corpus incorporated in \textbf{ESGenius}: the set begins with the 2015 adoption texts—the 19-page plain-language overview of the 17 Goals and the 41-page General Assembly resolution ``Transforming Our World'' that enshrines the 2030 Agenda—then follows implementation through the Secretary-General’s \emph{Progress towards the SDGs 2024} and the flagship \emph{SDGs Report 2024}, is complemented by the UN Office for Partnerships’ \emph{SDG Briefing Book 2023} and deepened by thematic analyses such as UNDRR’s 2023 \emph{Global Assessment Report on Resilience} (GAR) and the independent \emph{Sustainable Development Report 2024} with its widely cited SDG Index and global dashboards. Across these seven key PDFs—totalling \textbf{73.33 MB}, \textbf{735 pages}, and distilled into \textbf{37} benchmark MCQs that anchors national, corporate, and investor sustainability strategies in the universally agreed 17-goal framework. Full SDG resources are freely available at \url{https://sdgs.un.org/goals}.}
    \label{tab:sdg_table}
  \end{table*}
  \clearpage

  \begin{figure*}[h!]
    \centering
    \includegraphics[width=\textwidth]{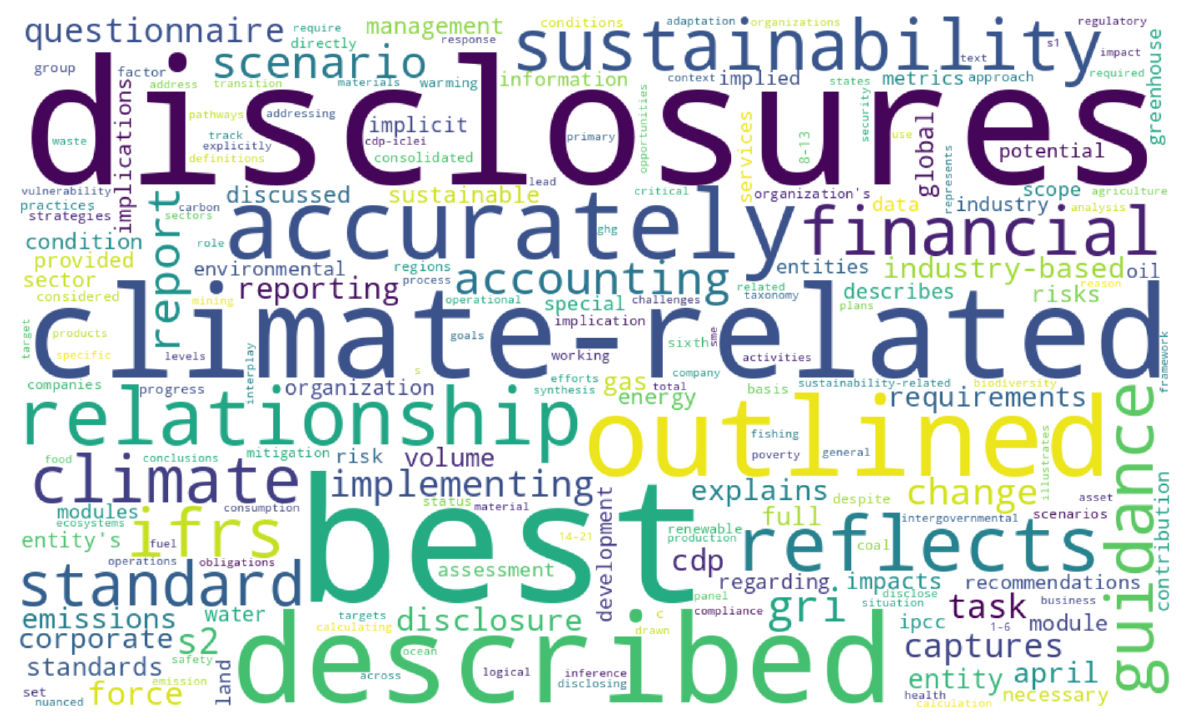}
    \caption{This cloud visualizes the 1,136 question stems after filtering generic fillers. Dominant terms such as ``disclosures'', ``climate-related'', ``sustainability'', ``accurately'', and ``IFRS'' reveal that the questions emphasize reporting frameworks and precision in interpreting ESG guidance. The prominence of verbs like ``described'', ``outlined'', and ``reflects'' indicates a consistent demand for higher-order reasoning (e.g., identifying relationships, implications, or best interpretations rather than simple fact recall).}
    \label{fig:esgenius_qa_question_wordcloud}
\end{figure*}

\begin{figure*}[h!]
  \centering
  \includegraphics[width=\textwidth]{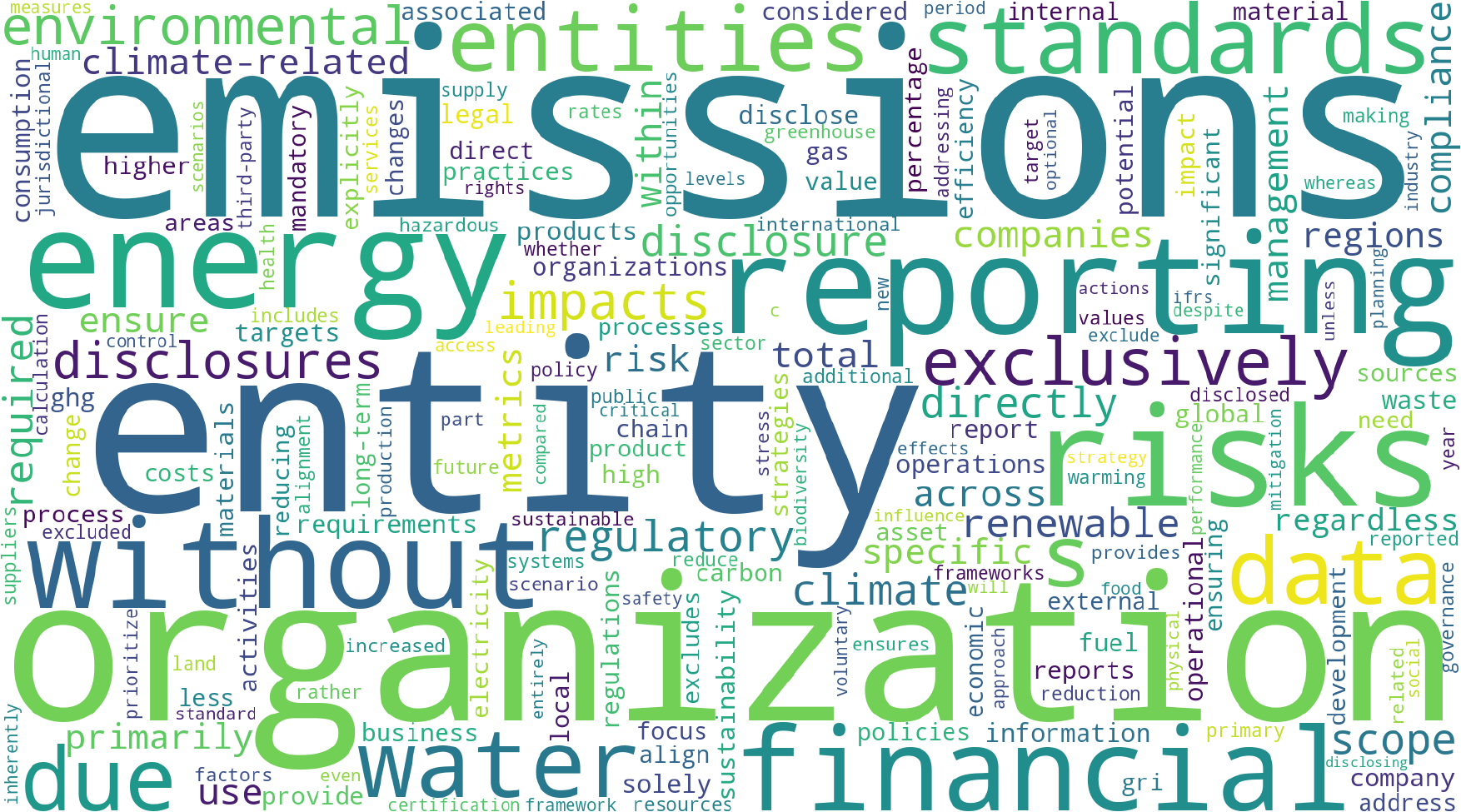}
  \caption{The aggregate vocabulary of the 5,680 answer choices centers on the same core ESG-reporting nouns that dominate the source text---``emissions,'' ``entity,'' ``organization,'' ``energy,'' ``water,'' ``risks,'' and ``reporting''---but it is studded with decisive qualifiers such as ``without,'' ``due,'' ``exclusively,'' ``primarily,'' and ``regardless.'' These modifiers reveal how distractors are engineered: they adjust scope, responsibility, or conditionality to make each option plausible while still allowing only one to satisfy the nuanced criteria posed by the question.}
  \label{fig:esgenius_qa_option_wordcloud}
\end{figure*}

\begin{figure*}[h!]
  \centering
  \includegraphics[width=\textwidth]{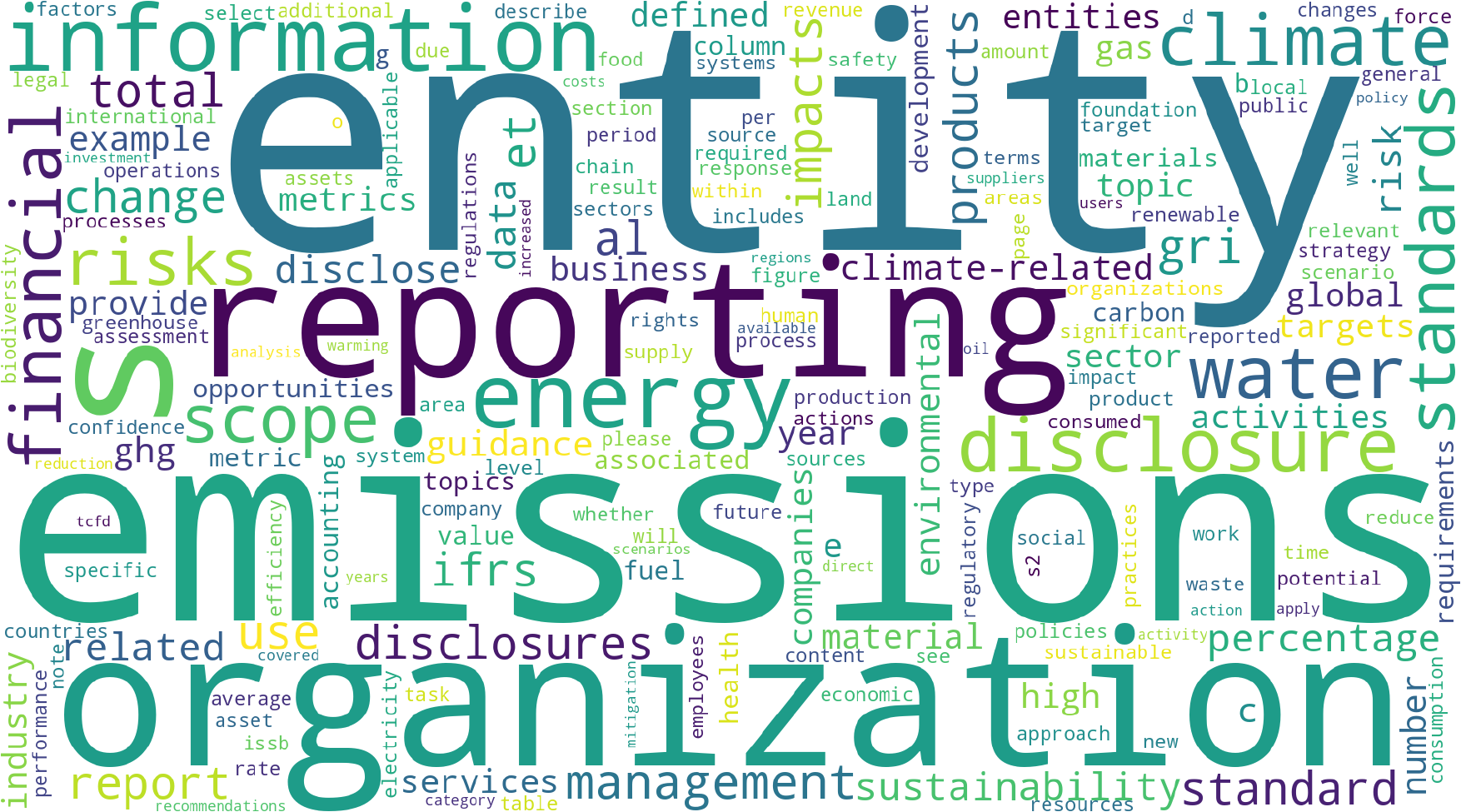}
  \caption{Drawn from the reference excerpts linked to each question, this word cloud highlights the knowledge backbone behind the benchmark. Key nouns---``entity'', ``organization'', ``emissions'', ``energy'', ``water'', ``reporting'', ``standards''---underscore the dataset's strong focus on corporate disclosure obligations, accounting boundaries (Scope 1/2/3 GHG), and resource-specific metrics. Frequent technical modifiers such as ``financial'', ``material'', ``percentage'', ``management'' show that passages often quantify impacts or prescribe measurement criteria, aligning with the analytical depth expected of the MCQs.}
  \label{fig:esgenius_corpus_wordcloud}
\end{figure*}

\clearpage

\clearpage

\begin{figure*}[h!]
  \centering
  \includegraphics[width=\textwidth]{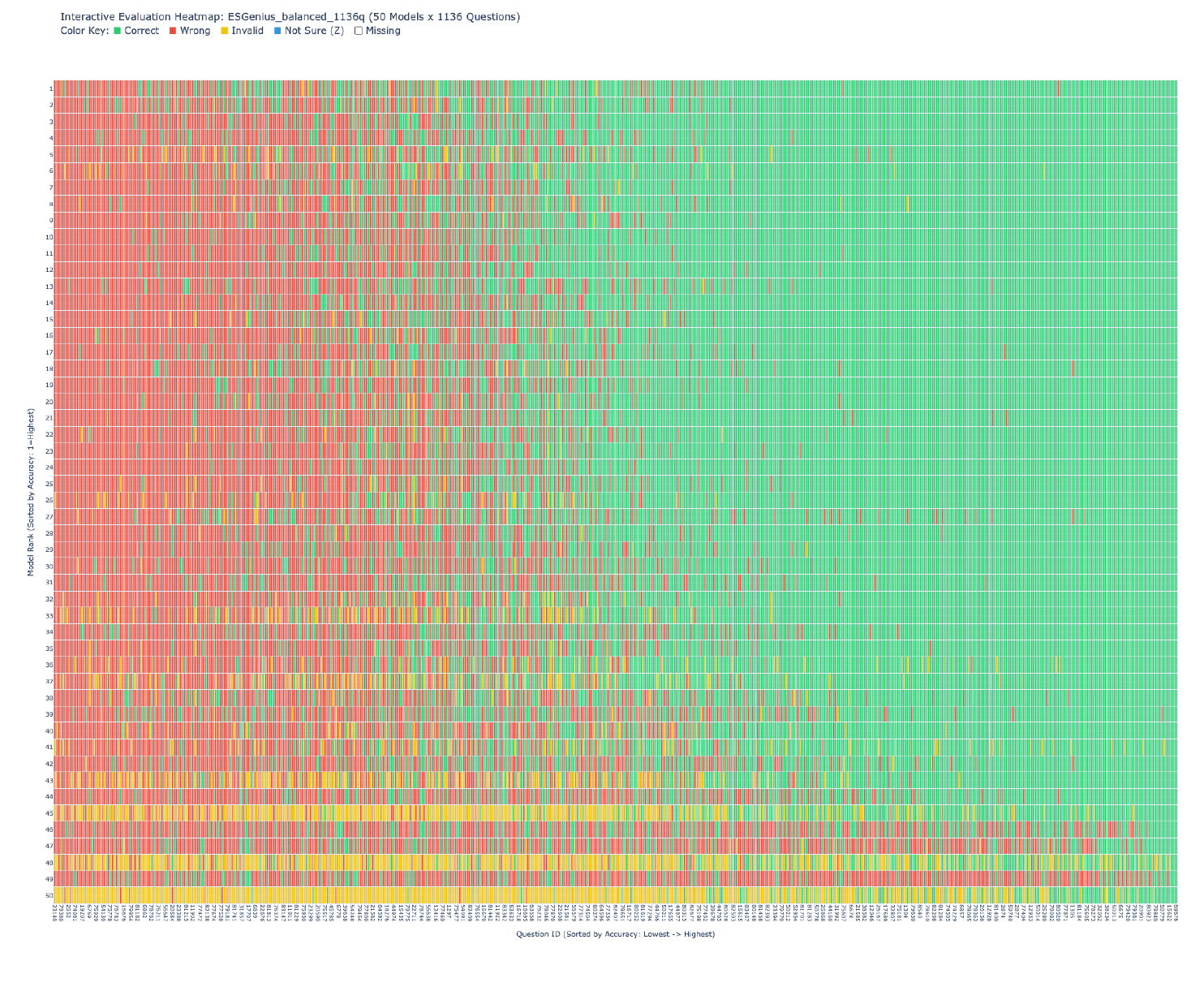}
  \caption{Example zero-shot performance heatmap showing model accuracy patterns across different ESG question types and topics within ESGenius (based on initial data). Darker colors indicate higher accuracy.}
  \label{fig:zero_shot_heatmap}
\end{figure*}

\begin{figure*}[h!]
  \centering
  \includegraphics[width=\textwidth]{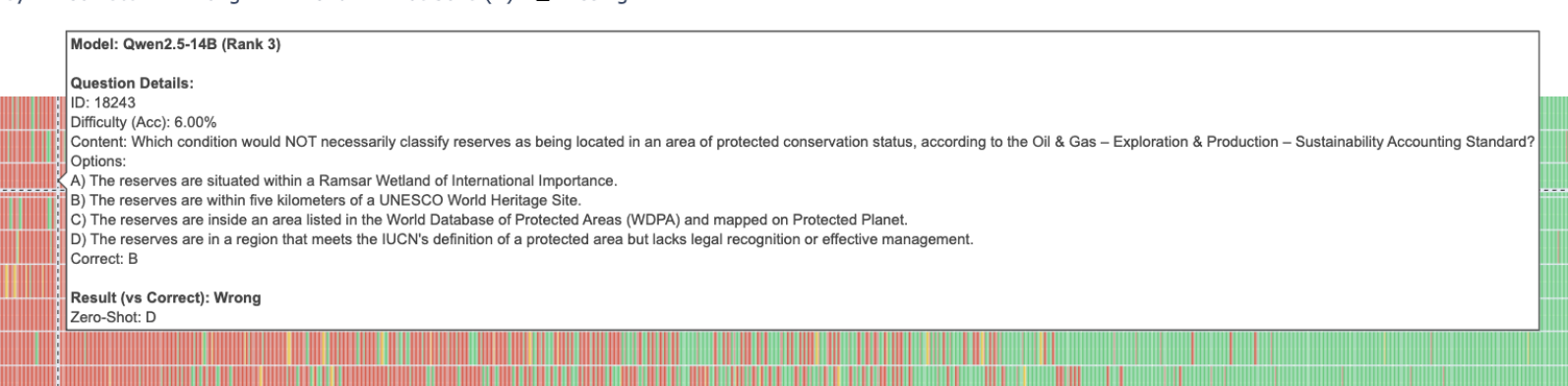}
  \caption{Conceptual information display for interactive heatmap cells (Figure~\ref{fig:zero_shot_heatmap}), showing model and accuracy score.}
  \label{fig:heatmap_cell_info}
\end{figure*}
% \section*{Author Checklist Compliance}
% \label{sec:checklist}

% This work complies with the ARR Author Checklist guidelines.\footnote{\url{https://aclrollingreview.org/authorchecklist}} Specifically:
% \begin{itemize}
%     \item \textbf{Use of AI Assistants:} We used AI assistants (e.g., ChatGPT) to polish and refine the presentation of this paper, particularly for improving clarity, fluency, and organization. No content, ideas, or experiments were generated by AI tools.
%     \item \textbf{Ethical Considerations:} No personally identifiable information (PII), sensitive data, or unethical content was used or produced.
%     \item \textbf{Bias and Fairness:} We recognize limitations in model generalizability and discuss them in Section~\ref{sec:limitations}.
%     \item \textbf{Reproducibility:} Experimental settings, evaluation protocols, and model specifications are described clearly to facilitate reproducibility.
% \end{itemize}
\end{document}